\documentclass[notitlepage,11pt]{article}

\usepackage{natbib}
\usepackage{appendix}
\usepackage{setspace}
\usepackage{epsfig}
\usepackage{fullpage}
\usepackage{geometry}
\usepackage{amssymb}
\usepackage{amsmath}
\usepackage{amsthm}
\usepackage{tikz}
\usepackage{graphicx}
\usepackage{algorithm}
\usepackage{algorithmic}
\let\Algorithm\algorithm
\renewcommand\algorithm[1][]{\Algorithm[#1]\setstretch{1.3}}
\usepackage{hyperref}
\usepackage{longtable}
\usepackage{supertabular}
\usepackage{upgreek}
\usepackage{multirow}
\usepackage{changepage}
\usepackage{pdflscape}
\usepackage[affil-it]{authblk}
\newcommand{\vect}[1]{\boldsymbol{#1}}

\title{PLUTO: Penalized Unbiased Logistic Regression Trees}
\author{Wenwen Zhang and Wei-Yin Loh}
\affil{Department of Statistics, University of Wisconsin-Madison}

\date{November 24, 2014}

\begin{document}

\maketitle
%\doublespacing
\begin{center}
\section*{Abstract}
\end{center}
We propose a new algorithm called PLUTO for building logistic regression trees to binary response data. PLUTO can capture the nonlinear and interaction patterns in messy data by recursively partitioning the sample space. It fits a simple or a multiple linear logistic regression model in each partition. PLUTO employs the cyclical coordinate descent method for estimation of multiple linear logistic regression models with elastic net penalties, which allows it to deal with high-dimensional data efficiently. The tree structure comprises a graphical description of the data. Together with the logistic regression models, it provides an accurate classifier as well as a piecewise smooth estimate of the probability of ``success". PLUTO controls selection bias by: (1) separating split variable selection from split point selection; (2) applying an adjusted chi-squared test to find the split variable instead of exhaustive search. A bootstrap calibration technique is employed to further correct selection bias. Comparison on real datasets shows that on average, the multiple linear PLUTO models predict more accurately than other algorithms. \\~\\
\emph{Keywords:} Adjusted chi-squared test, Elastic net penalty, Logistic regression tree, Recursive partitioning, Selection bias.

\newpage
\section{Introduction}
Logistic regression and tree-based models are two popular methods for modeling binary response data. The former is a traditional method relying on classical statistical principle, while the latter is a nontraditional machine learning tool. Logistic regression provides a smooth estimate to the probability of ``success". However, when the pattern that holds in the data is complex, it is often difficult to find a logistic regression model that is satisfactory. Furthermore, the fitted model may be hard to interpret. On the other hand, tree-based models can detect nonlinear and interaction patterns in the data automatically by recursively partitioning the sample space. Besides, the tree structure gives a graphical presentation of the data.

Logistic regression tree is the result of combining logistic regression and tree-based models. It partitions sample space by a sequence of splits, and fits logistic regression models in the nodes. Logistic regression tree inherits the interpretability of tree-based models, and provides a smooth solution to classification problems in the nodes like logistic regression. Several methods have been proposed to build logistic regression trees, but there is still space for improvement.

In this paper, we present a new algorithm called PLUTO, which incorporates the logistic regression model to the tree structure. PLUTO stands for \emph{P}enalized, \emph{L}ogistic regression, \emph{U}nbiased splitting, \emph{T}ree \emph{O}perator.

The rest of the paper is organized as follows:
\begin{description}
\item[Section 2] provides an overview of the logistic regression, regularized logistic regression, and tree-based models. Existing logistic regression tree algorithms are also reviewed in this section.
\item[Section 3] describes the PLUTO algorithm in detail, in order of split variable selection, split point/subset selection, tree size determination, and importance ranking.
\item[Section 4] discusses the selection bias problem, as well as how we make the selection bias negligible using a bootstrap technique. Simulation results of split variable selection and selection bias correction are also presented.
\item[Section 5] presents a comparison between PLUTO and competing algorithms in terms of prediction accuracy on real datasets.
\item[Section 6] shows an application of PLUTO to the census income dataset.
\item[Section 7] summarizes this paper and suggests future work.
\end{description}

%%%%%%%%%%%%%%%%%%%%%%%%%%%%%%%%%%%%%%%%%%%%%%%%%%%%%%%%%%%%%%%%%%%%%%%%%%%%%%%%%%%%%%%%%%%%%%%%%
\section{Background}
PLUTO is built upon classic statistical models and innovative machine learning methods. This section reviews the fundamental building blocks that inspired and enabled us to develop the PLUTO algorithm. A brief overview of existing logistic regression tree algorithms is presented at the end of this section.

\subsection{Logistic regression}
\subsubsection{Notation and parameter estimation}
Logistic regression is a widely used method for modeling binary data. It employs a regression method to solve classification tasks. Denote a binary response by $Y$ and its two possible outcomes by 0 and 1. The distribution of $Y$ is specified by the probability of ``success" $p = \mbox{Pr}(Y=1)$. Given $K$ predictors $\vect{X} = (X_1,\ldots,X_K)$, linear logistic regression relates $p$ to a linear predictor $\eta = \beta_0 + \beta_1 X_1 + \ldots + \beta_K X_K$ via a logit link function $\eta = \mbox{logit}(p) = \log [p/(1-p)]$. Here predictors $X_k,\ k = 1,\ldots,K$ may be numerical variables or dummy-coded categorical variables, and categorical variables may be ordinal or nominal. Let $N$ denote the sample size and let $\{\vect{x_i}=(x_{i1},\ldots,x_{iK})',y_i\}$ denote the values for the $i$th observation of $\{X,Y\}$, $i = 1,\ldots,N$. The coefficients $\beta_0$ and $\boldsymbol{\beta} = (\beta_1,\ldots,\beta_K)'$ are estimated by maximizing the log-likelihood function:
\begin{equation}\label{equ:1}
\ell(\beta_0,\boldsymbol{\beta}) = \sum_{i = 1}^N [ y_i \log p_i + (1-y_i) \log (1-p_i) ],
\end{equation}
where
\begin{equation}\label{equ:2}
p_i = \exp (\beta_0 + \vect{x_i}'\boldsymbol{\beta})/\{1+\exp (\beta_0 + \vect{x_i}'\boldsymbol{\beta}) \}.
\end{equation}
Pluging (\ref{equ:2}) into (\ref{equ:1}), $\ell(\beta_0,\boldsymbol{\beta})$ can be rewritten as:
\begin{equation}\label{equ:11}
\ell(\beta_0,\boldsymbol{\beta}) = \sum_{i = 1}^N \{ y_i(\beta_0 + \vect{x_i}'\boldsymbol{\beta}) - \log [1+\exp (\beta_0 + \vect{x_i}'\boldsymbol{\beta})] \}.
\end{equation}
The maximum likelihood estimates (MLE) of $\beta_0$ and $\boldsymbol{\beta}$ are the solution to a set of score equations:
\begin{equation}\label{equ:12}
S_i(\beta_0,\boldsymbol{\beta}) = \partial \ell(\beta_0,\boldsymbol{\beta})/\partial \beta_i,\quad i = 0, \ldots, K,
\end{equation}
which are nonlinear in $\beta_0$ and $\boldsymbol{\beta}$. Hence, there is no closed-form expression for the MLEs of $\beta_0$ and $\boldsymbol{\beta}$. However, they can be solved iteratively by numerical methods such as the Newton-Raphson method \citep{mccu:neld:1989}. Suppose the current estimates of the parameters are $(\tilde{\beta}_0,\tilde{\boldsymbol{\beta}})$. A quadratic approximation to the log-likelihood function $\ell(\beta_0,\boldsymbol{\beta})$ (\ref{equ:11}) is:
\begin{equation}\label{equ:6}
\ell_{Q}(\beta_0,\boldsymbol{\beta}) = -\frac{1}{2}\sum_{i = 1}^N w_i(v_i-\beta_0-\vect{x_i}'\boldsymbol{\beta})^2 + \mathcal{C}(\tilde{\beta}_0,\tilde{\boldsymbol{\beta}}),
\end{equation}
where
\begin{equation}
\begin{split}
\label{equ:7}
v_i &= \tilde{\beta}_0+\vect{x_i}'\tilde{\boldsymbol{\beta}}+\frac{y_i - \tilde{p}_i}{\tilde{p}_i (1-\tilde{p}_i)},\\
w_i &= \tilde{p}_i (1-\tilde{p}_i).
\end{split}
\end{equation}
The term $\mathcal{C}(\tilde{\beta}_0,\tilde{\boldsymbol{\beta}})$ is a constant and $\tilde{p}_i$ is calculated from (\ref{equ:2}) using current parameter values. The Newton-Raphson method updates $(\beta_0,\boldsymbol{\beta})$ by minimizing $\ell_{Q}$. If we consider $v_i$ as the response and $w_i$ as the weight, minimizing $\ell_{Q}$ also leads to an iteratively reweighted least squares (IRLS) procedure.

\subsubsection{Model checking}
When the model fits well to the data, logistic regression provides not only an accurate
classifier, but also a smooth estimate for the probability of ``success". Furthermore, if the fitted model is simple, we can interpret the estimated coefficients conveniently, in terms of $p$ and the odds ratio $p/(1-p)$.

However, there are also weaknesses with logistic regression. First, in the presence of nonlinearity, collinearity and/or interactions among predictor variables, it is often challenging for us to manually select a well-fitted model with the limited diagnostic tools of logistic regression. Second, model checking for logistic regression can be difficult. The well-known goodness-of-fit analysis uses a Pearson $\chi^2$ or likelihood-ratio $G^2$ test statistic. Under
logistic regression model, especially with numerical predictors, the large sample theory may not apply \citep[see][page 112]{agresti1996}. In other words, $\chi^2$ and $G^2$ may not follow chi-squared distributions asymptotically. Lastly, complex logistic regression models are generally hard to interpret.

\subsubsection{An example}
To illustrate, we consider the heart disease dataset obtained from the UCI Machine Learning Repository \citep{uci}. The purpose of this dataset is to predict the presence or absence of heart disease given the results of various medical tests carried out on a patient. It contains 270 observations and 13 predictors. The variables are listed in Table \ref{T1:heart}.
\begin{table}[!htb]
\begin{center}
\caption{Variables for heart disease data}\label{T1:heart}
\begin{tabular}{l l c}
\hline
\hline
Name & Values & Type\textsuperscript{\dag} \\
\hline
\texttt{age} & age in years & N \\
\texttt{sex} & gender (1 = male; 0 = female) & C\\
\texttt{cp} & chest pain type (values 1, 2, 3, 4) & C \\
\texttt{rbp} & resting blood pressure & N\\
\texttt{chol} & serum cholesterol in mg/dl & N \\
\texttt{fbs} & I(fasting blood sugar $>$ 120 mg/dl) & C \\
\texttt{restecg} & resting electrocardiographic results (values 0, 1, 2) & C\\
\texttt{mhr} & maximum heart rate achieved & N\\
\texttt{eia} & exercise induced angina (1 = yes; 0 = no) & C\\
\texttt{oldpeak} & ST depression induced by exercise relative to rest & N\\
\texttt{slope} & the slope of the peak exercise ST segment & N\\
\texttt{nmv} & number of major vessels colored by fluoroscopy & N\\
\texttt{thal} & 3 = normal; 6 = fixed defect; 7 = reversible defect & C\\
\texttt{y} & presence of heart disease (1 = present; 0 = absent)  & R\\
\hline
\multicolumn{3}{l}{\textsuperscript{\dag}\footnotesize{Variable type:``N"--numerical predictor, ``C"--categorical predictor, ``R"--response.}}
\end{tabular}
\end{center}
\end{table}

We first fit a multiple linear logistic regression model with stepwise variable selection to the untransformed data using SAS software. The fitted model is:
\begin{equation}\label{equ:ex1}
\begin{split}
\mbox{logit}(p) = & -3.384 -0.861 \mbox{ I}(\texttt{sex}=0)-0.919 \mbox{ I}(\texttt{cp}=1)+ 0.065 \mbox{ I}(\texttt{cp}=2)\\
                  & -0.592 \mbox{ I}(\texttt{cp}=3) + 0.022 \texttt{ rbp} + 0.008 \texttt{ chol} -0.027 \texttt{ mhr} + 0.620  \texttt{ oldpeak}\\
                  & +1.042 \texttt{ nmv}-0.463 \mbox{ I}(\texttt{thal}=3)-0.517 \mbox{ I}(\texttt{thal}=6).
\end{split}
\end{equation}
Eight predictor variable are selected. Next, we use likelihood-ratio tests to compare Model (\ref{equ:ex1}) with reduced models. Test results suggest that the parameters of indicators $\mbox{I}(\texttt{cp}=2)$ and $\mbox{I}(\texttt{thal}=6)$ are not significant. Also, the parameters of $\mbox{I}(\texttt{cp}=1)$ and $\mbox{I}(\texttt{cp}=3)$ are found to be equivalent. Therefore, we introduce transformed variables $\texttt{cpI} = \mbox{I}(\texttt{cp} \in \{1,3\})$ and $\texttt{thalI} = \mbox{I}(\texttt{thal}=3)$. Set $\texttt{sexI} = \mbox{I}(\texttt{sex}=0)$ and $\texttt{eiaI} = \mbox{I}(\texttt{eia}=0)$. Fitting a stepwise logistic regression using the transformed variables, we obtain a simpler model:
\begin{equation}\label{equ:ex2}
\begin{split}
\mbox{logit}(p) = & -3.014 -0.804 \texttt{ sexI} -0.805 \texttt{ cpI} + 0.018 \texttt{ rbp}+0.008 \texttt{ chol}-0.022 \texttt{ mhr}\\
                  & -0.435 \texttt{ eiaI} + 0.624 \texttt{ oldpeak}+ 1.071 \texttt{ nmv} -0.653  \texttt{ thalI}.
\end{split}
\end{equation}
To check if any of the numerical predictors has a non-linear effect, we perform power transformations and use the deviances for comparison. No such transformation is found to be necessary. Finally, interaction effects are examined. We fit a logistic regression model with all the main effects in Model (\ref{equ:ex2}) and their pairwise interactions. Then we apply stepwise variable selection to the fitted model, resulting in the final model:
\begin{equation}\label{equ:ex3}
\begin{split}
\mbox{logit}(p) = & -4.765 -0.787 \texttt{ sexI} -0.825 \texttt{ cpI} + 0.018 \texttt{ rbp}+0.016 \texttt{ chol}-0.023 \texttt{ mhr}\\
                  & -0.431 \texttt{ eiaI} + 2.289 \texttt{ oldpeak}+ 1.094 \texttt{ nmv} -0.706 \texttt{ thalI}\\
                  & -0.006 \texttt{ chol}*\texttt{oldpeak}.
\end{split}
\end{equation}
According to Wald tests, all the parameters in Model (\ref{equ:ex3}) are significant at $5\%$ significance level except $\texttt{chol}*\texttt{oldpeak}$ ($p$-value = 0.09). Due to the complexity of the final model, it is hard to interpret the results. Besides this, the ``hand-crafted" model building procedure is very subjective and extremely inefficient for routine data analysis.

\subsection{Regularized logistic regression} \label{sec:bg:glmnet}
There is a flourishing interest in regularization methods over the past couple decades within the machine learning community. Ridge regression \citep{RidgeReg} solves least square linear regression with an $\ell_2$ penalty. Because of the nature of the $\ell_2$ norm constraint, it tends to shrink the coefficients. \citet{LASSO} introduced the LASSO, which stands for ``Least Absolute Shrinkage and Selection Operator". LASSO is similar to ridge regression, but it employs the $\ell_1$ penalty instead of $\ell_2$ to achieve a sparse solution. LASSO not only shrinks the coefficients as ridge regression does, but also performs variable selection. \citet{ElasticNet} proposed the elastic net regression, which applies a mixture of the LASSO and ridge penalties. The elastic net possesses the model shrinking ability like the ridge penalty, and it can perform variable selection like the LASSO. These regularization methods have been extended to generalized linear models \citep{GLMNET}.

Adding an elastic net penalty to logistic regression, the coefficients $\beta_0$ and $\boldsymbol{\beta}$ are estimated by maximizing the penalized log-likelihood function:
\begin{equation}\label{equ:4}
\max_{(\beta_0,\boldsymbol{\beta}) \in \mathbb{R}^{K+1}} \left \{\sum_{i = 1}^N [ y_i \log p_i + (1-y_i) \log (1-p_i)]-\lambda P_{\alpha}(\boldsymbol{\beta}) \right \},
\end{equation}
where
\begin{equation}\label{equ:5}
P_{\alpha}(\boldsymbol{\beta}) = (1-\alpha)\frac{1}{2}\|\boldsymbol{\beta}\|_{\ell_2}^2 + \alpha \|\boldsymbol{\beta}\|_{\ell_1}.
\end{equation}
$P_{\alpha}$ is the elastic net penalty which includes the ridge penalty ($\alpha = 0$) and the LASSO penalty ($\alpha = 1$) as special cases. This penalty is particularly useful in the $K \ge N$ situation, or any situation where there are many correlated predictor variables.

\citet*{GLMNET} developed fast algorithms for estimation of generalized linear models with elastic net penalties (GLMNET). Their algorithms use cyclical coordinate descent, computed along a regularization path. For each value of $\lambda$, an outer loop is created which updates the quadratic approximation $\ell_{Q}$ (\ref{equ:6}) using the current parameter values $(\tilde{\beta}_0,\tilde{\boldsymbol{\beta}})$. Then the cyclical coordinate descent algorithm is applied to solve a penalized weighted least-squares problem:
\begin{equation}\label{equ:8}
\min_{(\beta_0,\boldsymbol{\beta}) \in \mathbb{R}^{K+1}} \{ -\ell_{Q}(\beta_0,\boldsymbol{\beta}) + \lambda P_{\alpha}(\boldsymbol{\beta}) \}.
\end{equation}
The GLMNET algorithm is implemented in the R programming system \citep{Rcite} as a package ``glmnet" and is very efficient.

The convergence properties of coordinate descent in convex problems is discussed in \citet{Tseng2001}.

\subsection{Tree-based models}
Over the past fifty years, tree-based models have been developing steadily as computation power increases. By recursively partitioning the sample space, a tree model can automatically capture the nonlinear and interaction patterns in the data. In each partition, the relation between response and predictors become less complicated, and may be described in terms of a constant category/number or a linear model containing a small number of variables. Also, the tree structure provides a graphical interpretation of the data. In practice, tree models often provide insight into the data that may be hard to achieve using other methods.

\citet*{morg:sonq:1963} and \citet{Fielding1977} first incorporated the tree structure in the Automated Interaction Detection (AID) algorithm. They failed however, to provide a stopping rule that is satisfactory.

\citet*{cart} then developed the CART\textregistered \ algorithm which generates binary classification trees and piecewise-constant regression trees. It improved AID by introducing a pruning process for determining the final tree. Both AID and CART trees apply exhaustive search for partitions and have the problem of selection bias towards variables with more splits.

Following CART, a number of tree-based algorithms were developed to employ different splitting and node fitting rules. To name a few of the more well known ones, there is C4.5 \citep{Quinlan:1992},
SUPPORT \citep{Chaudhuri:Huang:1994}, QUEST \citep{Loh:Shih:1997}, and CRUISE \citep{Kim:Loh:2001}.

\begin{figure}[!ht]
                \centering
                \includegraphics[width=0.8\textwidth]{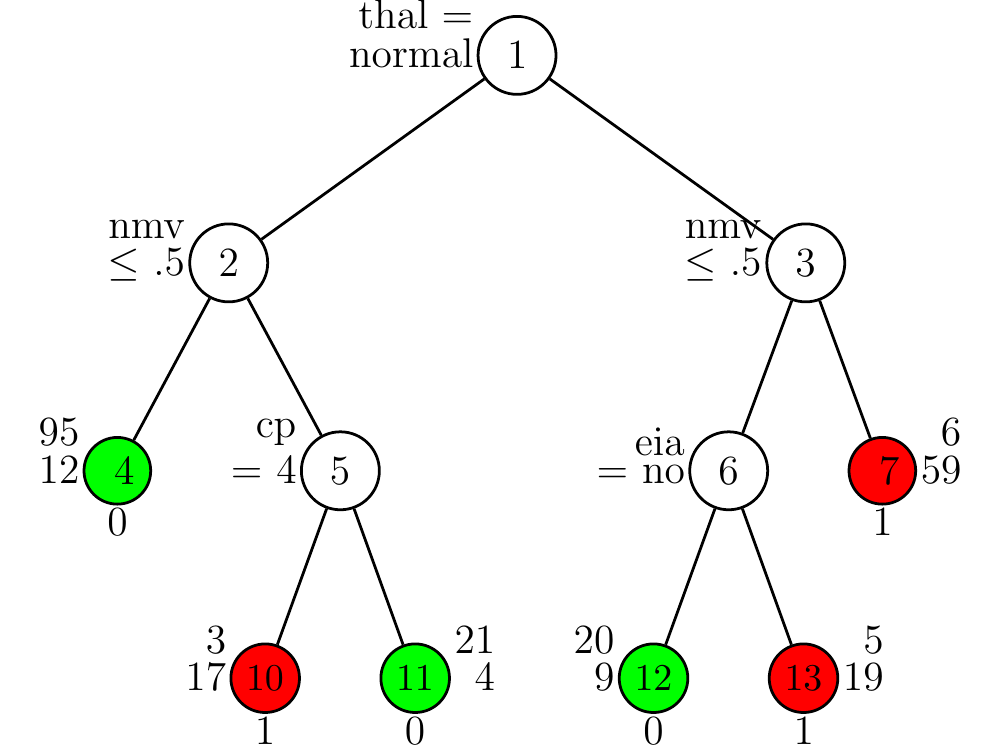}
\caption[GUIDE 0.2-SE classification tree for heart disease data.]{GUIDE 0.2-SE classification tree for heart disease data (Table \ref{T1:heart}).
At each intermediate node, an observation goes to the left branch if and only if the condition is satisfied. Predicted classes (based on estimated misclassification cost) given below terminal nodes; sample sizes for outcome 0, and 1, respectively, given beside nodes. Outcome 1 indicates presence of heart disease; 0 otherwise.}
\label{P1:guide}
\end{figure}

The GUIDE algorithm \citep{guide} was then proposed and it was designed to solve the selection bias problem of CART. GUIDE stands for ``Generalized, Unbiased Interaction Detection and Estimation". It controls selection bias by separating the split variable selection and split point selection procedures. At each node, a GUIDE regression tree obtains the residuals from fitting a constant or linear model. Then it constructs contingency tables by cross-tabulating the signs of the residuals and the categories or discretized groups of the split variable candidates. It employs the chi-squared test of association for the two-way contingency table as a criterion and selects the one with the smallest $p$-value to split the node. In addition, GUIDE applies bootstrap calibration to further correct selection bias. Over the years, the GUIDE algorithm has evolved gradually to accommodate growing demands of data analysis. The GUIDE program is well maintained and constantly updated. Figure \ref{P1:guide} is an example of GUIDE classification tree for the heart disease data (Table \ref{T1:heart}).

\subsection{Logistic regression trees}
Classic tree-based models were mostly designed to solve two sets of problems: classification and ordinary regression. For data with dichotomous response, classification tree models can be applied to build a binary classifier such as the GUIDE 0.2-SE tree shown in Figure \ref{P1:guide}, but they do not provide good estimates of $p$, while ordinary regression tree models may predict a probability of success out of the range $[0,1]$. Therefore, it is a natural idea to combine the tree structure and logistic regression, resulting in logistic regression tree models.

One obvious approach is to grow a classification tree first and run logistic regression in the terminal nodes. \citet*{Steinberg:Cardell:1998} implemented this idea using CART and found it unsuccessful. They stated that, ``by the time CART has declared a node terminal, the information remaining in the nodes is insufficient to support further statistical analysis." Meanwhile, they developed the hybrid CART-Logit model. Instead of fitting a logistic regression model in each terminal node, they constructed dummy variables as terminal node identifiers which contain the CART tree partitioning information. Denote the dummy variables by $U_i$, $i = 1,\ldots,|\tilde{\mathcal{T}}|$, where $|\tilde{\mathcal{T}}|$ is the number of terminal nodes. The hybrid CART-Logit model solves a logistic regression with the following mixed linear predictor:
\begin{equation}\label{equ:hybrid}
\eta^* = \beta_0 + \beta_1 X_1 + \ldots + \beta_K X_K + \beta_{K+1} U_1 + \ldots + \beta_{K+|\tilde{\mathcal{T}}|} U_{|\tilde{\mathcal{T}}|}.
\end{equation}
One drawback of this algorithm is that the final fitted model is hard to interpret. Also, there may exist complex collinearity among the dummy variables $U_i$ and the predictors $X_j$. Furthermore, this approach inherits the selection bias of CART, which may lead to inaccurate information in terminal node identifiers.

LOTUS, which stands for ``LOgistic Tree with Unbiased Selection", was developed by \citet*{lotus}. It is an analogue to the GUIDE piecewise linear regression tree for binary response data. Like GUIDE, LOTUS performs split variable selection and split point selection separately at each node. Due to the nature of logistic regression, its residuals $y_i-\hat{p}_i$ do not fluctuate around 0. In fact, the signs of the residuals are determined by the response $Y$. Therefore, the ``curvature tests" of GUIDE cannot be applied to logistic regression tree models. Instead, LOTUS employs a trend-adjusted chi-square test, which allows for linear effects in the model, to obtain the significance probabilities. Furthermore, LOTUS allows users the option to fit either multiple or simple linear logistic node models. However, the trend-adjusted chi-square test of LOTUS is not as comprehensive as the ``curvature tests" of GUIDE. Second, LOTUS is not able to handle high-dimensional data well when the multiple linear logistic regression option is selected, due to the lack of an advanced variable selection method. Furthermore, when a complete separation occurs, that is, when the response variable completely separates a predictor variable or a combination of predictor variables, LOTUS crashes.

Another work on combining a tree model and logistic regression is the logistic model tree (LMT) algorithm \citep{Landwehr:2005:LMT}. LMT grows a single tree containing binary splits on numeric predictors, and multiway splits on nominal ones, which follows the way trees are built by the M5 \citep{Quinlan:M5} algorithm. At each node, it employs the LogitBoost algorithm \citep{friedman00} for logistic regression model estimation as well as variable selection. Finally, it uses the CART pruning method to determine the tree. In practice, the LMT algorithm can be computationally intensive due to the slow LogitBoost procedure.

\citet*{MOB08} introduced the Model-Based Recursive Partitioning (MOB) algorithm, which is a
general framework for applying the tree structure to various types of parametric models,
including logistic regression. Similar to GUIDE and LOTUS, MOB separates split variable selection
from split point selection. First, it fits a parametric model to the data. Then it applies the
generalized M-fluctuation tests \citep{MOB07} to evaluate the stability of the fitted model
corresponding to every possible split variable, and chooses the one that is associated with
the highest instability as the split variable. The MOB algorithm does not perform pruning.
Therefore, the tree it provides may not have the best size.

%%%%%%%%%%%%%%%%%%%%%%%%%%%%%%%%%%%%%%%%%%%%%%%%%%%%%%%%%%%%%%%%%%%%%%%%%%%%%%%%%%%%%%%%%%%%%%%%%
\section{The PLUTO Algorithm}
In this section, we present a new algorithm for growing logistic regression trees with unbiased splits, called PLUTO: \emph{P}enalized, \emph{L}ogistic regression, \emph{U}nbiased splitting, \emph{T}ree \emph{O}perator.

\begin{figure}[!ht]
                \centering
                \includegraphics[width=\textwidth]{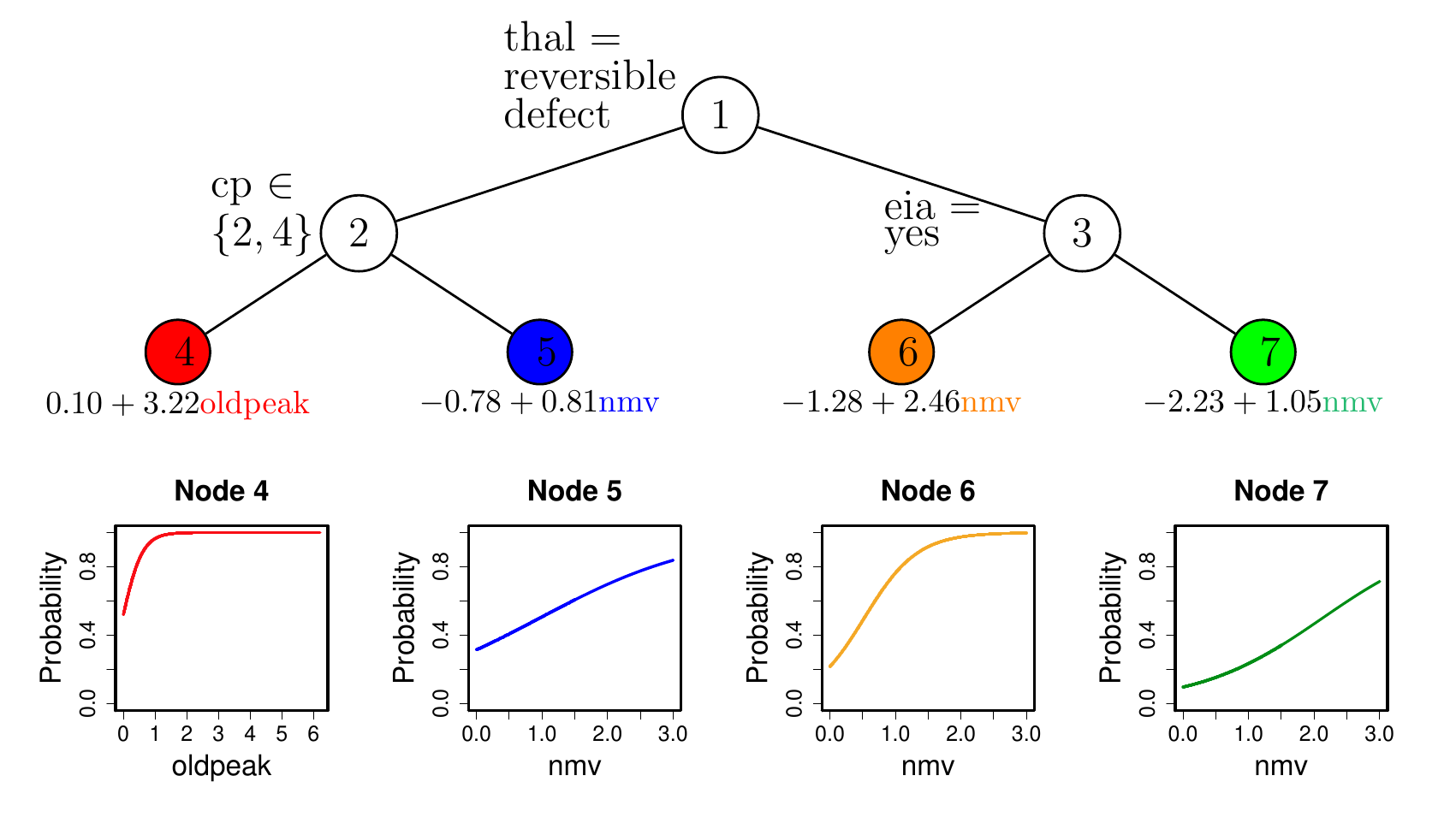}
\caption[PLUTO 0-SE simple linear logistic regression tree for heart disease data.]{PLUTO 0-SE simple linear logistic regression tree for heart disease data. At each intermediate node, an observation goes to the left branch if and only if the condition is satisfied. The fitted simple linear logistic regression lines are given below the terminal nodes and visualized as s-curves.}
\label{P2:pluto}
\end{figure}

To demonstrate the PLUTO tree, we apply our algorithm to the heart disease dataset (Table \ref{T1:heart}), with simple linear logistic regression models fitted in the nodes. The resulting tree is shown in Figrure \ref{P2:pluto}. We can see that, as a combination of recursive partitioning and logistic regression, PLUTO automatically captures the nonlinear and interaction patterns in the data like other tree-based models, and provides a smooth estimate of the probability of ``success" in each node as logistic regression. It also gives a vivid graphical display of the data that is rather interpretable.

The rest of this section shows the steps of growing a PLUTO tree in detail.

\subsection{Preparatory work}
Before we start growing the tree, we need to first define the roles of predictors and the type of node models to be fitted. Each numerical predictor can serve one, both or none of the following two roles: splitting and fitting. Each categorical predictor can be either included as a split variable candidate or be excluded. PLUTO does not include categorical predictors for fitting because the main and interaction effects of categorical predictors can be captured as splits. We refer to the set of split variable candidates as $\{X_1,\ldots,X_L\}$, among which there are $a$ categorical variables and $b$ numerical variables. The set of regressor candidates is denoted by $\{X_{N_1},\ldots,X_{N_K}\}$.

At each node, we can choose to fit a best simple linear logistic regression model or a multiple linear logistic regression model with all the regressor candidates included. The ordinary multiple model which applies no variable selection works on small datasets. However, with a large number of predictors, we often want to determine a smaller subset that exhibits strong effects on the response. Moreover, as the datasets grow even wider--i.e. many more features than samples, we won't be able to estimate all the parameters in ordinary multiple models. Hence, PLUTO employs regularized multiple linear logistic regression. Regularized logistic regression was introduced in Section \ref{sec:bg:glmnet}.

A best simple linear logistic regression model is a model that contains only one best numerical regressor. We use \emph{Deviance} as a criterion to search for the best regressor. This procedure is described in Algorithm \ref{A2:alg1}.

\begin{algorithm}
\caption{\strut Best Simple Linear Logistic Regression Model Selection}
\label{A2:alg1}
%\begin{small}
%Suppose $X_1,\ldots,X_K$ are the numerical variables that can be used for fitting. The following steps are carried out at each node.
\begin{algorithmic}[1]
%\hline
\STATE If the node data has pure response values, keep it as a pure classification node (one kind of terminal node). Stop.
\FOR{$k = 1$ to $K$}
\STATE Fit the simple linear model $\log [p/(1-p)]= \beta_0+\beta_1 X_{N_k}$. Let $\mathcal{D}_k$ denote its deviance (Degrees of freedom is defined as the number of fitted observations minus the number of estimated parameters, which do not vary in this case.).
\STATE If the model does not converge, define $\mathcal{D}_k = \infty$.
\ENDFOR
\STATE Let $k^* = \mbox{argmin}\ \mathcal{D}_k$.
\STATE If $\mathcal{D}_{k^*} = \infty$, delete the node and its sibling then turn its parent into a terminal node. Otherwise, select the simple linear logistic regression model with predictor $X_{N_{k^*}}$.
\end{algorithmic}
%\end{small}
\end{algorithm}

\subsection{Split variable selection}
\begin{figure}[H]
                \centering
                \includegraphics[width=0.45\textwidth]{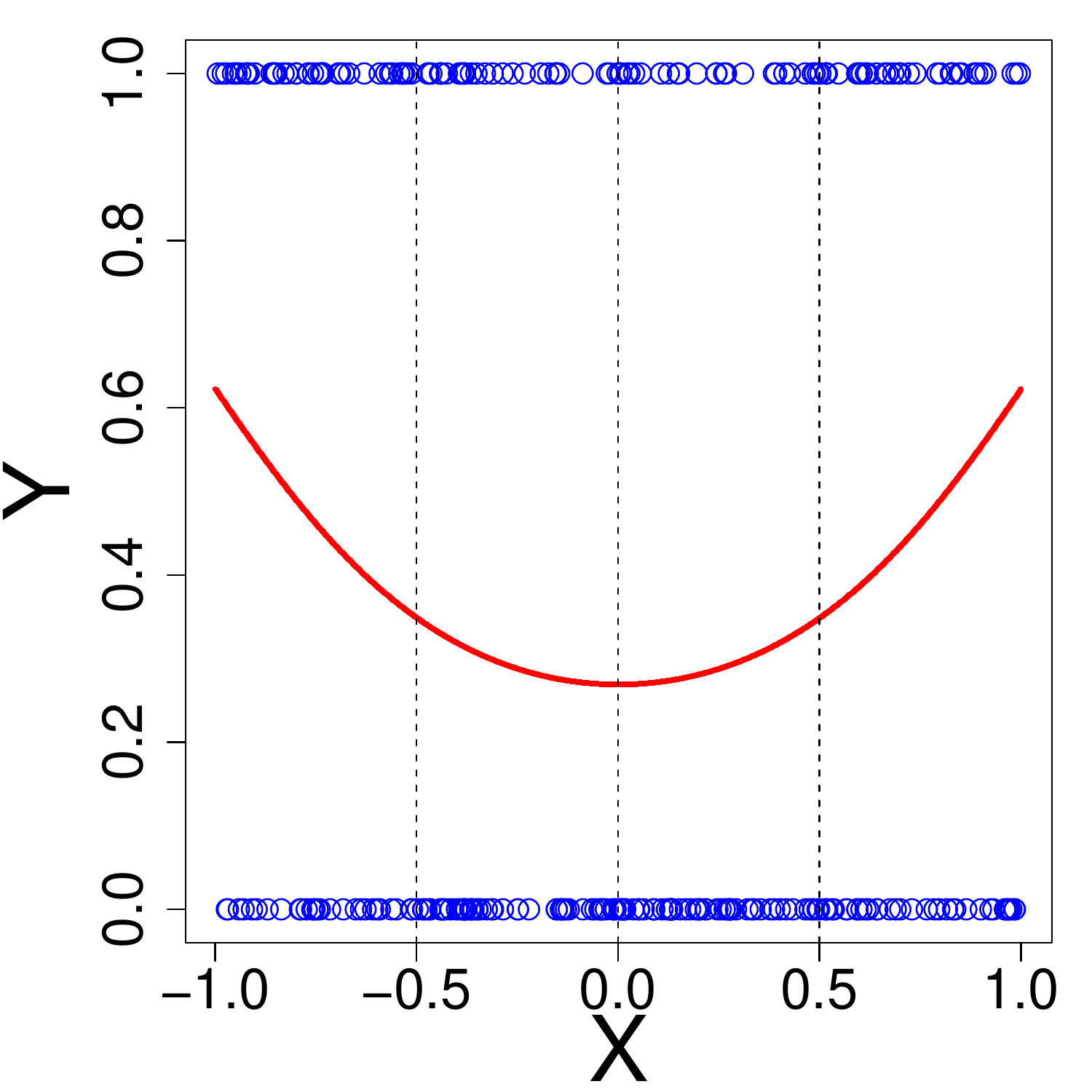}
                \includegraphics[width=0.45\textwidth]{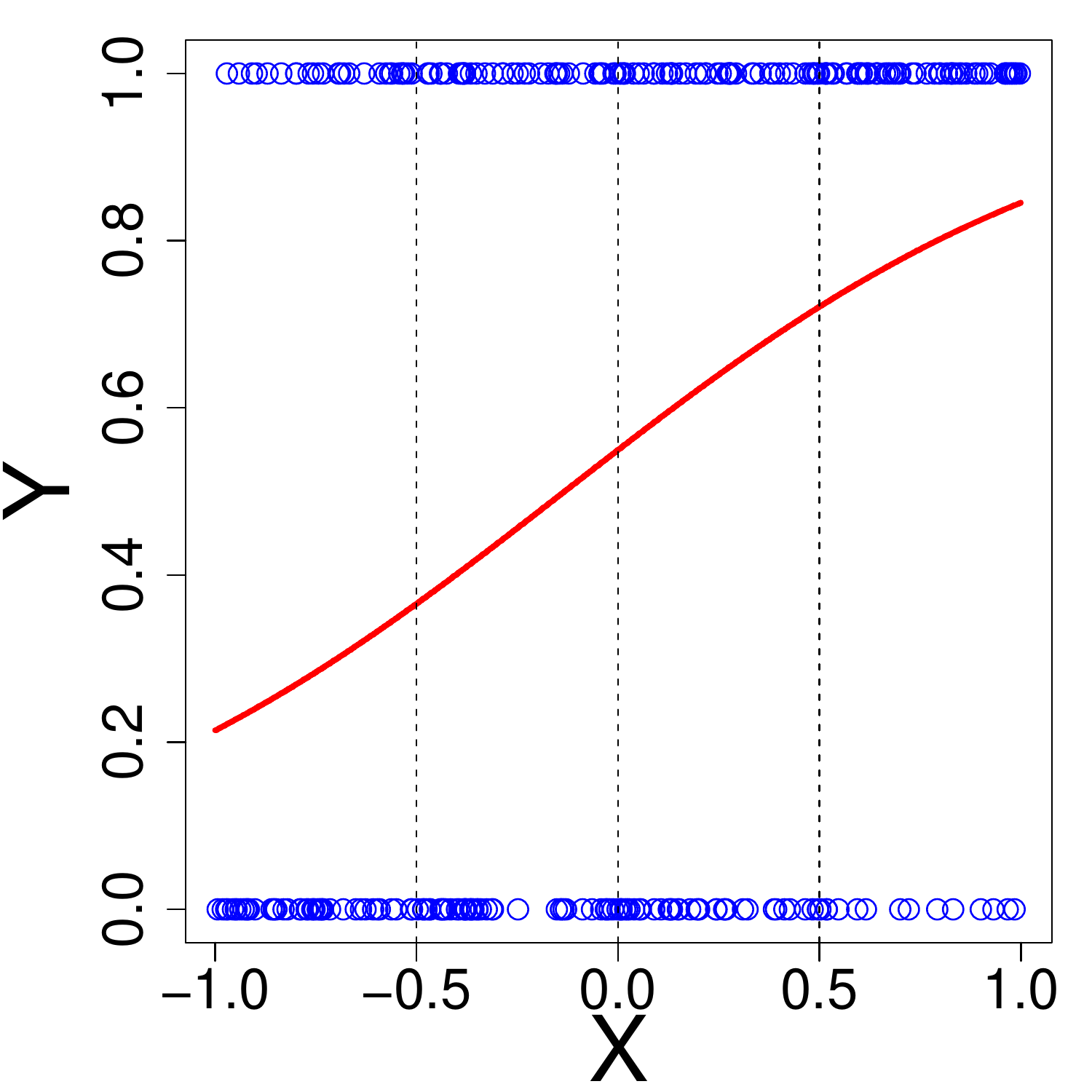}
\caption[Simulated data of which the predictors $X$ have nonlinear (Left) and linear (Right) effects on the response.]{Simulated data of which the predictors $X$ have nonlinear (Left) and linear (Right) effects on the response. Simulation models are shown as the solid red line: $\mbox{logit}(p) = -1+1.5X^2$ (Left); $\mbox{logit}(p) = 0.2+1.5X$ (Right).}
\label{P2:var:select}
\end{figure}

A crucial element of tree-based models is the splitting rule. Inspired by the GUIDE split variable selection algorithm, we propose to search for the split variable using chi-squared tests.
However, applying the ordinary Pearson chi-squared test is not satisfactory. Figure \ref{P2:var:select} shows a simple simulation study. The predictor $X$ has a nonlinear effect on the response for the dataset on the left, while it has a linear effect on the response for the dataset on the right.

We divide the values of $X$ into intervals, construct contingency tables, and conduct Pearson chi-squared tests for both datasets. The results are given in Table \ref{T2:chi}. We find that the Pearson chi-squared test not only detects nonlinear dependence, but also identifies strong linear dependence. This approach may select a linear predictor for splitting, resulting in a tree with misleading interpretation.

\begin{table}[!htb]
%\begin{center}
\caption{Contingency tables and chi-squared tests for simulated data shown in Figure
\ref{P2:var:select}.}\label{T2:chi}
\begin{adjustwidth}{-0.5cm}{}
\begin{scriptsize}
%\begin{footnotesize}
\begin{tabular}{ccccccccccc}
  & \multicolumn{4}{c}{Left} && & \multicolumn{4}{c}{Right}\\
  \cline{1-5} \cline{7-11} \noalign{\smallskip}
  & \multicolumn{4}{c}{$X$} && & \multicolumn{4}{c}{$X$}\\
  \cline{2-5} \cline{8-11} \noalign{\smallskip}
  & $[-1.0,-0.5)$ & $[-0.5,0.0)$ & $[0.0,0.5)$ & $[0.5,1.0]$ &  & & $[-1.0,-0.5)$ & $[-0.5,0.0)$ & $[0.0,0.5)$ & $[0.5,1.0]$\\
  \cline{1-5} \cline{7-11} \noalign{\smallskip}
  $Y=1$ & 42 & 28 & 24 & 32 && $Y=1$ &  27 & 39 & 46 &54\\
  $Y=0$ & 28 & 53 & 57 & 36 &&$Y=0$ &  43 & 42 & 35 &14\\
  \cline{1-5} \cline{7-11}  \noalign{\smallskip}
  \multicolumn{5}{c}{$\chi^2_3 = 16.950$, $p = 0.0007$} &&  \multicolumn{5}{c}{$\chi^2_3 = 25.670$, $p < 0.0001$} \\
  \cline{1-5} \cline{7-11}
\end{tabular}
\end{scriptsize}
%\end{footnotesize}
\end{adjustwidth}
%\end{center}
\end{table}

To solve this problem, we develop split variable selection algorithms that take into consideration the model fitting while splitting a node, which are presented next. Depending on the type of node model to be fit, a different split variable selection algorithm will be applied.

\subsubsection{Simple linear logistic regression option}
We use Algorithm \ref{A2:alg1} to select an $X$ variable to fit a simple linear logistic regression model at each node. Once the best regressor is determined, we conduct an adjusted chi-squared test on each split variable candidate to choose the one that shows most lack of fit. For a categorical split candidate $X_c$ with $C$ distinct values, one could regard the data as a $2\times C$ contingency table, in which the two cells in each column give the counts of the number of successes ($Y = 1$) and the number of failures ($Y = 0$), for that category. We then estimate the expected counts in each cell using the fitted value of model $\log [p/(1-p)] = \beta_0+\beta_1 X_{N_{k^*}}$, where $X_{N_{k^*}}$ is the best regressor as defined in Algorithm \ref{A2:alg1}. For instance, let the fitted probabilities of ``success" for observations falling into column/category 1 be a vector $\hat{\boldsymbol{\pi}}^{1} = (\hat{\pi}_1^{1},\ldots,\hat{\pi}_{n_{1}}^{1})$, where $n_{1}$ is the number of observations in column 1. Then the expected count of cell (1,1) is $\sum_{i = 1}^{n_{1}} \hat{\pi}_i^{1}$, and the expected count of cell (0,1) is $\sum_{i = 1}^{n_{1}} (1-\hat{\pi}_i^{1})$. Thus we can get the chi-squared statistic via:
\begin{equation}\label{equ:3}
\chi^2 = \sum \frac{(\mbox{observed} - \mbox{expected})^2}{\mbox{expected}}.
\end{equation}
Since $X_c$ is not a predictor in the simple model, under $H_0: X_{N_{k^*}} \mbox{ is not lack of fit}$, the expected counts in each cell will be proportional, which we count as 1 parameter. Hence, the degrees of freedom of the above $\chi^2$ is $C-1$.

A numerical split candidate $X_n$ may take numerous distinct values. One could regard the data as a $2 \times D$ contingency table, where $D$ refers to the number of distinct values of $X_n$. The cell counts in this table can be small, as are the fitted counts. The large-sample theory for Pearson $\chi^2$ and likelihood-ratio $G^2$ test statistics applies when there is a fixed number of cells and when the fitted count in each cell is large. This theory is violated for the $2 \times D$ table in two ways. First, fitted counts are often found to be small. Second, when more data are collected, additional distinct values may occur, so the contingency table will contain more cells rather than a fixed number. Therefore, $\chi^2$ and $G^2$ for logistic regression with numerical predictors do not have approximate chi-squared distributions. To check the nonlinear pattern of a numerical variable adjusting for the best simple linear logistic model, we compare the observed and fitted values in grouped form. We divide the observed and fitted values into $M$ intervals using the $100/M,200/M,\ldots,100(M-1)/M$ sample percentiles of $X_n$ as cutoff points, and form a $2 \times M$ table. Then, we calculate the fitted values and chi-squared statistic the same way we did for a categorical split candidate. This Pearson-like statistic does not actually have a chi-squared distribution, but simulations show that its distribution is roughly approximated by chi-squared \citep[see][page 114]{agresti1996}. If $n \neq N_{k^*}$, similar as for categorial variable, $df = M-1$. Otherwise, if $n = N_{k^*}$, the simple logistic regression model has two parameters, so $df = M-2$.

For each split variable candidate, we calculate the $p$-value according to the chi-squared statistics and $df$. Then we choose the variable with strongest evidence of lack of fit (the one with smallest $p$-value) as the split variable. The whole procedure is summarized in Algorithm \ref{A2:alg2}.
\begin{algorithm}
\caption{\strut Split Variable Selection Algorithm I - Simple Linear Logistic Regression Option}
\label{A2:alg2}
%\begin{small}
%Refer to the set of split variable candidates as $\{X_1,\ldots,X_L\}$.
\begin{algorithmic}[1]
\STATE Run Algorithm \ref{A2:alg1} and denote the best simple linear logistic model by\\
$\log [p/(1-p)] = \beta_0+\beta_1 X_{N_{k^*}}$. Calculate and record the fitted probability for each observation.
\FOR{$l = 1$ to $L$}
\IF{$X_l$ is a categorical variable with $C$ categories}
\STATE Construct the data into a $2\times C$ contingency table by cross-tabulating the response $Y$ and $X_l$. Set $df = C-1$.
\ELSE
\STATE Group the data into $M$ categories using the $100/M,200/M,\ldots,100(M-1)/M$ sample percentiles of $X_l$ as cutoff points. Put the data into a $2\times M$ contingency table. If $l \neq N_{k^*}$, $df = M-1$. Otherwise, $df = M-2$.
\ENDIF
\STATE Let $n_{(r,c)}$ denote the observed count in cell $(r,c)$ of the contingency table, and $n_{c}=n_{(1,c)}+n_{(0,c)}$. $n_{c}$ is the total number of observations in column $c$. Also, let $\hat{\boldsymbol{\pi}}^{c} = (\hat{\pi}_1^{c},\ldots,\hat{\pi}_{n_{c}}^{c})$ denote the vector of fitted probabilities for observations that fall in column $c$. The expected count of cell $(1,c)$ is $\hat{n}_{(1,c)} = \sum_{i = 1}^{n_{c}} \hat{\pi}_i^{c}$, and the expected count of cell $(0,c)$ is $\hat{n}_{(0,c)} = n_{c}-\hat{n}_{(1,c)}$.
\STATE Let $\chi^2 = \sum_{r,c}{(n_{(r,c)}-\hat{n}_{(r,c)})^2}/{\hat{n}_{(r,c)}}$ and let $\rho_{l}$ denote its $p$-value.
\ENDFOR
\STATE Let $l^* = \mbox{argmin}\ \rho_l$. Select $X_{l^*}$ as the split variable.
\end{algorithmic}
%\end{small}
\end{algorithm}

%\clearpage
\subsubsection{Multiple Linear Logistic Regression Option}
PLUTO applies the GLMNET algorithm to fit a regularized multiple linear logistic regression model. The value of $\alpha$ for the elastic net penalty $P_{\alpha}$ is alterable, although the PLUTO default setting uses the lasso penalty. In each node, we tune the parameter $\lambda$ in (\ref{equ:4}) by $10$-fold cross-validation and select the one that minimizes the deviance. Then the fitted probability of each observation is estimated, and we carry out the adjusted chi-squared tests to search for the split variable candidate that shows most lack of fit. The procedure is summarized in Algorithm \ref{A2:alg3}.
\begin{algorithm}
\caption{\strut Split Variable Selection Algorithm II - Multiple Linear Logistic Regression Option}
\label{A2:alg3}
%\begin{small}
\begin{algorithmic}[1]
%\hline
\STATE If the data has pure response values, keep it as a pure classification node, and stop.
\STATE Select $\alpha \in [0,1]$, default $\alpha = 1$.
\STATE Run $10$-fold cross-validation of logistic regression model with elastic-net penalty $P_{\alpha}$ using the GLMNET algorithm. Select tuning parameter and denote it as $\lambda^*$. Calculate and record the fitted probabilities corresponding to $\lambda^*$.
\STATE If the model does not converge, delete the node and its sibling, turn its parent into a terminal node, and stop.
\FOR{$l = 1$ to $L$}
\IF{$X_l$ is a categorical variable with $C$ categories}
\STATE Put the data into a $2\times C$ contingency table. Set $df = C-1$.
\ELSE
\STATE Put the data into a $2\times M$ contingency table. Set $df = M-1$.
\ENDIF
\STATE Same as Algorithm \ref{A2:alg2}, let $n_{(r,c)}$ denote the observed count in cell $(r,c)$, $n_{c}$ denote the total number of observations in column $c$, and $\hat{\boldsymbol{\pi}}^{c}$ denote the vector of fitted probabilities for observations that fall in column $c$. Calculate the expected counts $\hat{n}_{(1,c)} = \sum_{i = 1}^{n_{c}} \hat{\pi}_i^{c}$ and $\hat{n}_{(0,c)} = n_{c}-\hat{n}_{(1,c)}$.
\STATE Let $\chi^2 = \sum_{r,c}{(n_{(r,c)}-\hat{n}_{(r,c)})^2}/{\hat{n}_{(r,c)}}$ and let $\rho_{l}$ denote its $p$-value.
\ENDFOR
\STATE Let $l^* = \mbox{argmin}\ \rho_l$. Select $X_{l^*}$ as the split variable.
\end{algorithmic}
%\end{small}
\end{algorithm}

\subsection{Split point/subset selection}
In each node, after we select the best split variable $X_{l^*}$ using a simple or multiple linear logistic regression model, we need to find the exact split criterion to partition the sample space. If $X_{l^*}$ is a numerical predictor, we search for the split point $\delta$ that forms the split criterion $X_{l^*} \leq \delta$. The candidates of $\delta$ will be any values that $X_{l^*}$ can take, except the minimum value. Ideally, we want to search over all these candidates for the one that minimizes the total deviance of the logistic regression models fitted to the two sub-datasets defined by the split. However, this approach is too compute-intensive when $X_{l^*}$ takes a large number of distinct values. Instead of exhaustive search, PLUTO chooses a set of split point candidates based on the percentiles of $X_{l^*}$ and search for the one that minimizes the sum of model deviances in the two sub-datasets. The PLUTO default setting uses the 0.2, 0.4, 0.6, and 0.8 percentiles of $X_{l^*}$ as split point candidates.

If $X_{l^*}$ is a categorical variable with values $\mathcal{V} = \{\nu_1,\ldots,\nu_C \}$, we want to identify a subset $A$ of $\mathcal{V}$ for a split criterion in the form of $X_{l^*} \in A$. When X is ordinal, we keep the natural order of its values. That is, we search over split subsets $A_i = \{\nu_1,\ldots,\nu_i\}, i = 1,\ldots,C-1$ and select the one that minimizes the sum of model deviances in the two sub-datasets. For a nominal split variable, the greedy algorithm searches over all $2^{C-1}-1$ possible split. When $C$ is large, it is computationally expensive. CART carries out exhaustive search for subsets of categorical split variables but as a result, it cannot deal with variables that have more than 32 classes due to the cost of computation. PLUTO sorts the values of $X_{l^*}$ manually in order of the proportions of ``success" among cases that fall into each category of $X_{l^*}$. Let $\{\nu'_1,\ldots,\nu'_C\}$ denote the ordered values. We search over split subsets $A'_i = \{\nu'_1,\ldots,\nu'_i\}, i = 1,\ldots,C-1$ for the best. This procedure is introduced in \citet[][page 101]{cart}.

When one or more sub-datasets have pure response, i.e. $Y = 0$ or $Y = 1$ for all the observations, the parameter estimates will fail to converge. PLUTO provides users the options to (1) keep this split, and set the deviance of the sub-dataset with pure response to 0; or (2) ignore this split, and search over the rest split points/subsets for the best one. The whole procedure of split point/subset selection is summarized in Algorithm \ref{A2:alg4}.
\begin{algorithm}
\caption{\strut Split Point/Subset Selection}
\label{A2:alg4}
%\begin{small}
\begin{algorithmic}[1]
%\hline
\IF{the best split variable $X_{l^*}$ is a numerical variable}
\FOR{$i = 1$ to $M-1$}
\STATE Set split point $\delta_i = 100 \times i/M \mbox{ sample percentile}$. Split the dateset using criterion $X_{l^*} \leq \delta_i$.
\STATE In each of the two sub-datasets, search for the best simple linear logistic regression
model using Algorithm \ref{A2:alg1} (Simple linear option) or fit a regularized multiple linear
logistic model using GLMNET (Multiple linear option). Calculate the deviances for both models
and denote their sum as $\mathcal{D}_i$.
%If the response is pure in any of the sub-dataset, depending on user's choice, we set the deviance to 0 or eliminate this split point candidate.
\ENDFOR
\ELSE
\IF{$X_{l^*}$ is an ordinal variable with ordered values $\mathcal{V} = \{\nu_1,\ldots,\nu_C \}$}
\FOR{$i = 1$ to $C-1$}
\STATE Set split setset $A_i = \{\nu_1,\ldots,\nu_i \}$. Split the dataset using criterion $X_{l^*} \in A_i$. Run Step 4.
\ENDFOR
\ELSE
\STATE $X_{l^*}$ is a nominal variable with values $\mathcal{V} = \{\nu_1,\ldots,\nu_C \}$. Let $Q_j$ denote the proportion of cases that have $Y = 1$ among the observations with $X_{l^*} = \nu_j$, $j = 1,\ldots,C$. Sort the value of $X_{l^*}$ in ascending order of $Q_j$ and denote the ordered values as $\mathcal{V}' = \{\nu'_1,\ldots,\nu'_C \}$.
\FOR{$i = 1$ to $C-1$}
\STATE Set split setset $A'_i = \{\nu'_1,\ldots,\nu'_i \}$. Split the dataset using criterion $X_{l^*} \in A'_i$. Run Step 4.
\ENDFOR
\ENDIF
\ENDIF
\STATE Let $i^* = \mbox{argmin}\ \mathcal{D}_i$. Select the corresponding $\delta_{i^*}$, $A_{i^*}$ or $A'_{i^*}$ as split point/subst of $X_{l^*}$.
\end{algorithmic}
%\end{small}
\end{algorithm}

\subsection{Determining tree size}
By iteratively applying the split variable selection algorithm and split point/subset selection algorithm, we grow a large PLUTO tree until we do not have a sufficient number of observations in the node or until the tree has grown up to a maximum depth. The largest tree usually overfits in most cases. To determine the right tree size, we adopt the minimal cost-complexity pruning algorithm of CART.

For any PLUTO tree $\mathcal{T}$, its complexity is defined as the total number of terminal nodes in the tree, denoted by $|\tilde{\mathcal{T}}|$. For PLUTO, we use the deviances of $\mathcal{T}$ as ``cost", and refer to it as $\mathcal{D}(\mathcal{T})$. The cost-complexity of PLUTO is then defined as:
\begin{equation}\label{equ:pru:1}
\mathcal{D}_{\kappa}(\mathcal{T}) = \mathcal{D}(\mathcal{T}) + \kappa|\tilde{\mathcal{T}}|.
\end{equation}
Here $\kappa$ is a pruning parameter. Similar to the tuning parameter $\lambda$ of LASSO, the greater $\kappa$ is, the tree that minimizes $\mathcal{D}_{\kappa}(\mathcal{T})$ will be smaller. Hence, we can prune back an overly grown tree $\mathcal{T}$ sequentially as $\kappa$ increases, until the tree consists of the root node only. By doing so, we obtain a sequence of nested subtrees of $\mathcal{T}$ and a sequence of the corresponding $\kappa$ values.

Ten-fold cross-validation is then employed to estimate the cost of each subtree in the sequence. In each iteration, the training dataset is used to grow a large PLUTO tree $\mathcal{T}^{\mbox{cv}}$. Using the sequence of $\kappa$ values obtained above, a sequence of nested subtrees of $\mathcal{T}^{\mbox{cv}}$ is computed by minimizing $\mathcal{D}_{\kappa}(\mathcal{T}^{\mbox{cv}})$. Each subtree is then used to predict the testing dataset. We refer to the response values of the testing dataset as $\{y_1,\ldots,y_n\}$ and its predicted probabilities of ``success" as $\{\hat{p}_1,\ldots,\hat{p}_n\}$.
The predicted deviance (DEV) is calculated by:
\begin{equation}
\label{E2:dev}
\mbox{DEV} = -2\sum_{i=1}^{n} [y_i\log{(\hat{p}_i)}+(1-y_i)\log{(1-\hat{p}_i)}].
\end{equation}

After the cross-validation, we obtain 10 DEV estimates for each subtree of $\mathcal{T}$.
The sample mean and sample standard deviation is calculated using these 10 estimates,
and we denote them by $\hat{\mathcal{D}}$ and $\mbox{SE}(\hat{\mathcal{D}})$ respectively.
The optimal tree is the one that has the smallest $\hat{\mathcal{D}}$ among the subtrees of $\mathcal{T}$, denoted by $\mathcal{T}^*$. We refer to the $\hat{\mathcal{D}}$ and $\mbox{SE}(\hat{\mathcal{D}})$ value of $\mathcal{T}^*$ as $\hat{\mathcal{D}}^*$ and $\mbox{SE}(\hat{\mathcal{D}}^*)$, respectively.

A $\theta\mbox{-SE}$ rule can be applied to select a final tree other than the optimal tree. The tree pruned by the $\theta\mbox{-SE}$ rule is denoted by $\mathcal{T}^*_\theta$, and it is the smallest subtree of $\mathcal{T}$ that satisfies:
\begin{equation}
\label{E2:pru:2}
\hat{\mathcal{D}}(\mathcal{T}^*_\theta) \leq \hat{\mathcal{D}}^* + \theta \mbox{SE}(\hat{\mathcal{D}}^*).
\end{equation}
We can see that the optimal tree $\mathcal{T}^*$ equals to $\mathcal{T}^*_0$. We show an example for the pruning process later in Section \ref{sec5} (Table \ref{T5:prune.rslts}).

\subsection{Importance ranking}
Often times, In data analysis, we are interested in determining a subset of variables that are more important than the rest. In other words, we want to rank the variables by their importance. However, ``importance" is a rather vague concept. We consider a variable that is more decisive in terms of predicting response variable using the PLUTO algorithm to be more ``important".
We adopt the importance ranking algorithm of Random Forest \citep{RandomForest} in PLUTO.
Giving a training dataset and a testing dataset, the following steps are performed to rank the
predictors variables:
\begin{enumerate}
\item Grow and prune a PLUTO tree using the training dataset.
\item Apply the final tree to the testing dataset and obtain a measure for the prediction accuracy, e.g., DEV.
\item For each variable: (i) replace its original values in the testing dataset by a bootstrap sample (permute the column with replacement); (ii)apply the final tree to the updated testing dataset and obtain an updated measure for the prediction accuracy.
\item Compare the updated measures with the original measure obtained on the un-permuted testing dataset.
\item The variables that have the updated measures further away from the original measure are more important.
\end{enumerate}

%%%%%%%%%%%%%%%%%%%%%%%%%%%%%%%%%%%%%%%%%%%%%%%%%%%%%%%%%%%%%%%%%%%%%%%%%%%%%%%%%%%%%%%%%%%%%%%%%
\section{Selection Bias} \label{sec3}
Controlling selection bias has fundamental importance for tree models. In this section, we discuss the selection bias in the PLUTO algorithms and a bias correction algorithm to remedy this problem. Simulation experiments are carried out to study the probabilities of variable selection and selection bias.

\subsection{Selection bias and bootstrap bias correction}
\citet{guide} presented simulation results where two predictors $X_1$ and $X_2$ allow $C_1$ and $C_2$ splits of the data, respectively, such that $C_1 \gg C_2$. Both predictors are independent of the response variable. CART selects $X_1$ with much higher probability over $X_2$. Due to the selection bias towards variables that allow more splits, a CART tree may choose less informative variables as split variables, leading to incorrect conclusions.

Our algorithm takes the GUIDE approach, separating split variable selection from split point/subset selection. By using adjusted chi-squared tests rather than exhaustive search like CART, we reduce the influence of variables that allow more splits. However, in the simulation study below, we find in Figures \ref{P3:SB1} and \ref{P3:SB3} that Algorithms \ref{A2:alg2} and \ref{A2:alg3} still show a preference for categorical split variables. This is because in our algorithm, only numerical variables can serve as regressors in node models. Therefore, the expected frequencies are correlated to the observed frequencies in the contingency tables of numerical variables, while categorical variables have $\chi^2$ values following the theoretical chi-squared distributions more closely. Another possible reason for the bias is that categorical variables and numerical variables possess different degrees of freedom.

PLUTO employs the GUIDE bootstrap calibration method to correct the selection bias towards categorical variables. In order to increase the chance of selecting numerical split variables, we want to reduce their corresponding $p$-values. However, when a $p$-value is close to 0, it is difficult to search for the shrinkage multiplier. We solve this problem by transforming each $p$-value into a $z$-value via $z = \Phi^{-1}(1-p/2)$, where $\Phi$ is the cumulative distribution function of the standard normal distribution. An outer loop is created to generate a new dataset for which the response variable is independent of the predictors using bootstrap sampling. We search over a grid of values of the multiplier, and choose the one that makes the frequency of selecting numerical split variables equal $b/L$, where $b$ is number of numerical variables among a total number of $L$ split variable candidates. The details of this procedure are given in Algorithm \ref{A3:alg1}.

\begin{algorithm}
\caption{\strut Bootstrap Calibration for Bias Correction}
\label{A3:alg1}
Let $\{X_1,\ldots,X_K,Y\}$ denote the matrix of training samples, where $Y = (y_1,\ldots,y_N)'$ is a vector of responses. Among the split variable candidates, let the numbers of categorical and numerical variables be $a$ and $b$ respectively.
\begin{algorithmic}[1]
%\hline
\FOR{$j = 1$ to $J$}
\STATE Draw a bootstrap sample $\{X_1,\ldots,X_K,Y^*\}$, where $Y^* = (y_1^*,\ldots,y_N^*)'$ and each $y_i^*$ is a random draw WITH replacement from the set $\{y_1,\ldots,y_N\}$.
\STATE Compute the $p$-values as described in Algorithm \ref{A2:alg2} or \ref{A2:alg3}.
\STATE Convert the $p$-values into $z$-values via the transformation $z = \Phi^{-1}(1-p/2)$.
\FOR{$\gamma = 1$ to $\Gamma$ by step $s$}
\STATE Let $z_n$ denote the largest $z$-value for the numerical variables and let $z_c$ denote the largest $z$-value for the categorical variables. Select the numerical variable if $\gamma z_n \ge z_c$.
\ENDFOR
\ENDFOR
\STATE For each $\gamma$, let $\pi(\gamma)$ denote the proportion of times that a numerical variable is chosen.
\STATE Linearly interpolate, if necessary, to find the value of $\gamma$ such that $\pi(\gamma) = b/(a+b)$, and denote it by $\gamma^*$.
\STATE Apply Algorithm \ref{A2:alg2} or \ref{A2:alg3}, and convert the $p$-values into $z$-values.
\STATE Adjust the $z$-values by multiplying those corresponding to the numerical variables by $\gamma^*$.
\STATE Select the variable with greatest adjusted $z$-value as the split variable.
\end{algorithmic}
%\end{small}
\end{algorithm}

\subsection{Simulation experiments}
In this subsection, we present simulation results of the selection probability and bias of both split variable selection Algorithms \ref{A2:alg2} and \ref{A2:alg3}. Then we assess the effect of our bootstrap bias correction Algorithm \ref{A3:alg1}. For each study, we generate five mutually independent predictor variables with their marginal distributions shown in Table \ref{T3:simu}. Predictors $X_1$ and $X_5$ both follow discrete uniform distributions with the same degree of discreteness. $X_3$ is symmetric while $X_2$ is skewed, and $X_4$ follows a bimodal distribution. In our simulation study, $X_5$ is used as a categorical variable (excluded from model fitting) while the others are numerical variables.

Next, we perform simulation experiments with random samples generated from the 9 models given in Table \ref{T3:simu}. The simulation results are based on 1000 iterations with 500 random samples in each iteration.
\begin{table}[!htb]
%\begin{center}
\begin{adjustwidth}{-0.25cm}{}
\caption{Variable and model descriptions for simulation experiments.}\label{T3:simu}
\begin{tabular}{cclcll}
\hline
\hline
Variable & Type\textsuperscript{\dag} & Distribution & & Model & logit($p$) \\
\cline{1-3}
\cline{5-6}
$X_1$ & N & Uniform\{-3,-1,1,3\} & & Null & 0 \\
$X_2$ & N & Exponential(1) & & Jump & $1+0.7I(X_1 > 0)$\\
$X_3$ & N & N(0,1) & & Int & $(0.5-0.5X_3)I(X_1 > 0)$ \\
$X_4$ & N & 0.5N(0,1)+0.5N(1,1) & & & $+(-0.5+0.5X_3)I(X_1 < 0)$ \\
$X_5$ & C & Discrete Uniform & & Quadratic & $1+0.08X_1^2$ \\
 & & \{-2,-1,1,2\} & & Cubic & $1+0.02X_1^3$ \\
 & &  & & Linear & $1+0.8X_2$ \\
 & &  & & LinQuad & $-1.5+X_2+X_3^2$ \\
 & &  & & LinLin & $-1+X_2+X_3$ \\
 & &  & & LinLinQuad & $1-0.1X_1^2+X_2+X_3$ \\
\hline
\multicolumn{6}{l}{\textsuperscript{\dag}\footnotesize{Variable type:``N"--numerical predictor, ``C"--categorical predictor.}}
\end{tabular}
\end{adjustwidth}
%\end{center}
\end{table}

\subsubsection{Selection probability of split variables and selection bias}
\begin{figure}[!ht]
%\begin{center}
\makebox[\textwidth][c]{\includegraphics[scale=0.5]{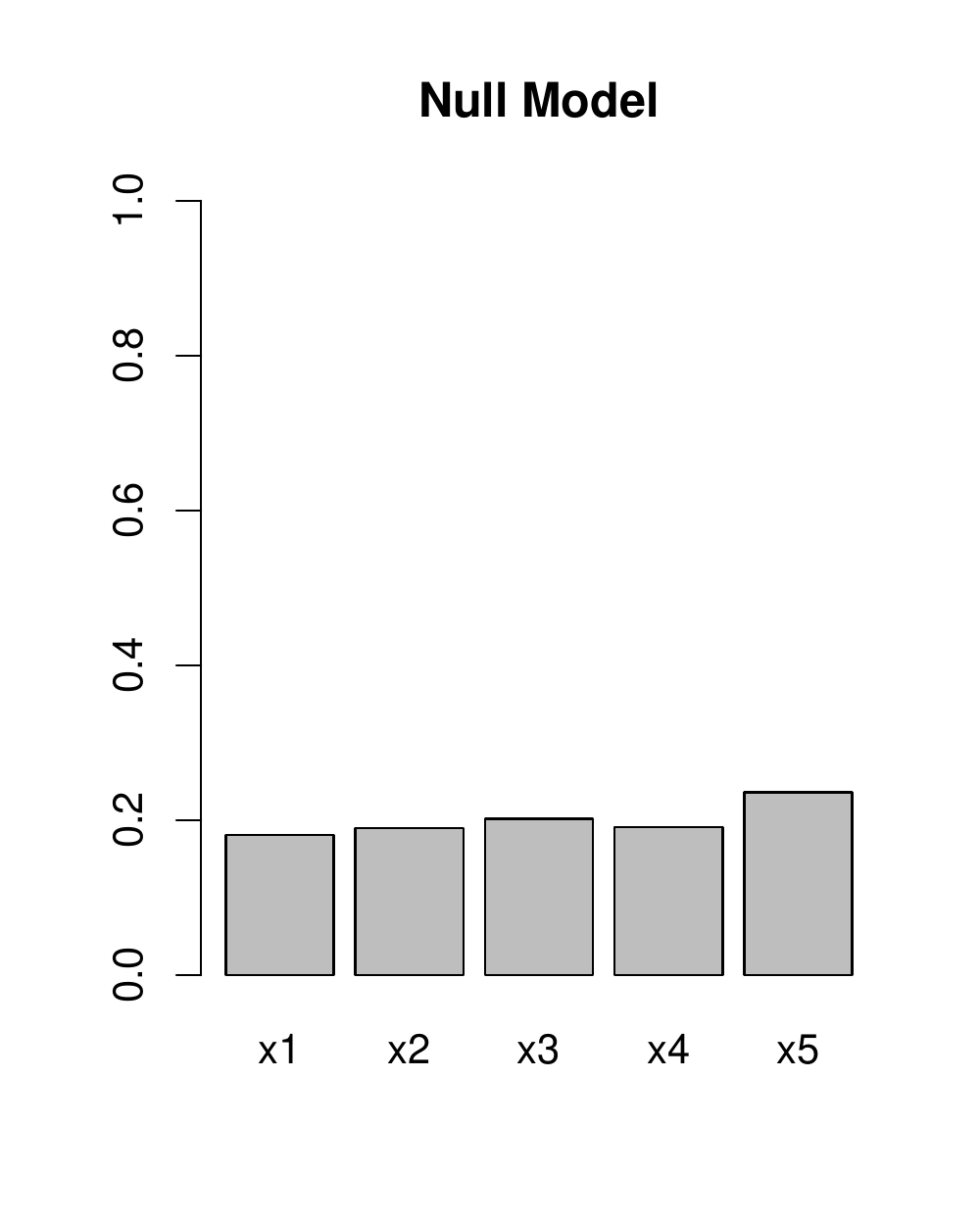}
\includegraphics[scale=0.5]{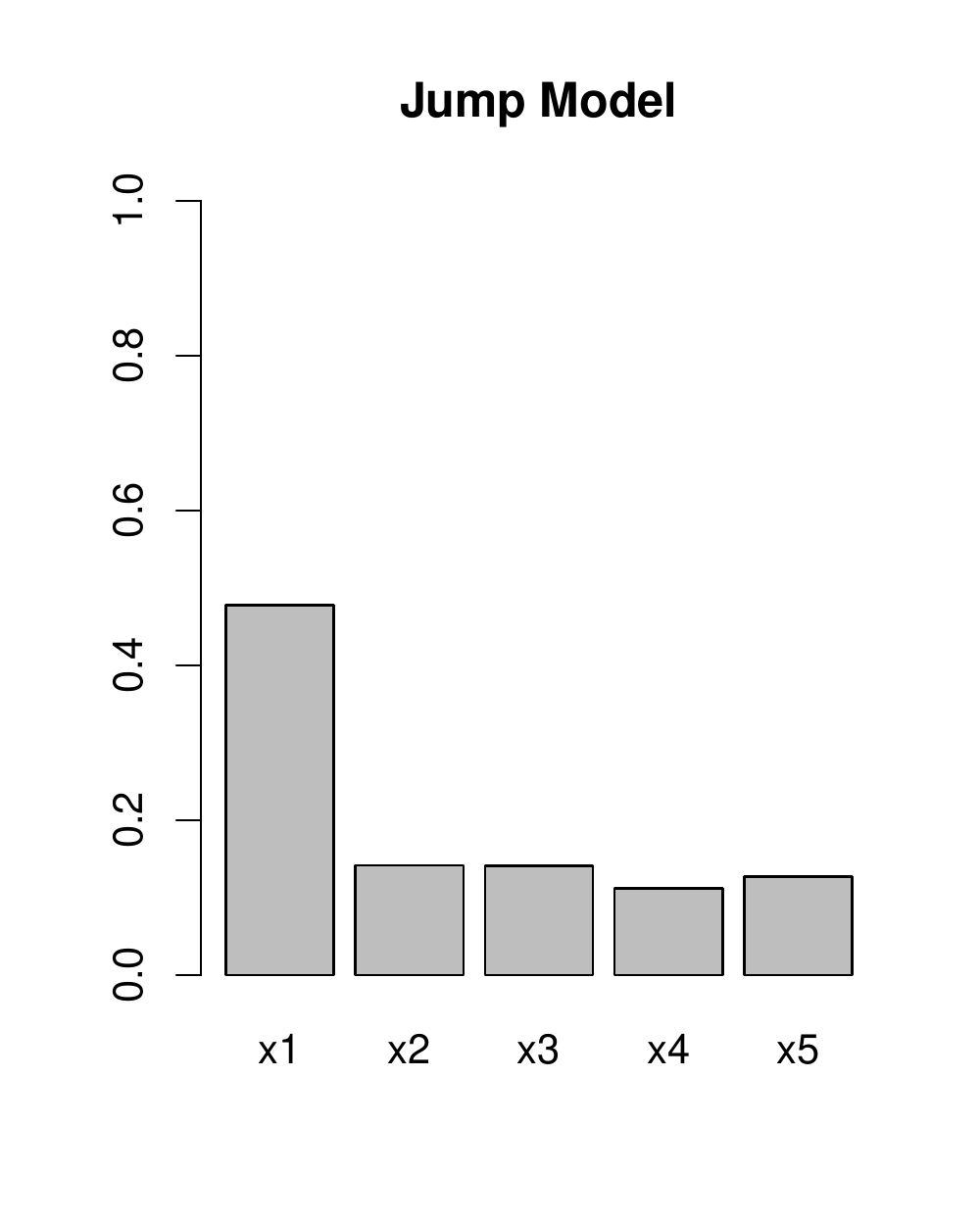}
\includegraphics[scale=0.5]{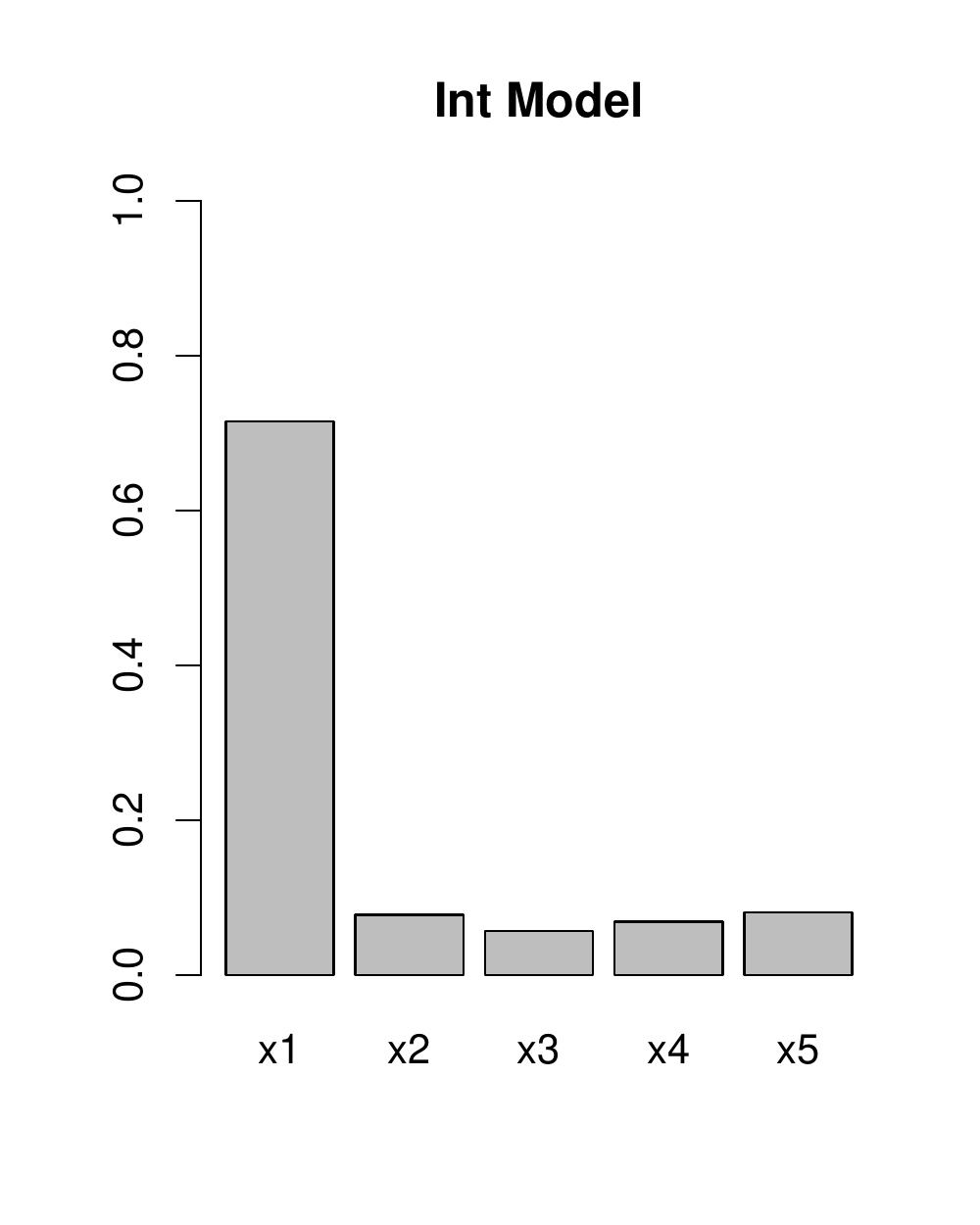}}
\makebox[\textwidth][c]{\includegraphics[scale=0.5]{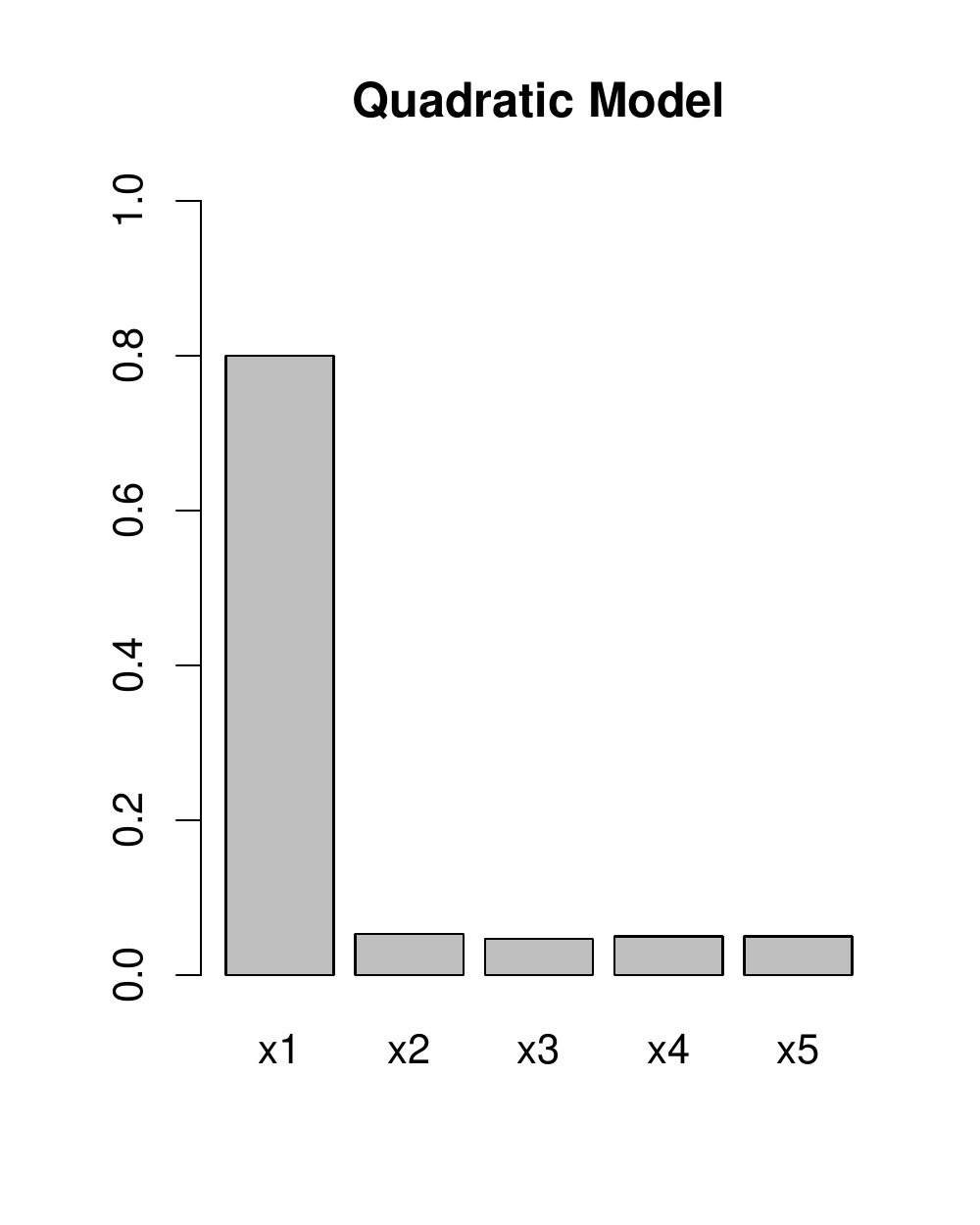}
\includegraphics[scale=0.5]{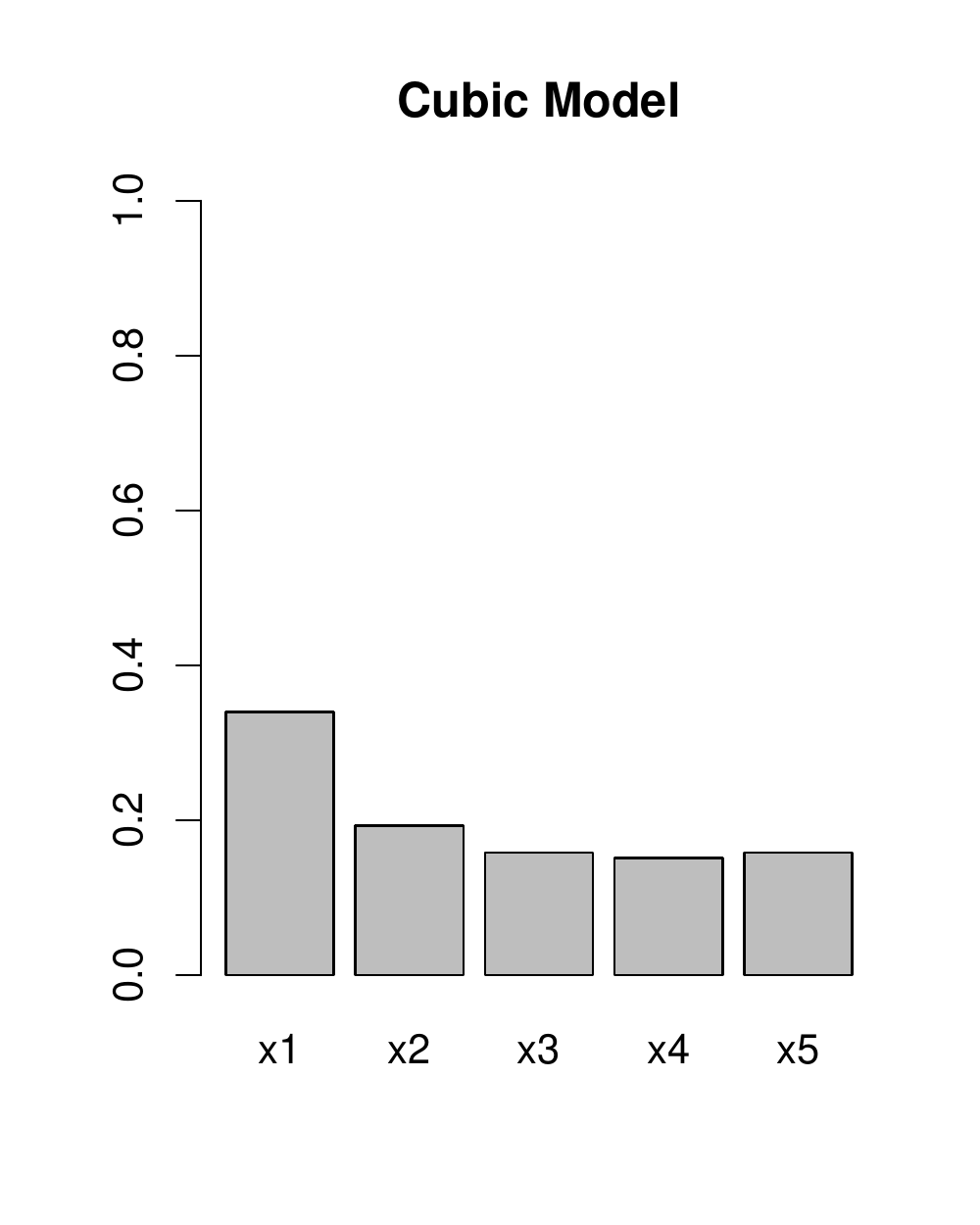}
\includegraphics[scale=0.5]{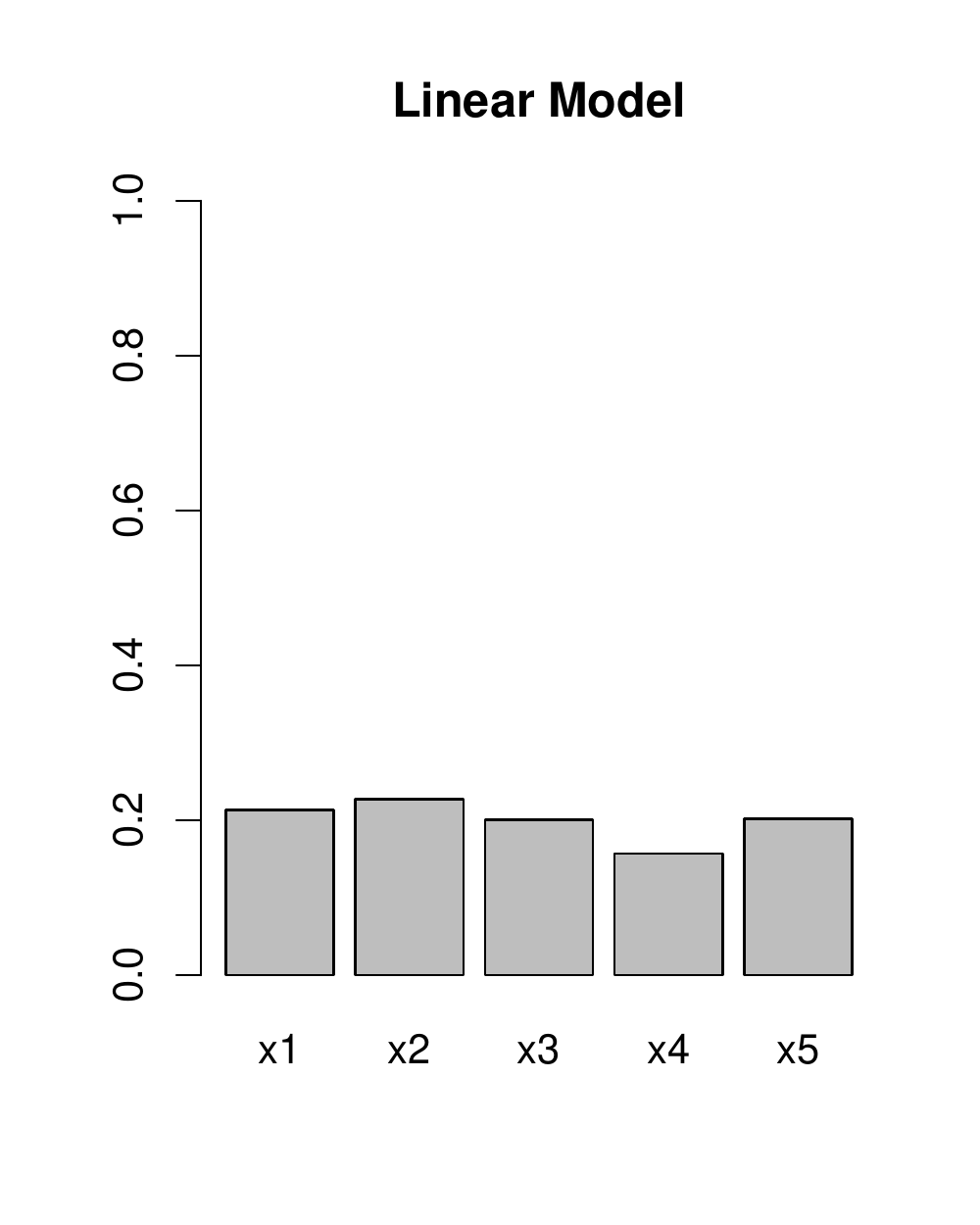}}
\makebox[\textwidth][c]{\includegraphics[scale=0.5]{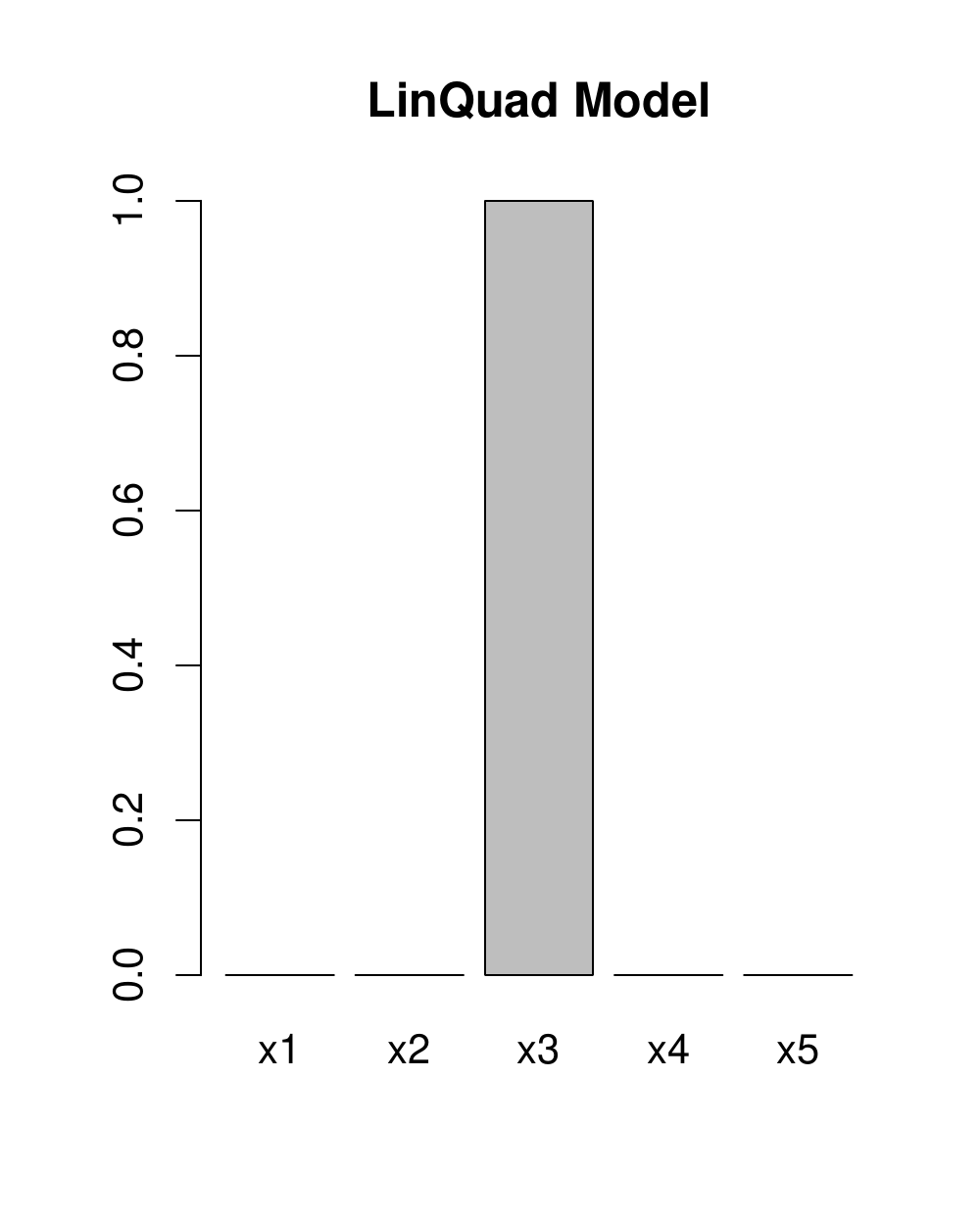}
\includegraphics[scale=0.5]{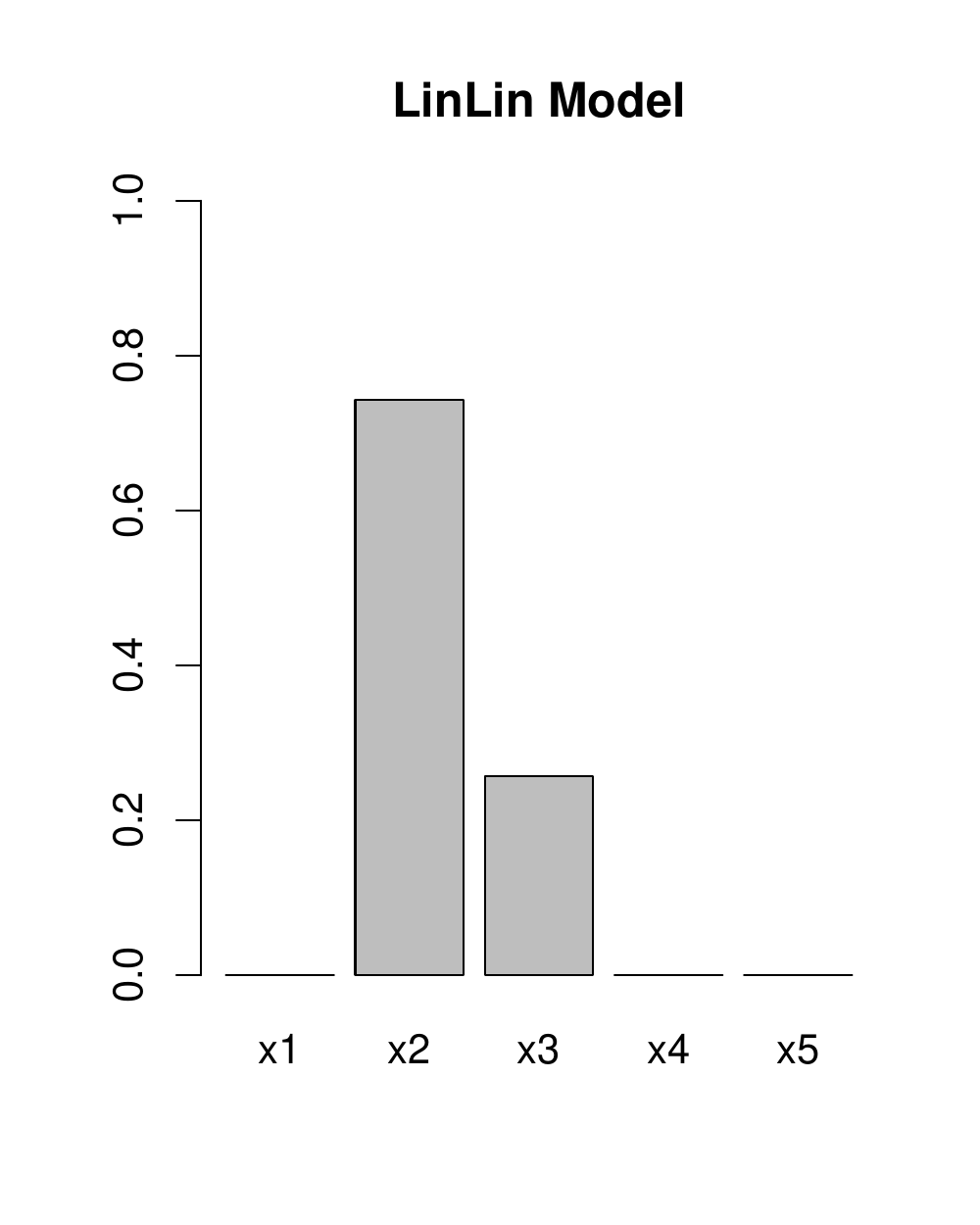}
\includegraphics[scale=0.5]{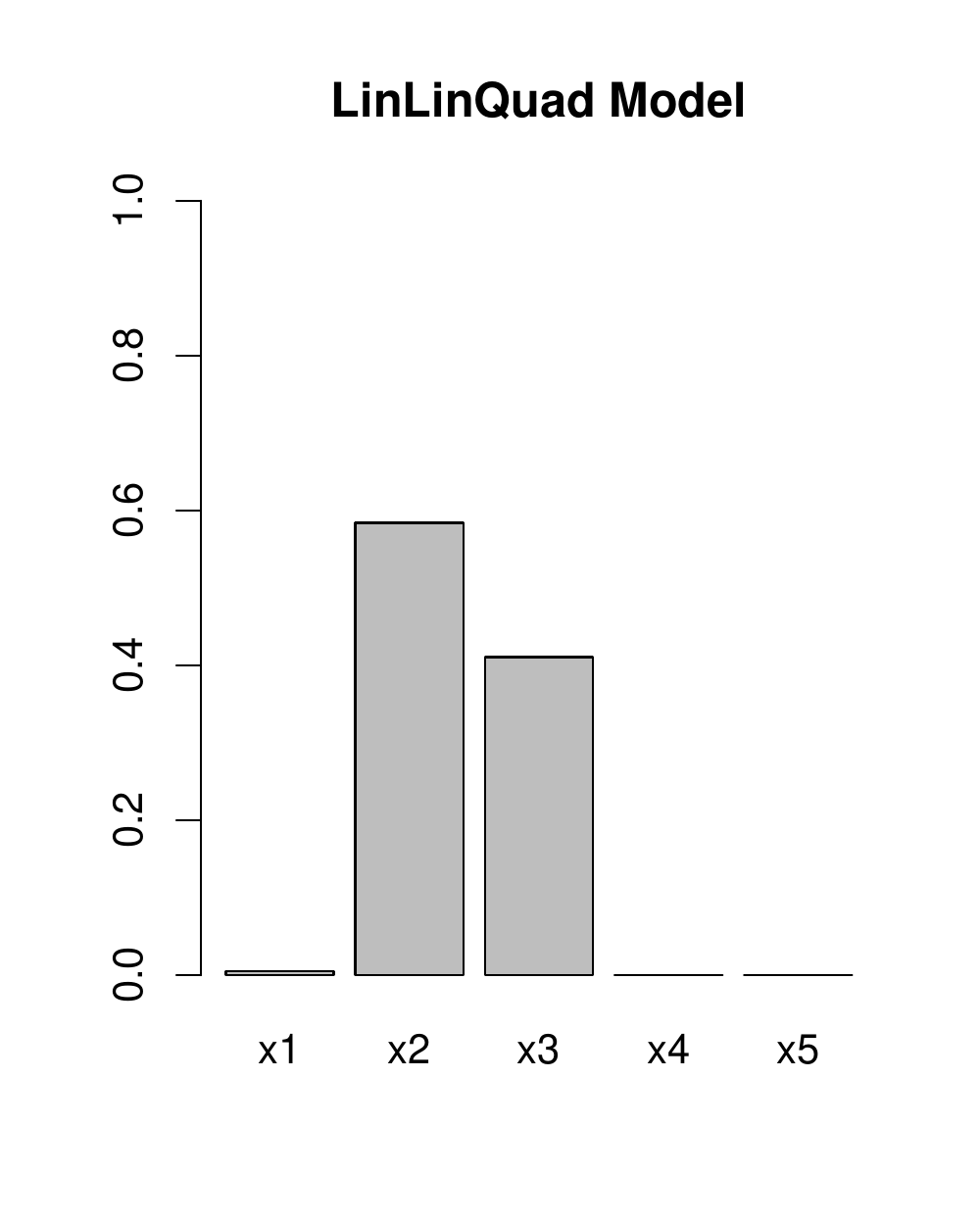}}
\caption[Estimated probabilities of split variable selection for simple linear logistic regression option.]{Estimated probabilities of split variable selection for simple linear logistic regression option (Algorithm \ref{A2:alg2}) under the models in Table \ref{T3:simu}.}\label{P3:SB1}
%\end{center}
\end{figure}

\begin{figure}[!ht]
%\begin{center}
\makebox[\textwidth][c]{\includegraphics[scale=0.5]{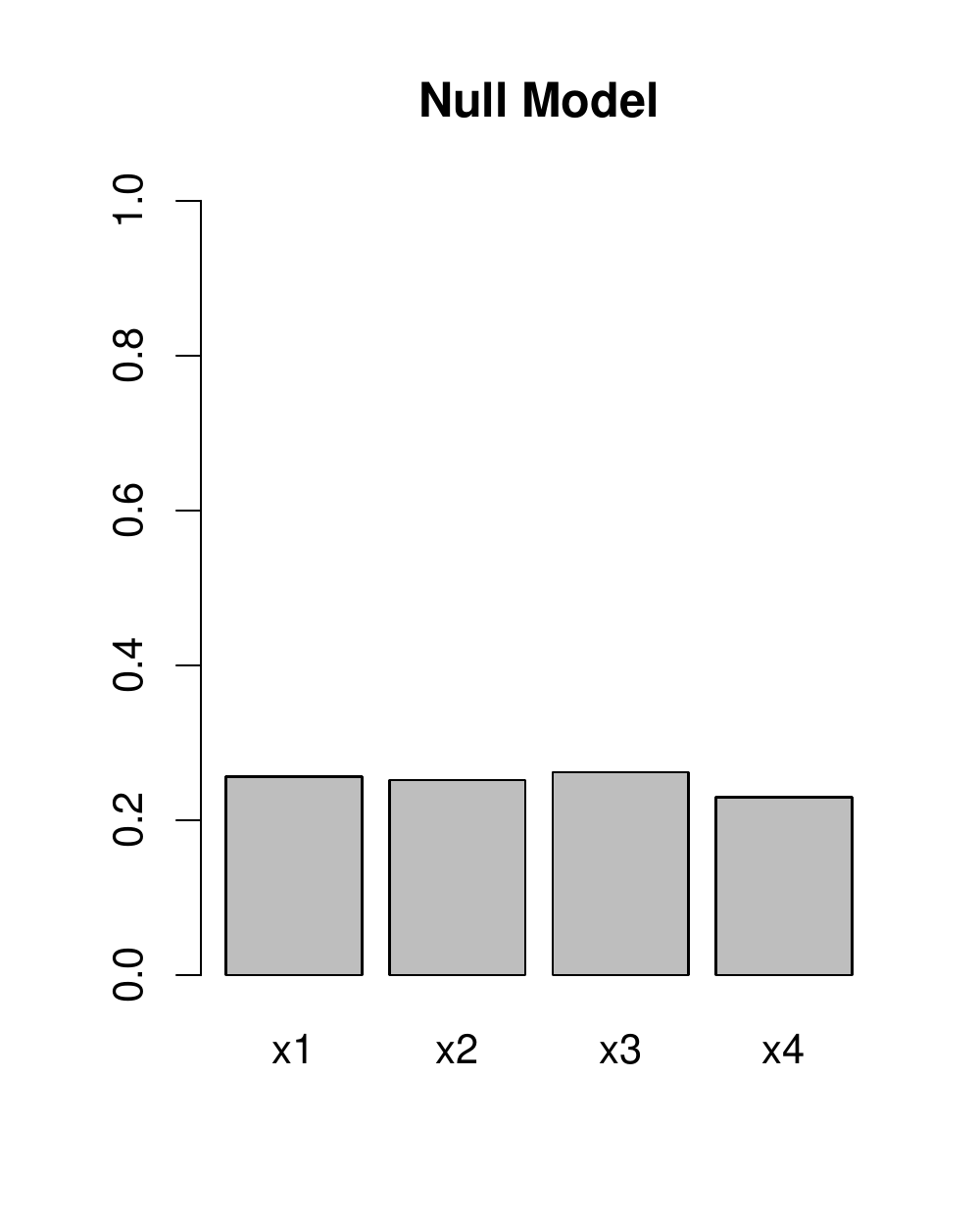}
\includegraphics[scale=0.5]{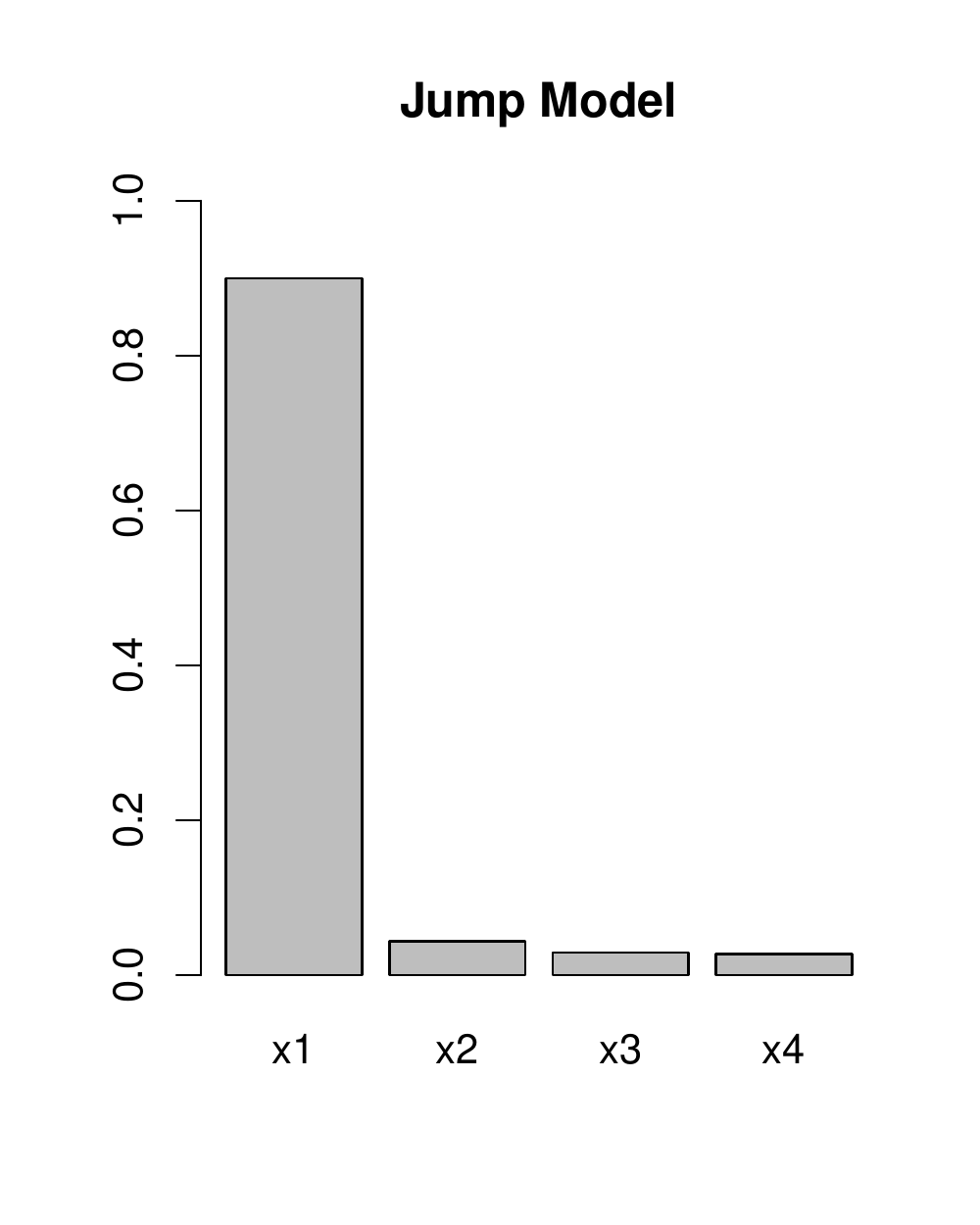}
\includegraphics[scale=0.5]{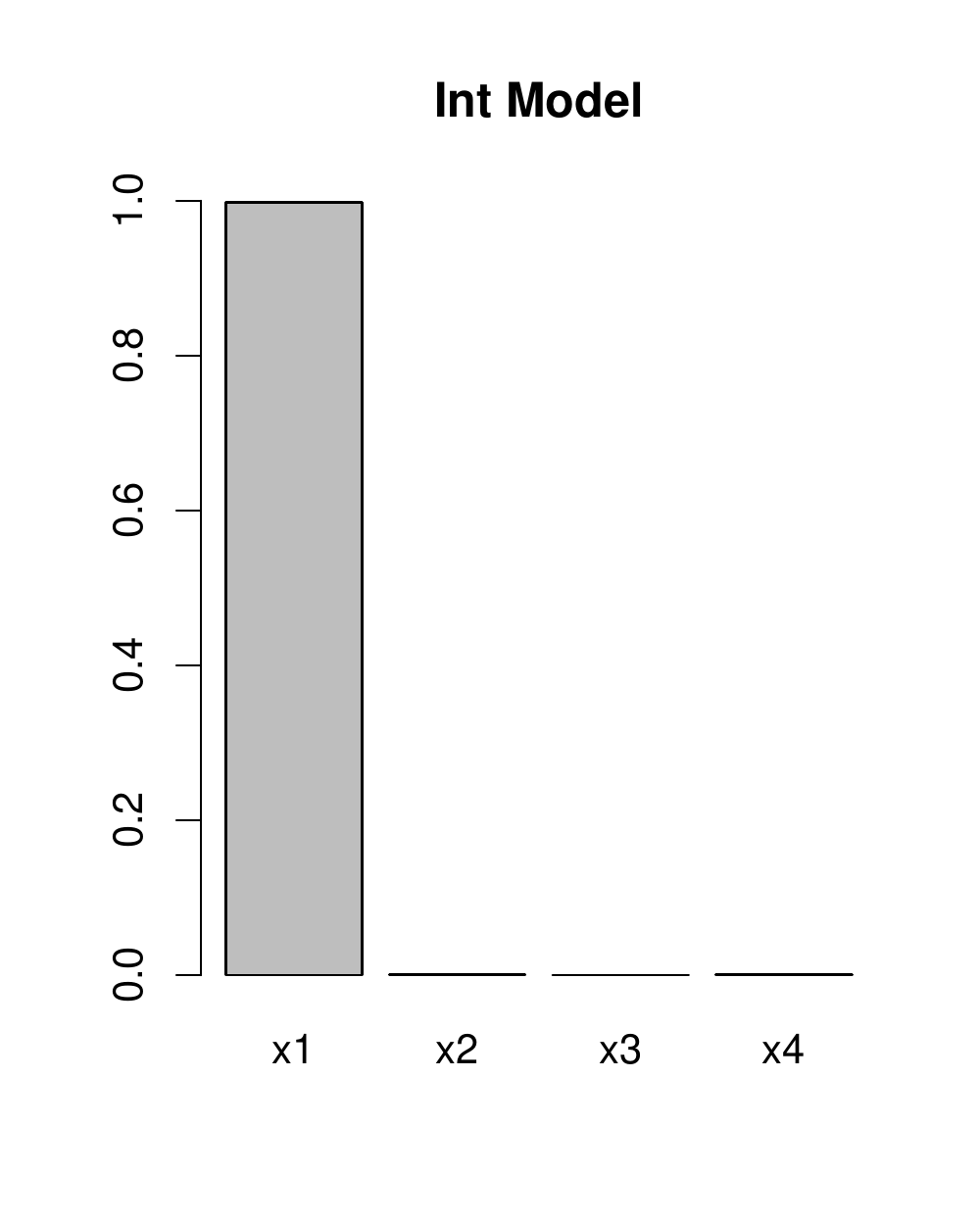}}
\makebox[\textwidth][c]{\includegraphics[scale=0.5]{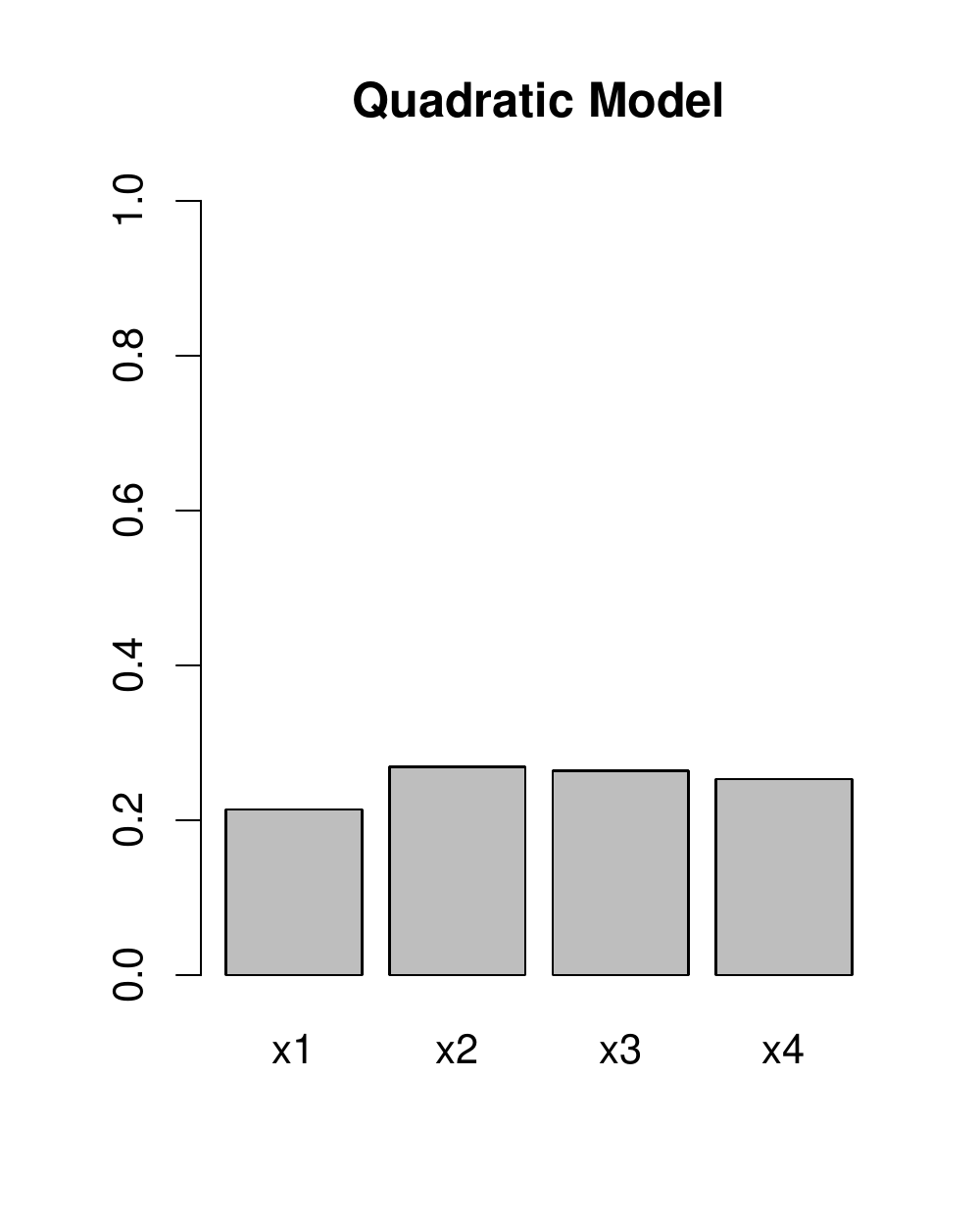}
\includegraphics[scale=0.5]{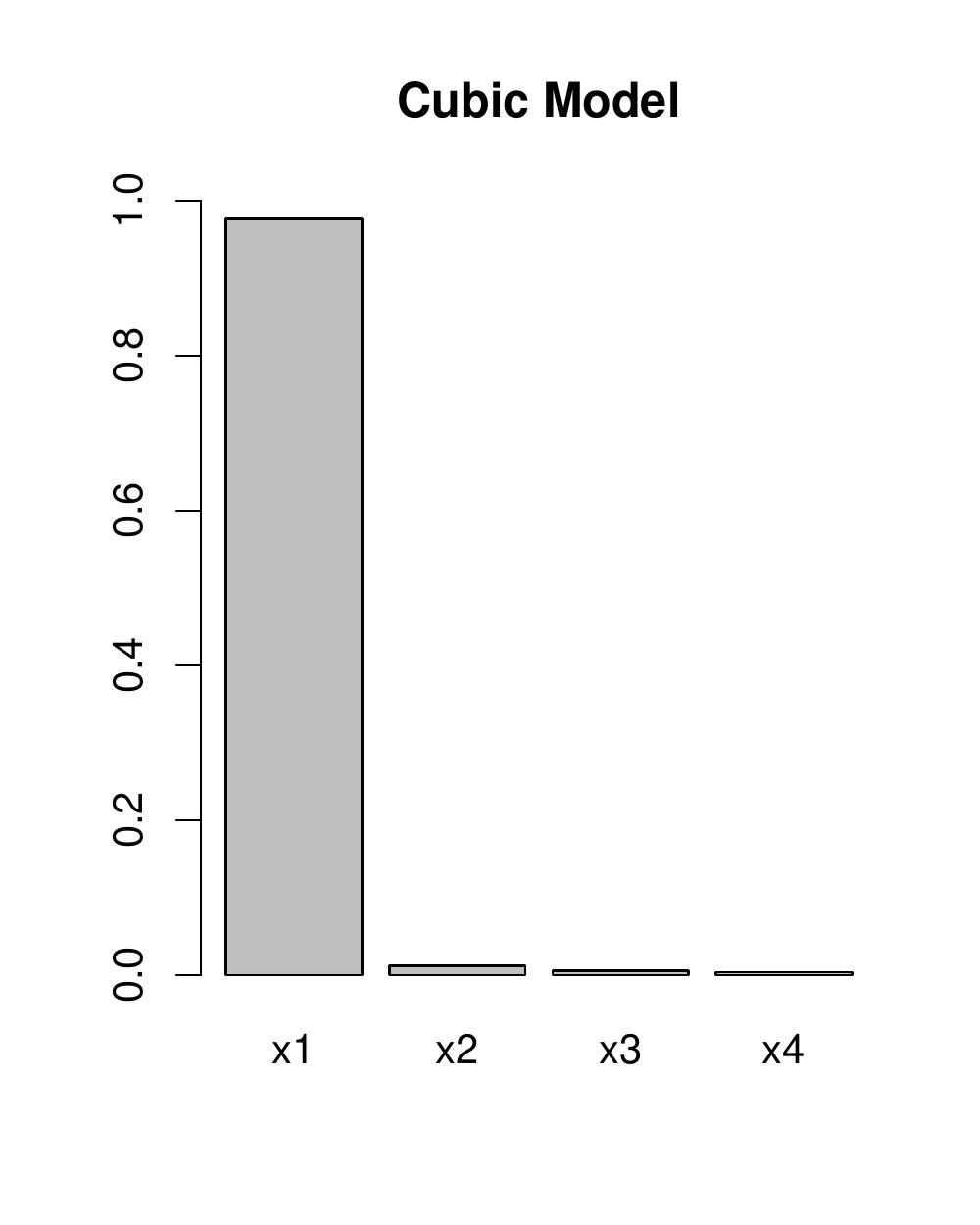}
\includegraphics[scale=0.5]{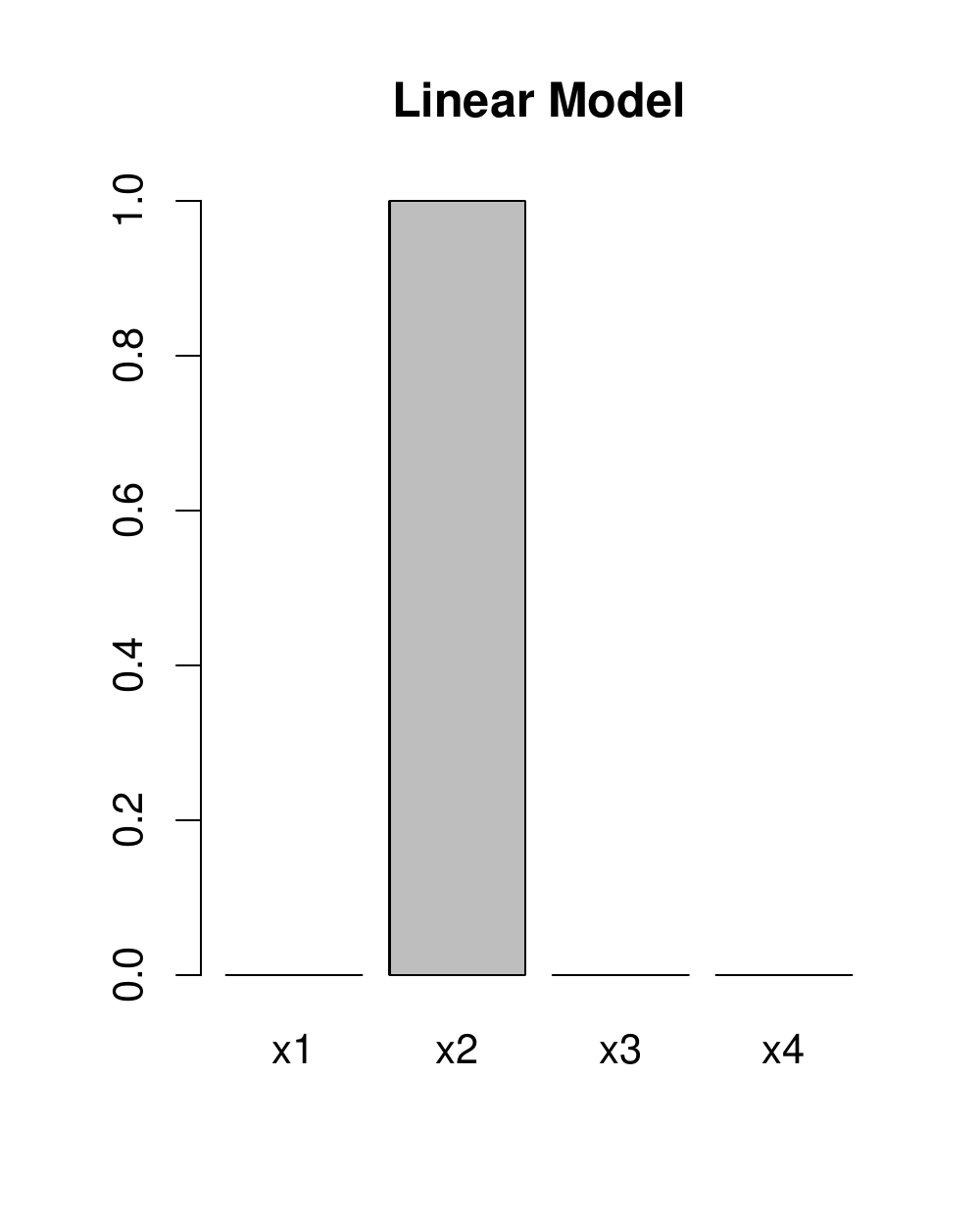}}
\makebox[\textwidth][c]{\includegraphics[scale=0.5]{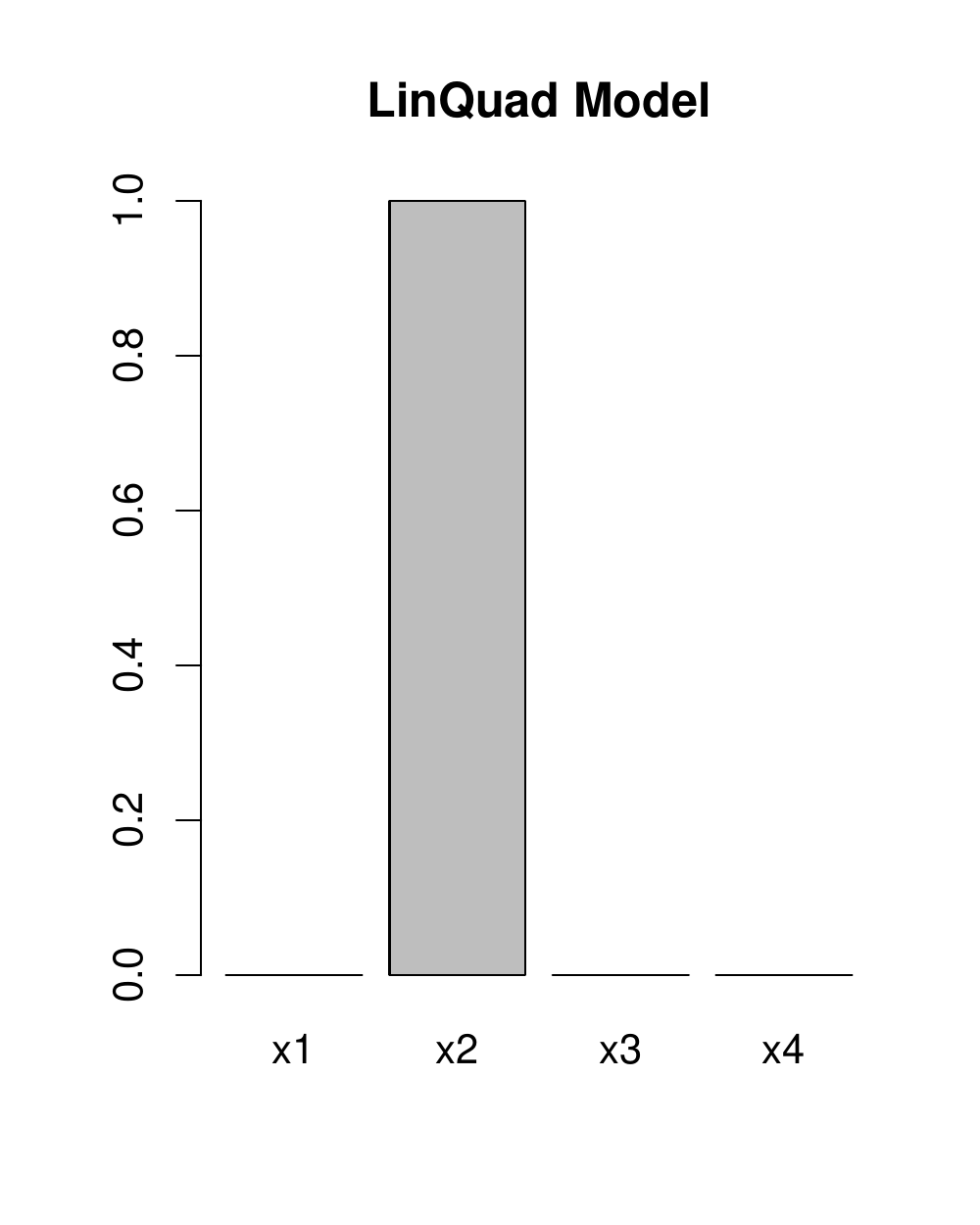}
\includegraphics[scale=0.5]{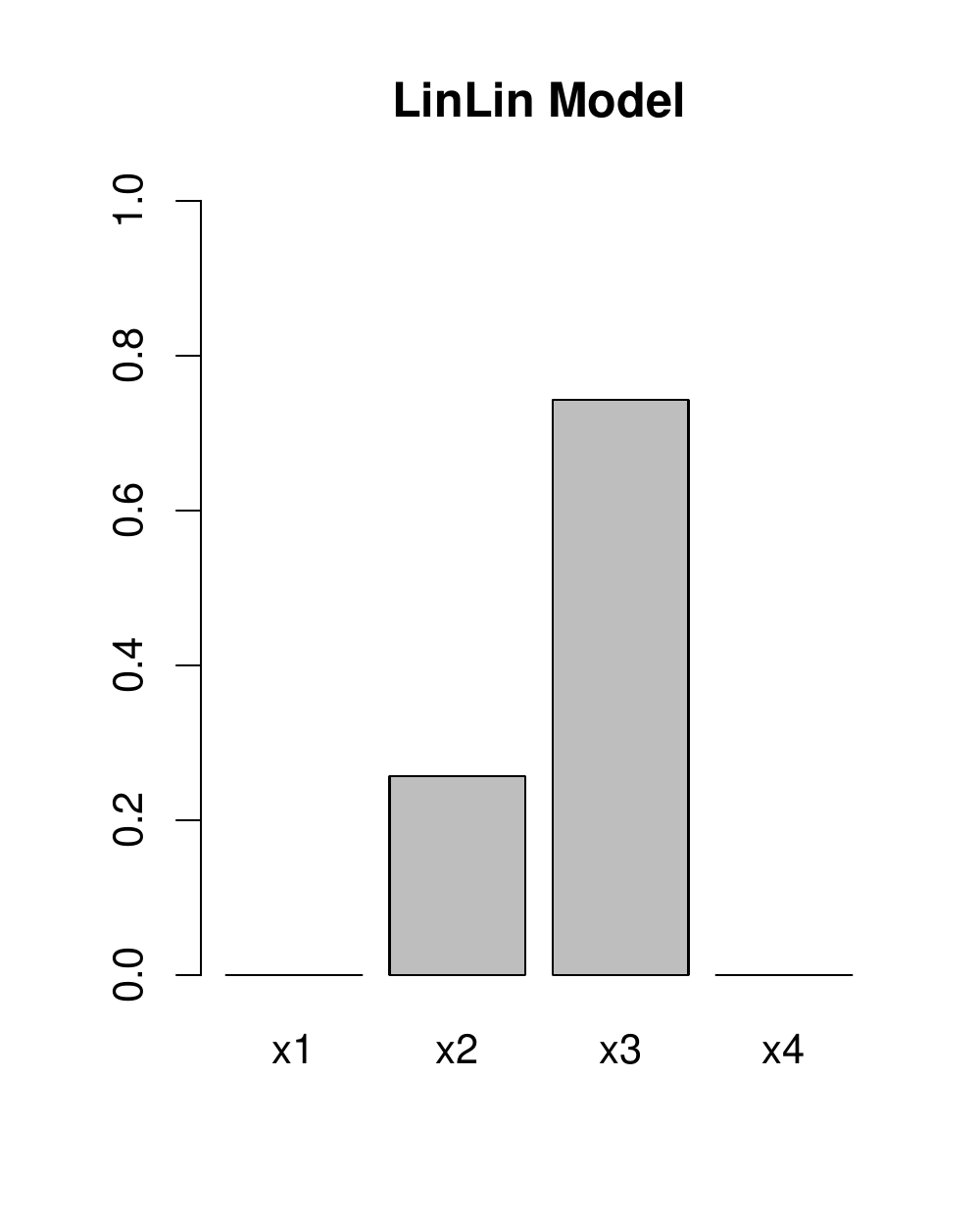}
\includegraphics[scale=0.5]{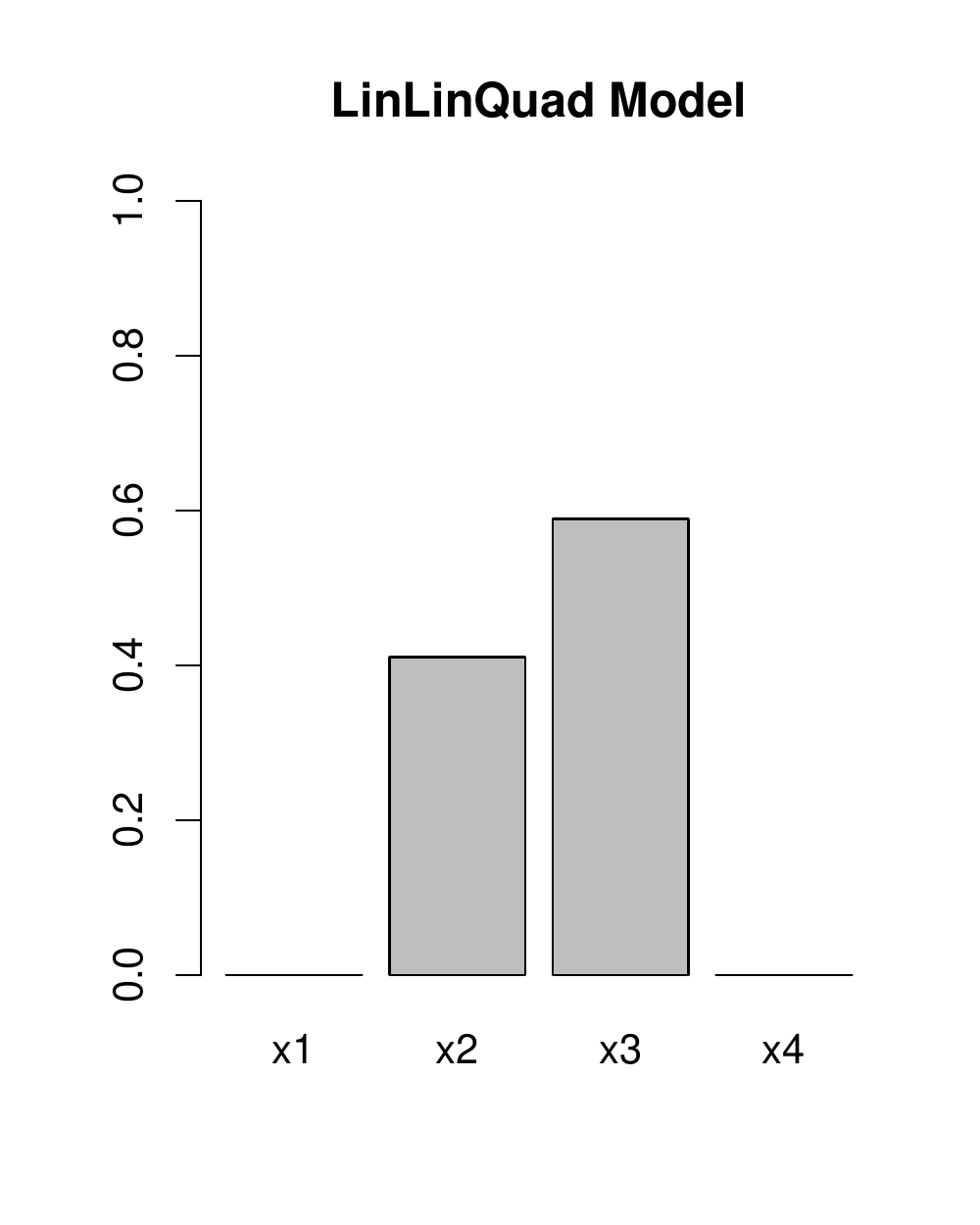}}
\caption[Estimated probabilities of best regressor selection for simple linear logistic regression option.]{Estimated probabilities of best regressor selection for simple linear logistic regression option using Algorithm \ref{A2:alg1} under the models in Table \ref{T3:simu}.}\label{P3:SB2}
%\end{center}
\end{figure}

We first assess the selection bias of Algorithm \ref{A2:alg2}, the simple linear logistic regression option. Figure \ref{P3:SB1} shows bar charts of the estimated probabilities of split variable selection under each of the 9 models. To help explain the plots in Figure \ref{P3:SB1}, the estimated probabilities that predictors were chosen as the best regressor in this simulation experiment are shown in Figure \ref{P3:SB2}.

\begin{figure}[!ht]
%\begin{center}
\makebox[\textwidth][c]{\includegraphics[scale=0.5]{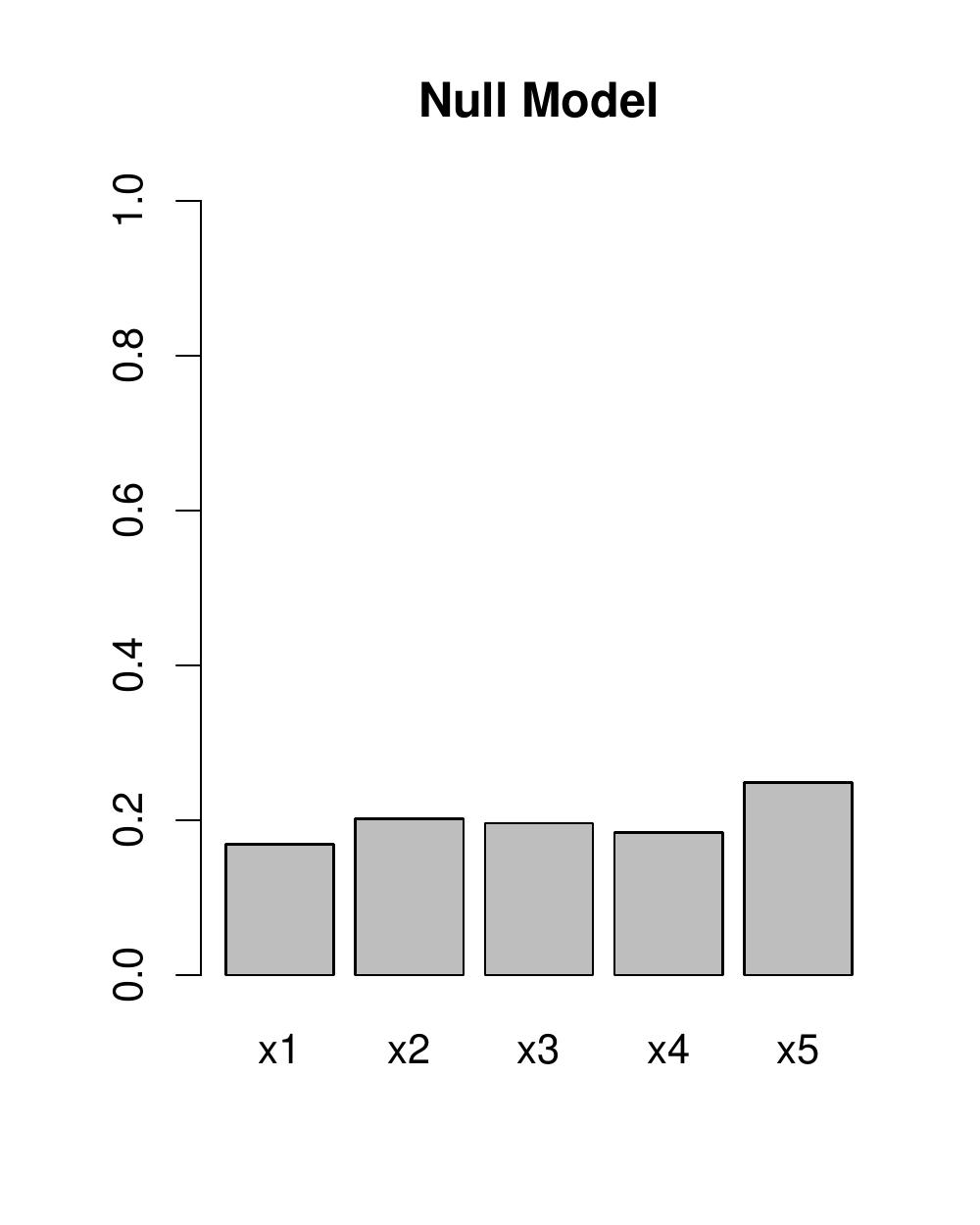}
\includegraphics[scale=0.5]{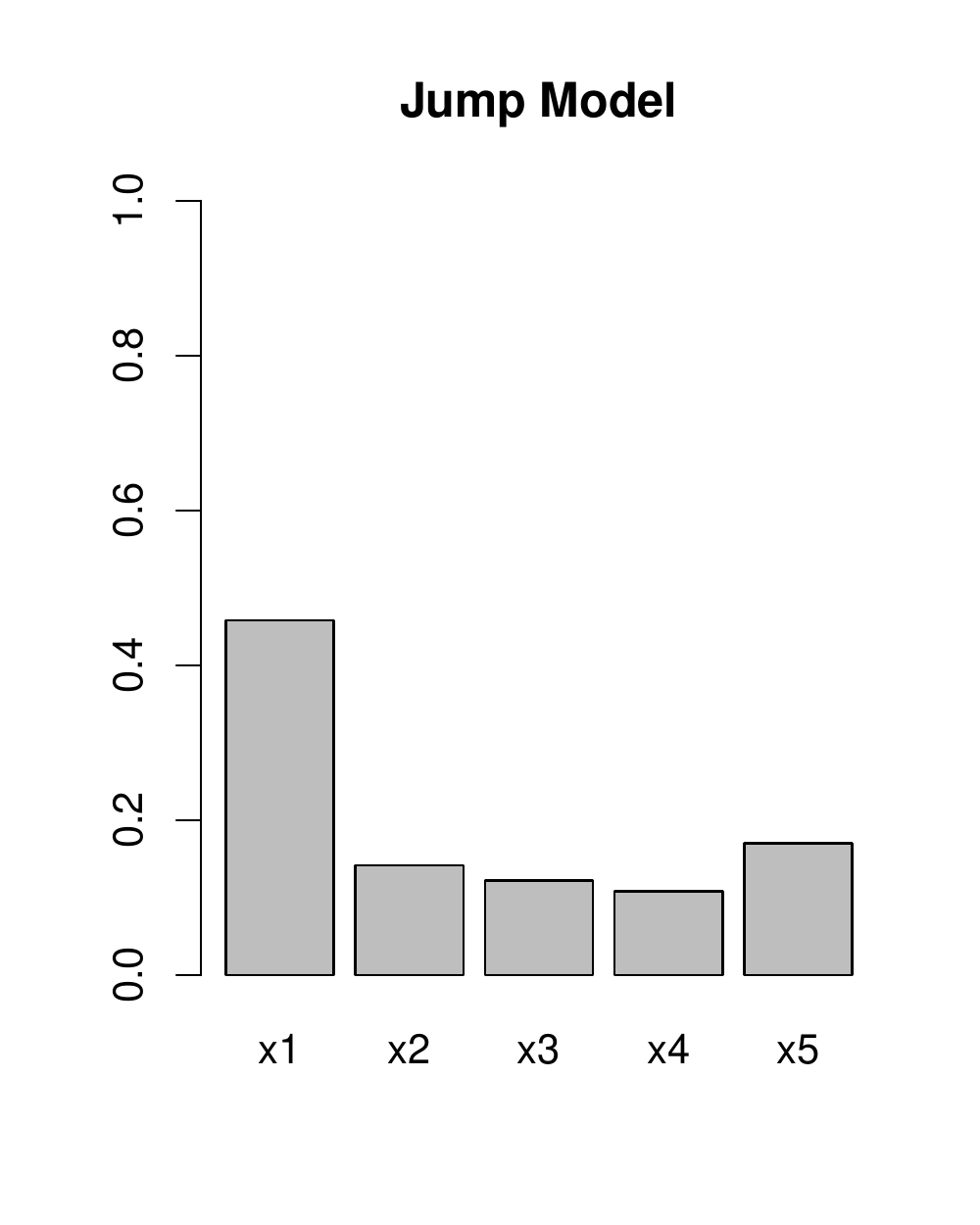}
\includegraphics[scale=0.5]{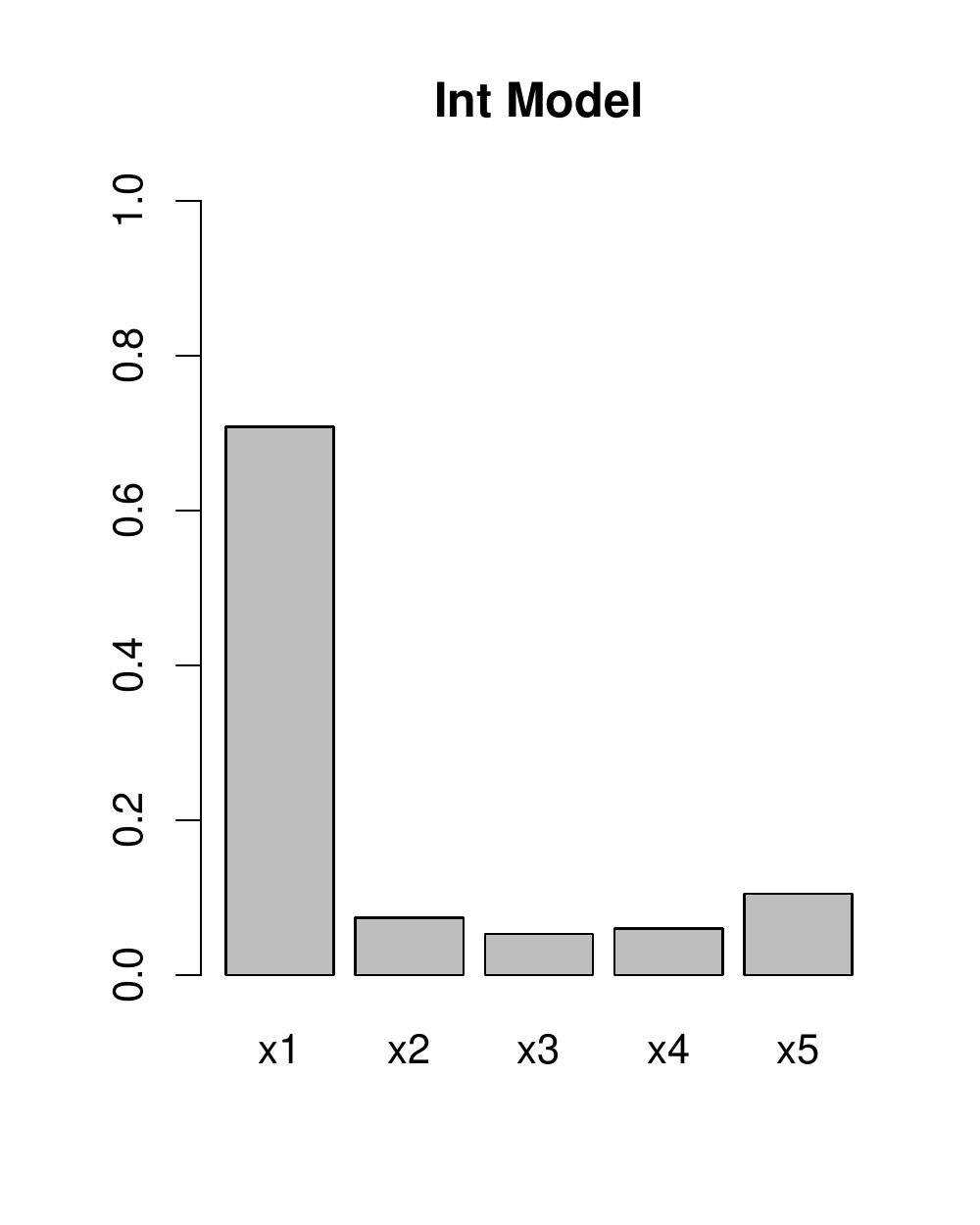}}
\makebox[\textwidth][c]{\includegraphics[scale=0.5]{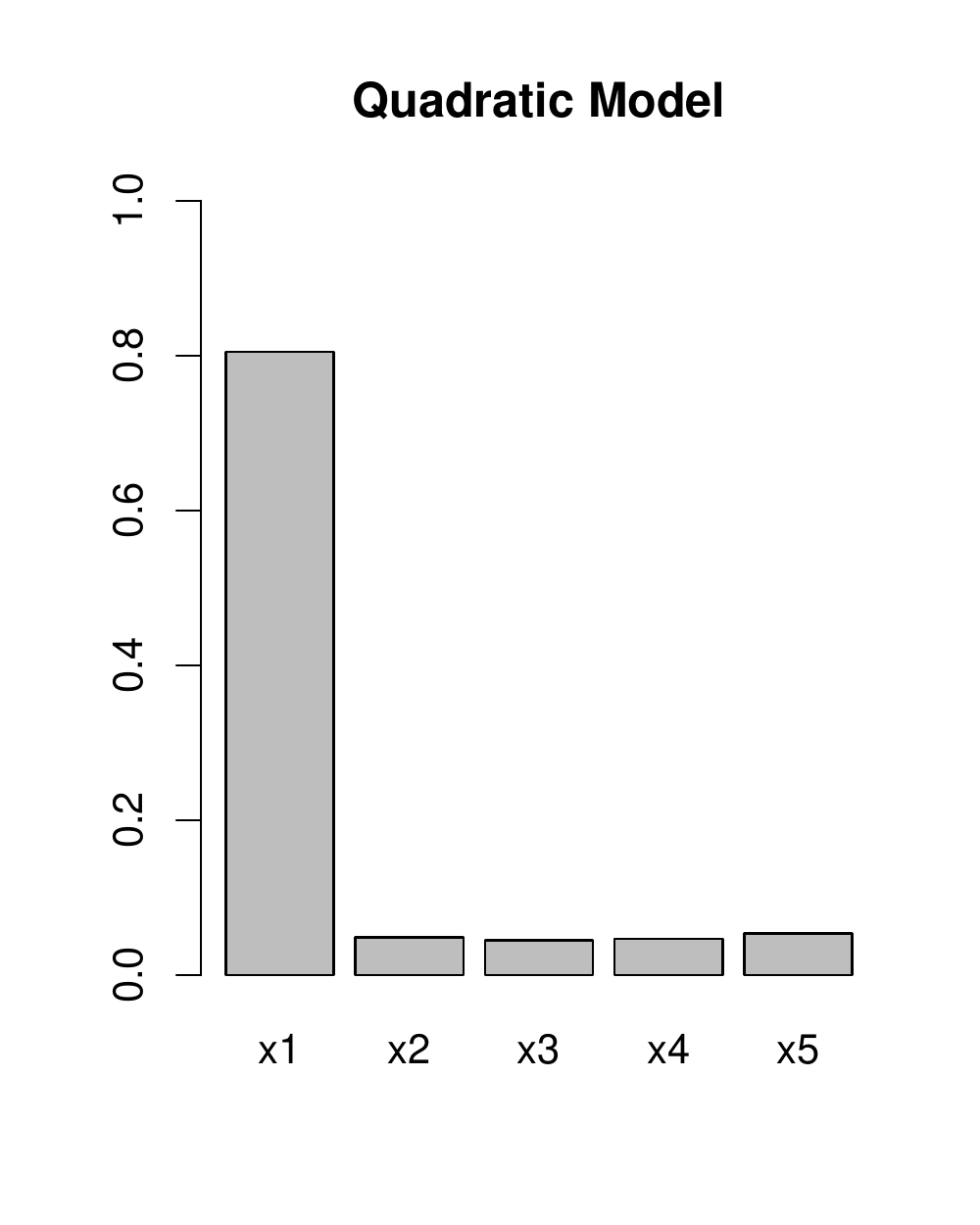}
\includegraphics[scale=0.5]{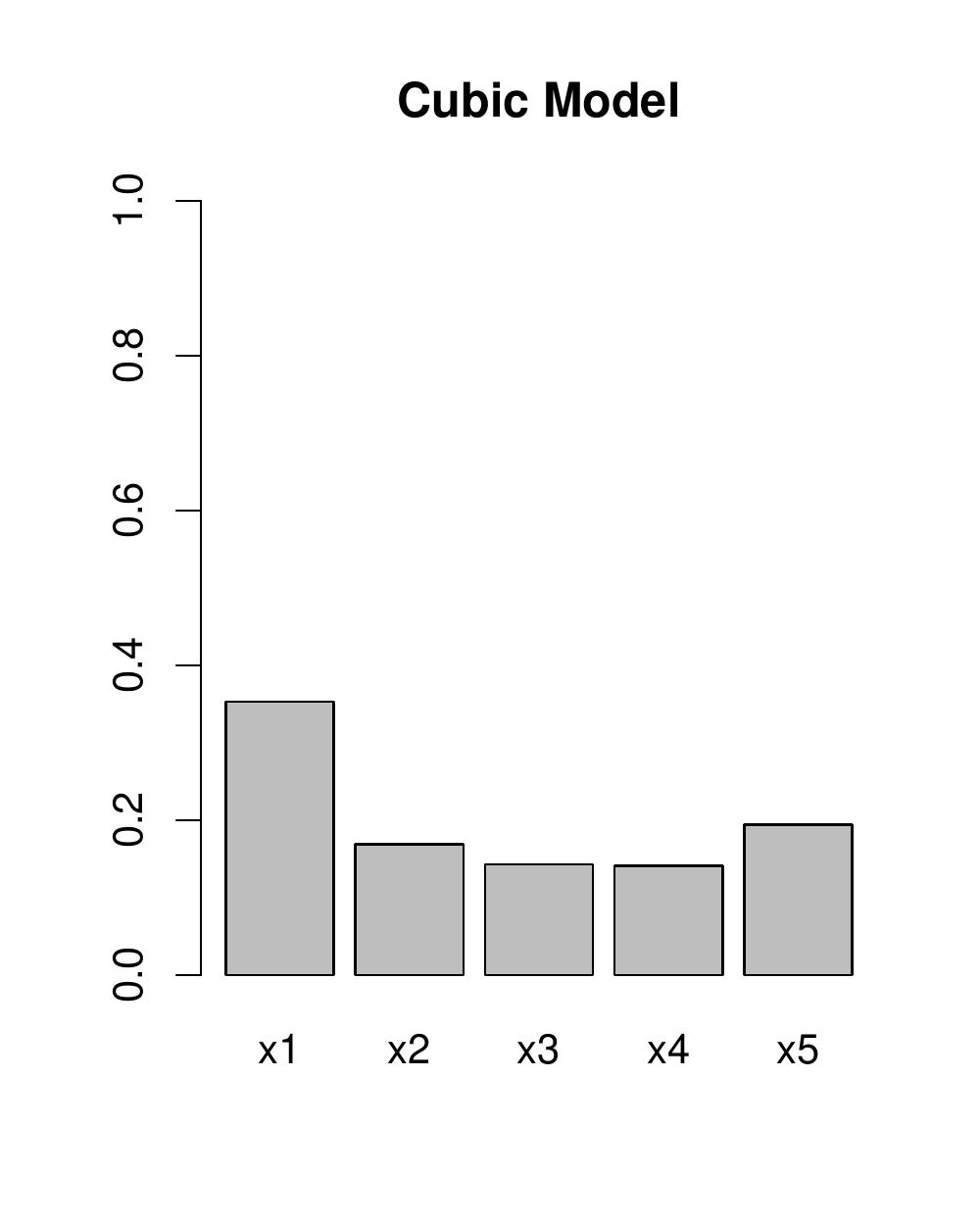}
\includegraphics[scale=0.5]{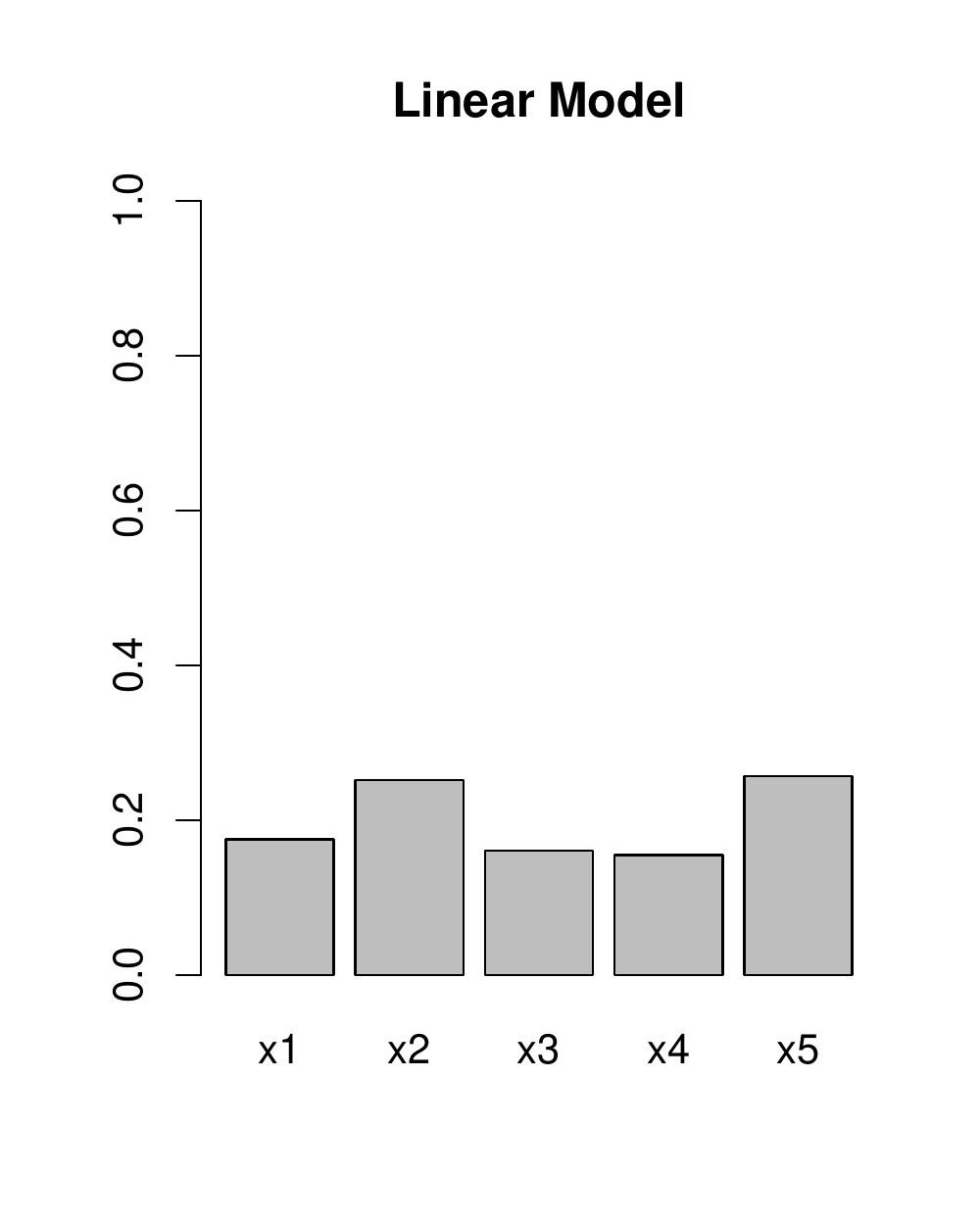}}
\makebox[\textwidth][c]{\includegraphics[scale=0.5]{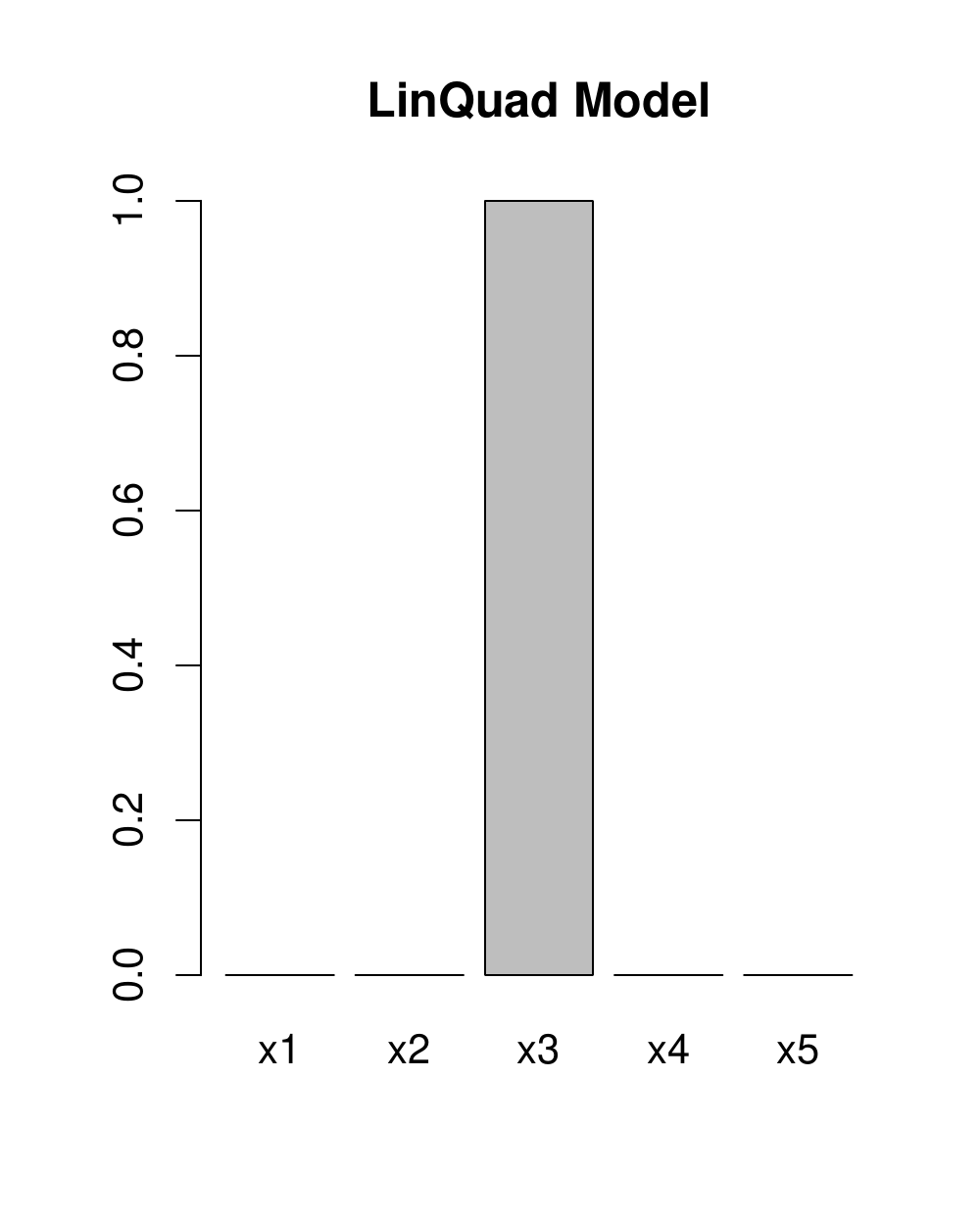}
\includegraphics[scale=0.5]{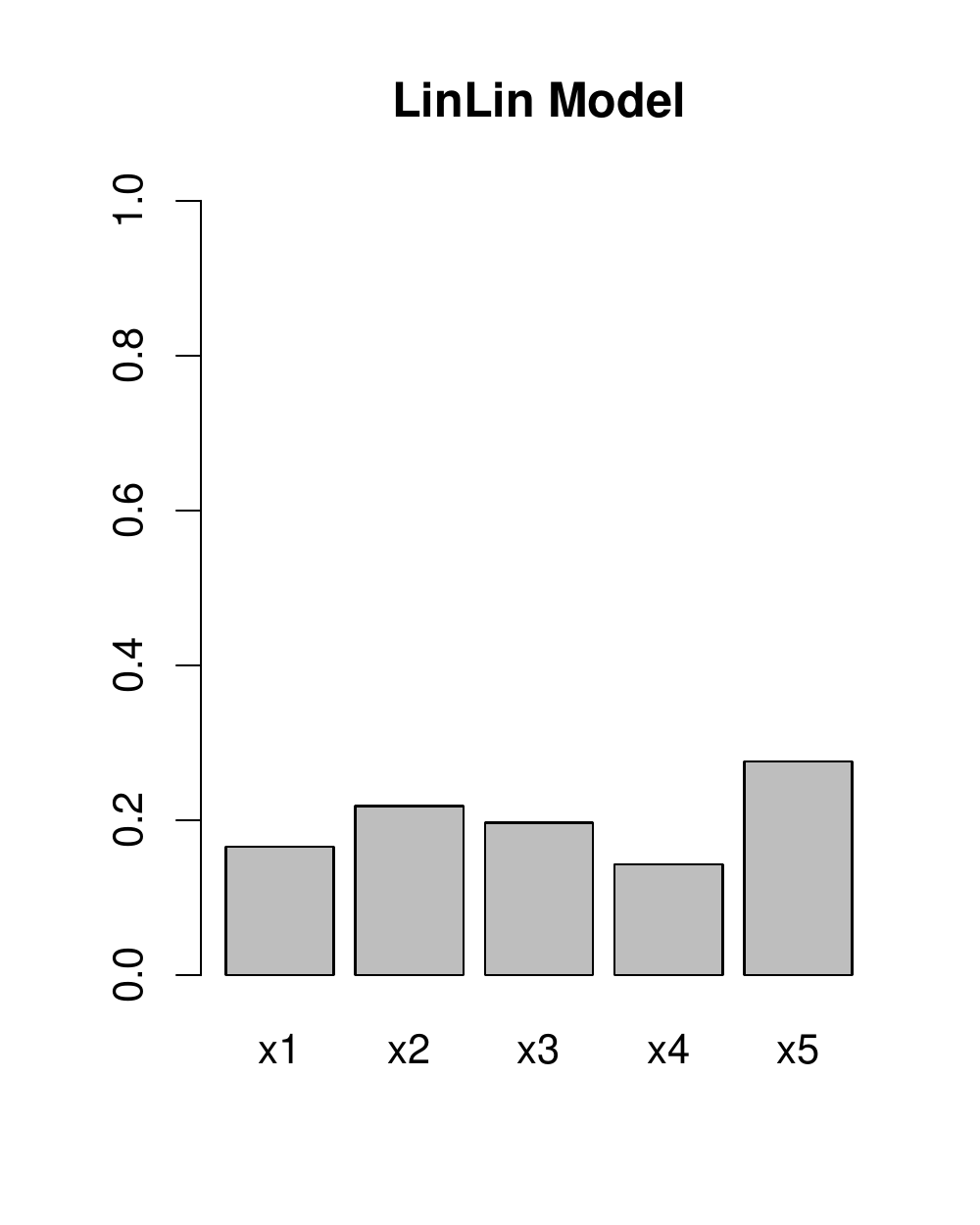}
\includegraphics[scale=0.5]{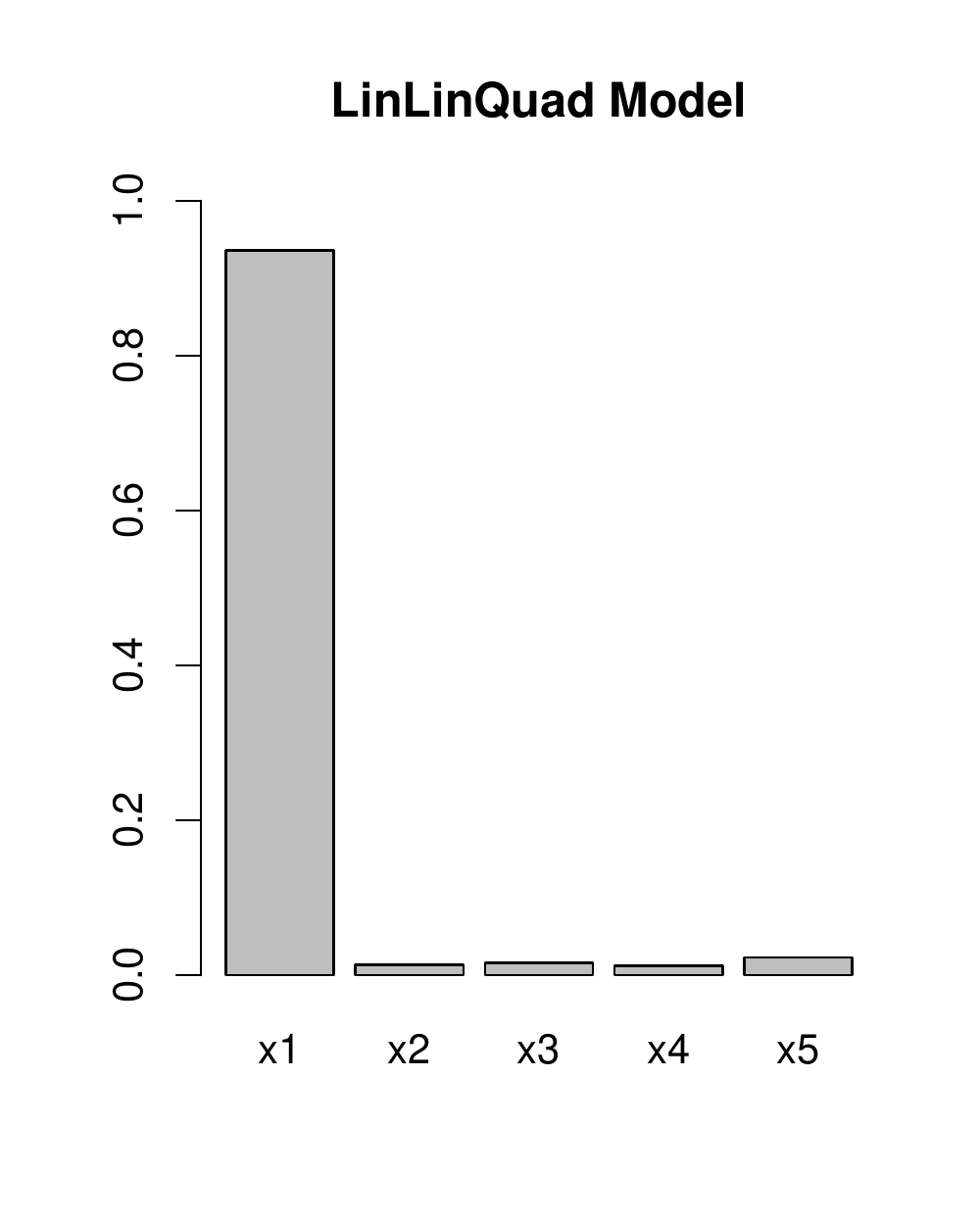}}
\caption[Estimated probabilities of split variable selection for multiple linear logistic regression option.]{Estimated probabilities of split variable selection for multiple linear logistic regression option (Algorithm \ref{A2:alg3}) under the models in Table \ref{T3:simu}.}\label{P3:SB3}
%\end{center}
\end{figure}

We repeat the above simulation experiment on Algorithm \ref{A2:alg3}, the multiple linear logistic regression option. The estimated selection probabilities are shown in Figure \ref{P3:SB3}.

We can see from these Figures that:
\begin{enumerate}
\item Under the Null model where $Y$ is distributed independently of the five predictors, both Algorithm \ref{A2:alg2} and \ref{A2:alg3} select each of the numerical predictors with approximately equal probability. However, we find that the categorical variable $X_5$ has a slightly greater chance of being selected than the numerical ones in both simulation experiments. In other words, both split variable selection algorithms show preference for categorical variables under the Null model.
\item For the Jump, Int, Quadratic, and Cubic models, where $X_1$ has nonlinear effects, both algorithms select the correct variable $X_1$ the most. The probabilities of $X_1$ being chosen to split the data depends on the strength of the signal. For example, both algorithms have a higher chance of selecting $X_1$ under the Quadratic model than under the Cubic model.
\item Selection bias is not apparent for the simple linear logistic regression option of PLUTO under the Linear model. While with the multiple linear option, there seems to be bias towards selecting $X_2$ and $X_5$ over others.
\item The LinLin and LinLinQuad models show the demerits of fitting simple linear logistic node models. A successful unbiased selection procedure should select each variable with equal probability under the LinLin model, and it should select $X_1$ as the split variable under the LinLinQuad model. Figure \ref{P3:SB1} shows that the simple linear logistic regression option does not possess this property. Because it only fits node models with one single best regressor, i.e., only one of $X_2$ and $X_3$ is fitted as shown in Figure \ref{P3:SB2}, the remaining one will then be selected frequently as the split variable. Similarly, under the LinLinQuad model, Algorithm \ref{A2:alg2} selects $X_2$ and $X_3$ much more frequently than $X_1$. Algorithm \ref{A2:alg3} shows a significant advantage under these two models. However, it still has a slight preference for the categorical split variable.
\end{enumerate}

\subsubsection{Bias correction}
\begin{figure}[!h]
\makebox[\textwidth][c]{\includegraphics[scale=0.5]{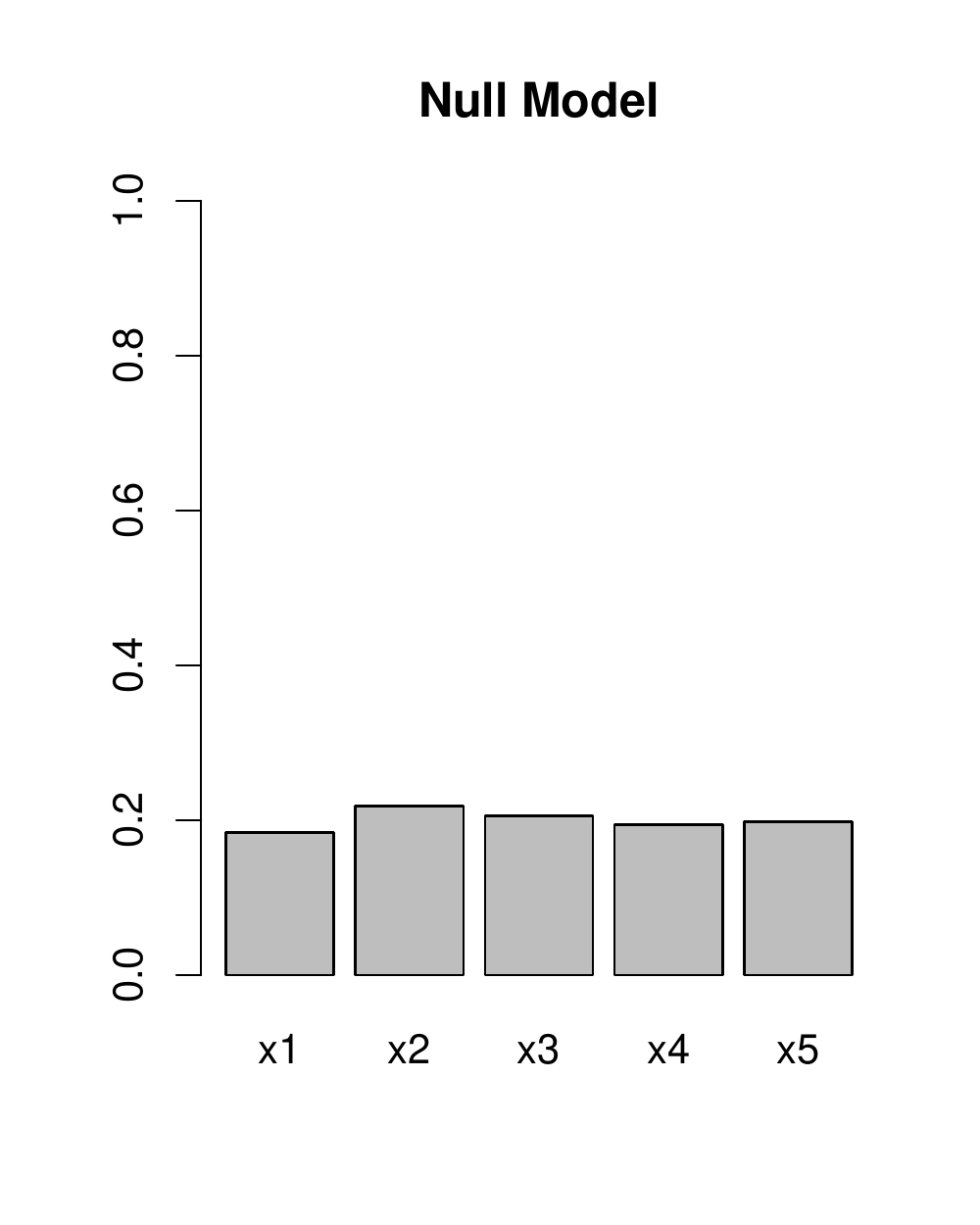}
\includegraphics[scale=0.5]{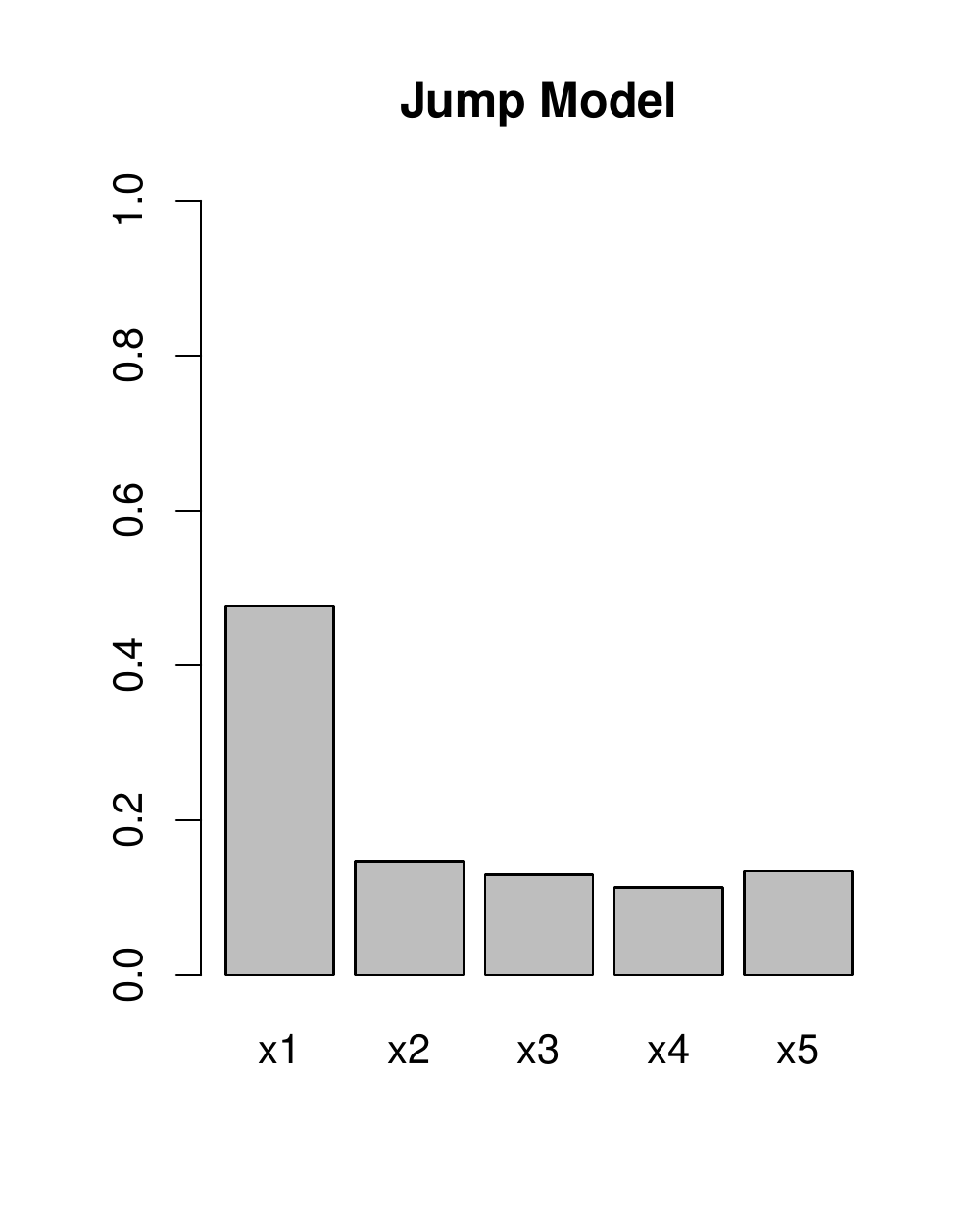}
\includegraphics[scale=0.5]{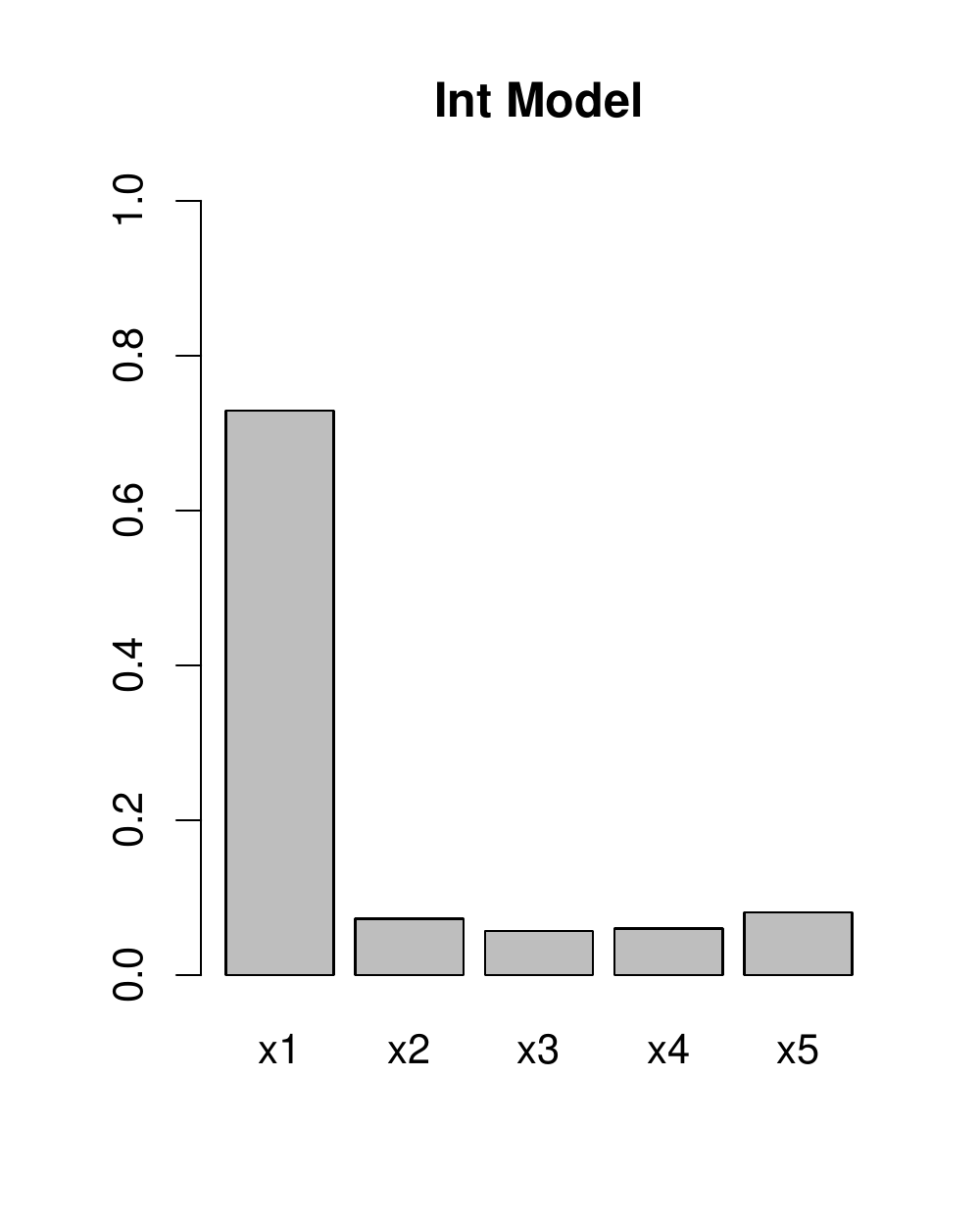}}
\makebox[\textwidth][c]{\includegraphics[scale=0.5]{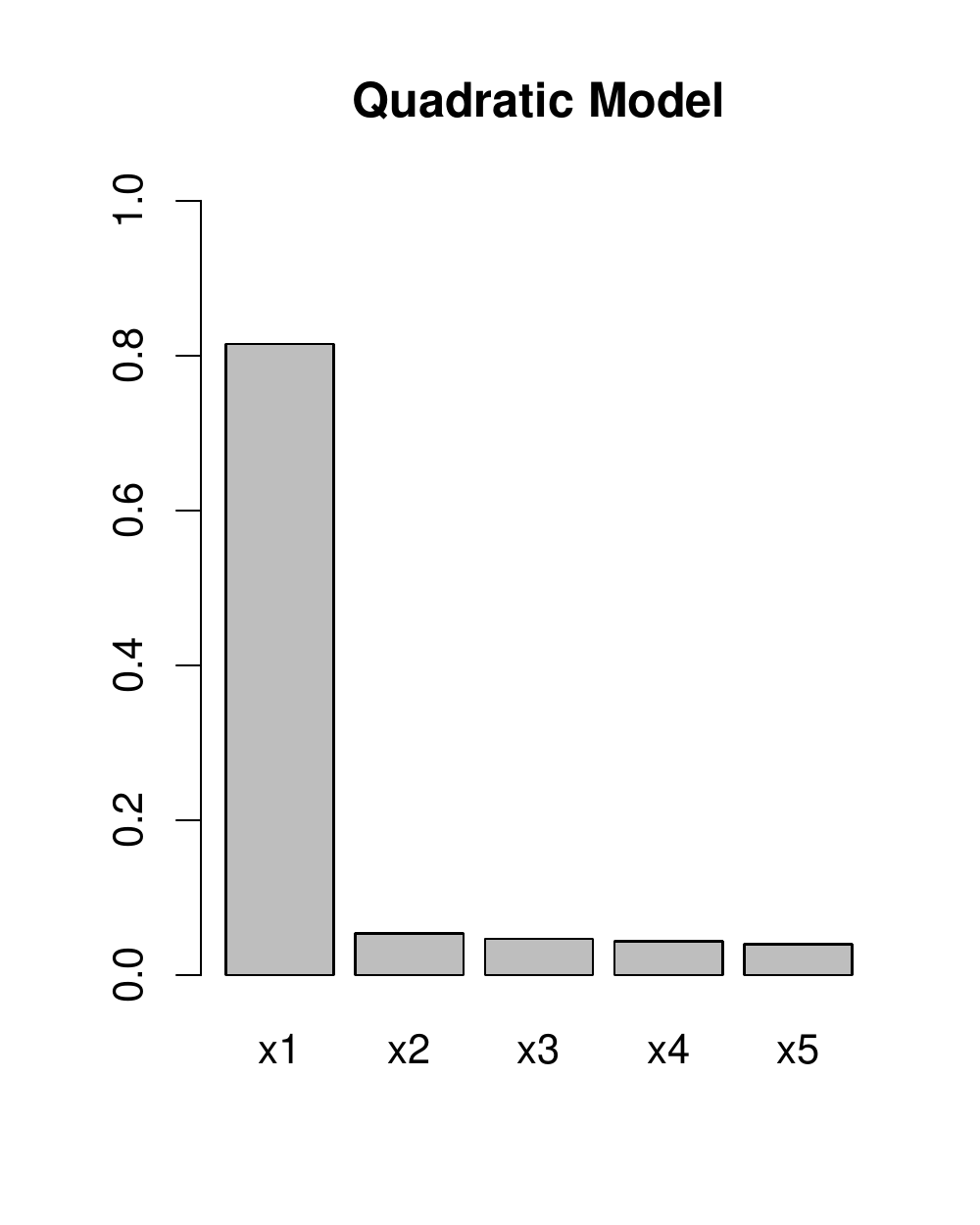}
\includegraphics[scale=0.5]{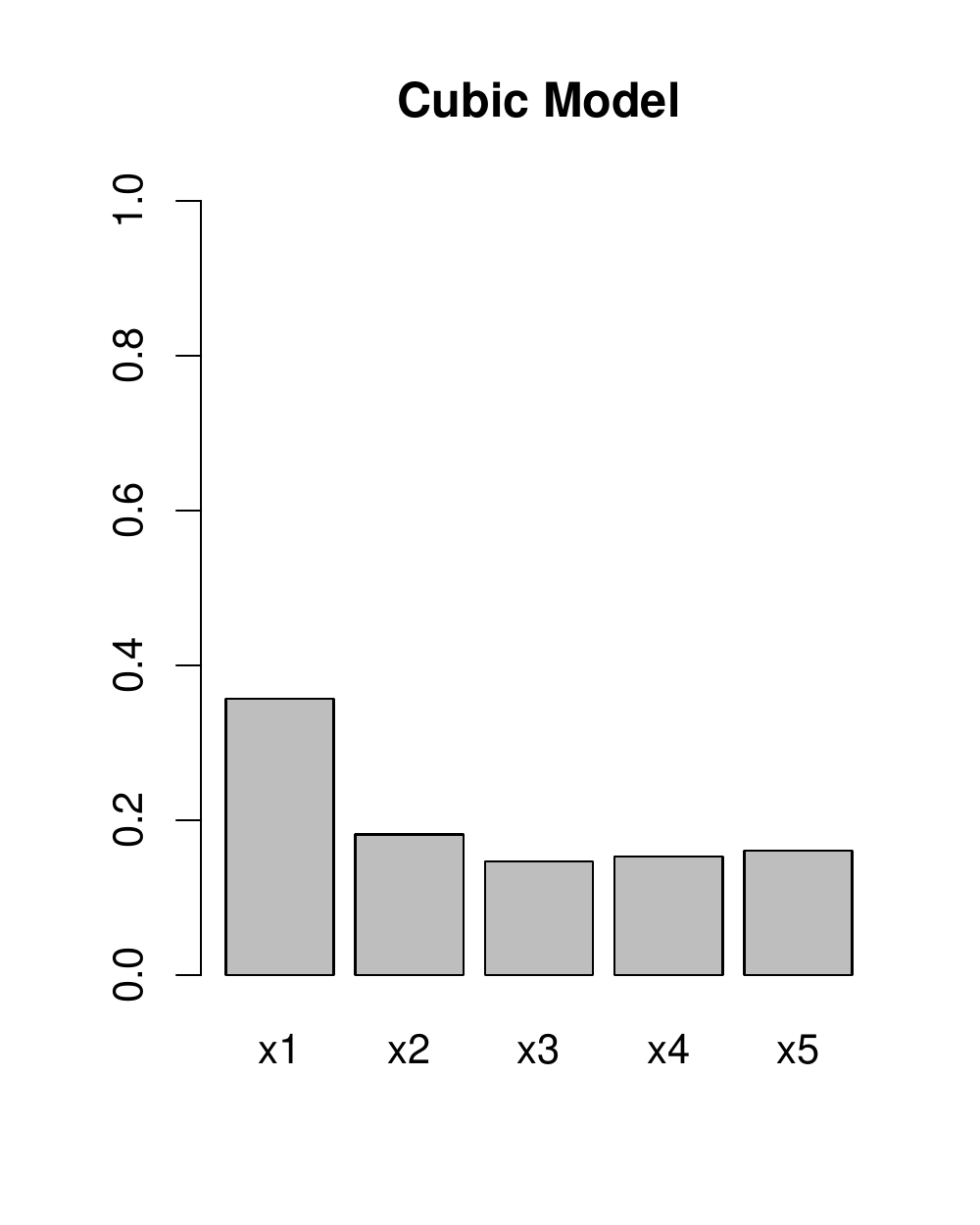}
\includegraphics[scale=0.5]{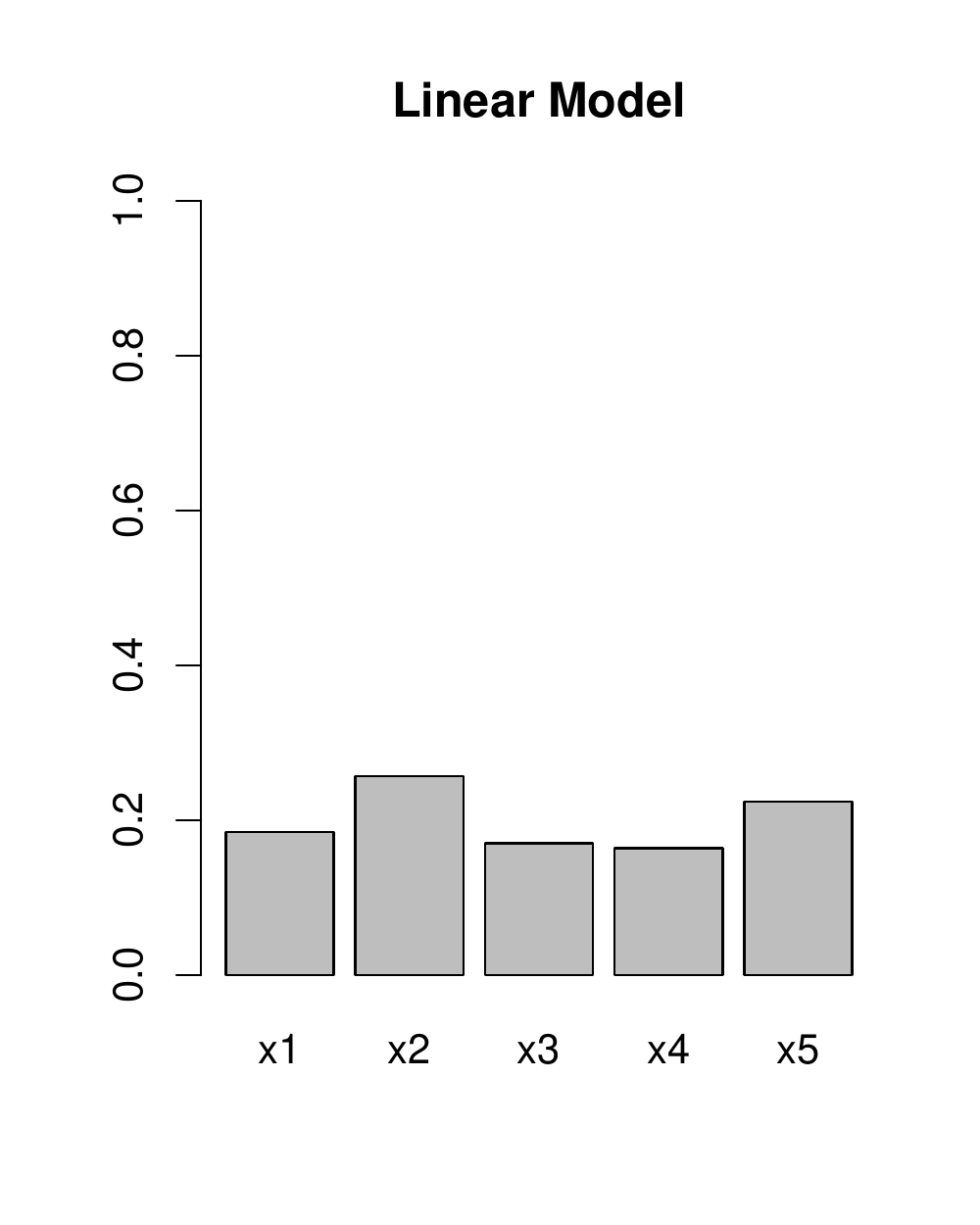}}
\makebox[\textwidth][c]{\includegraphics[scale=0.5]{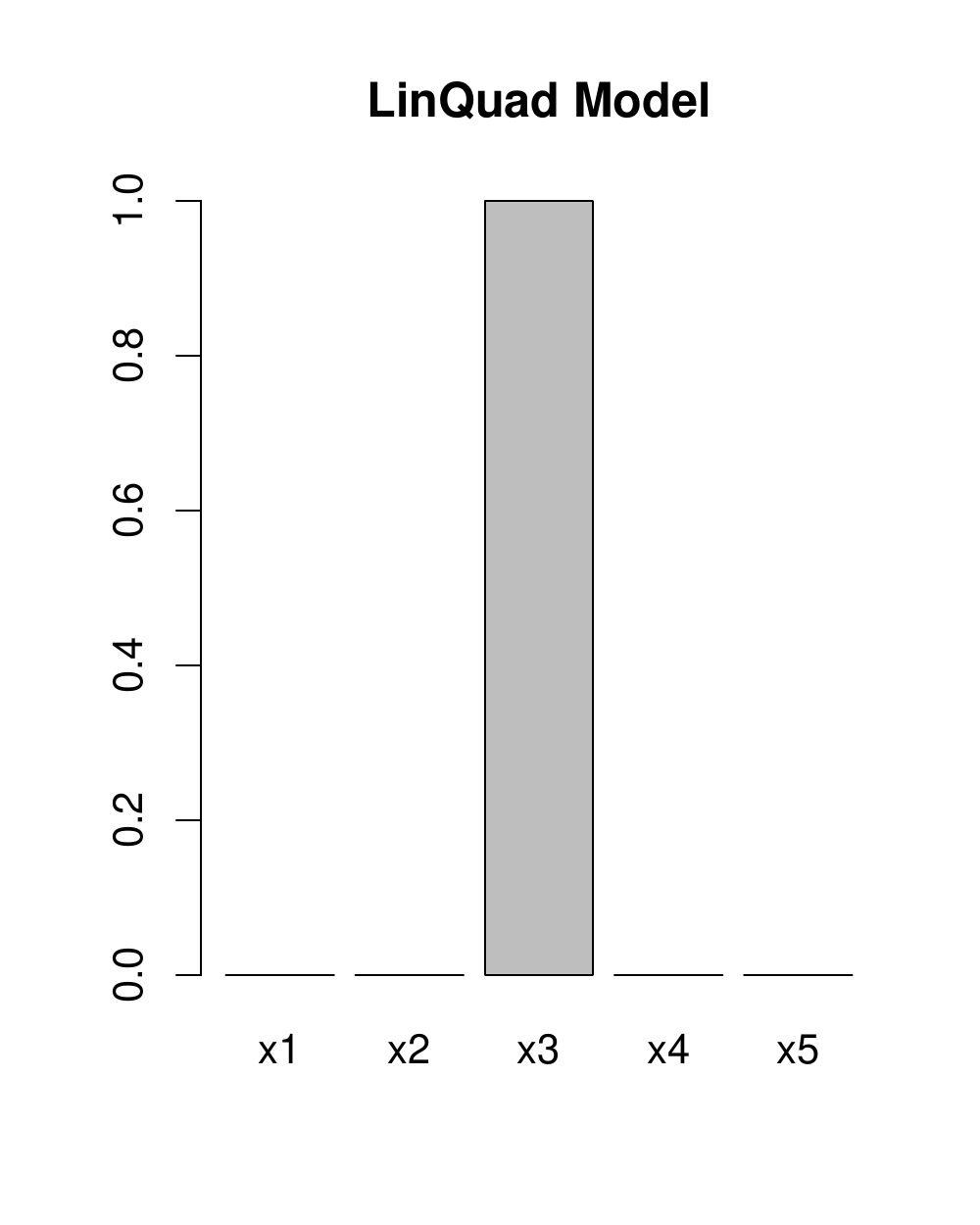}
\includegraphics[scale=0.5]{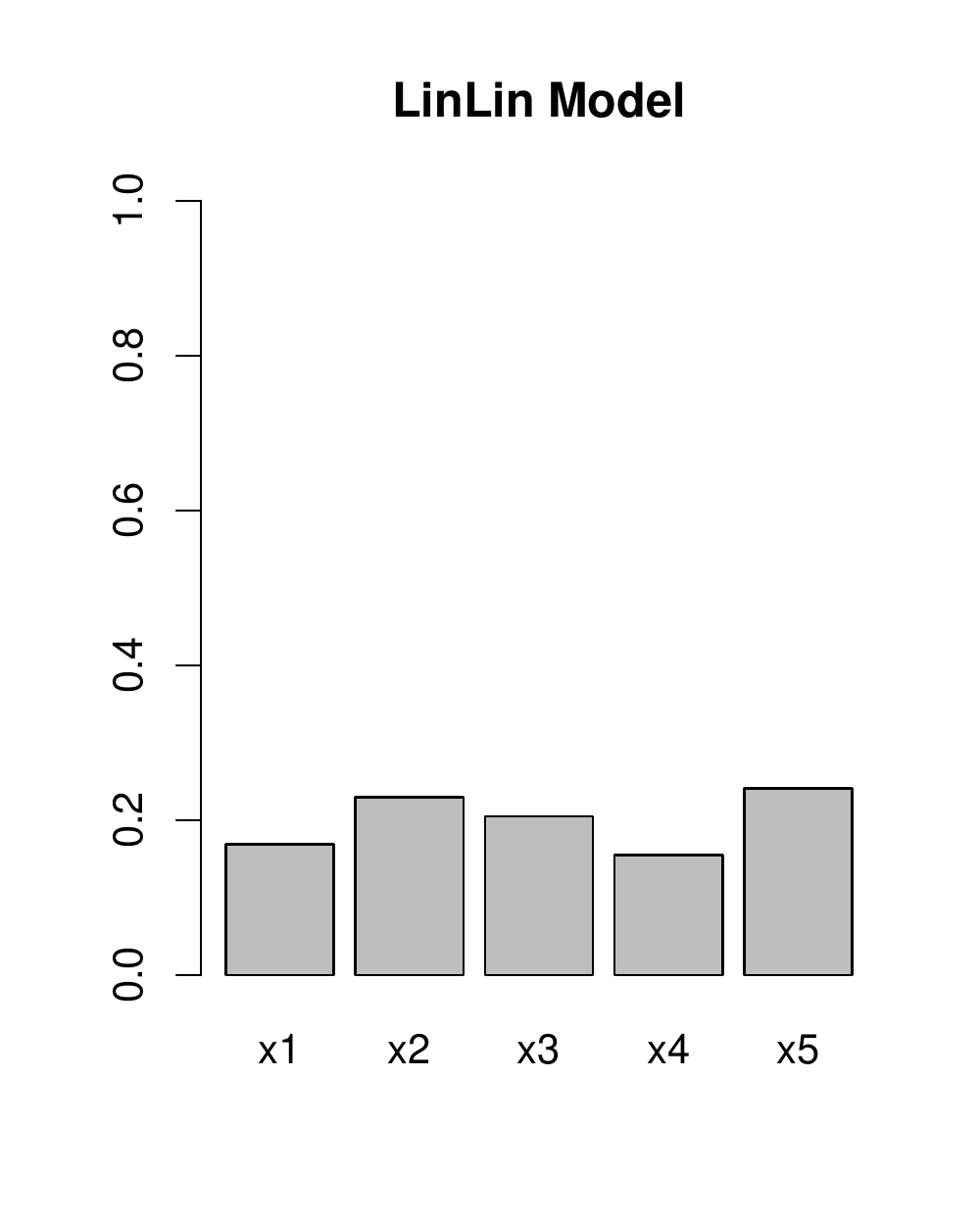}
\includegraphics[scale=0.5]{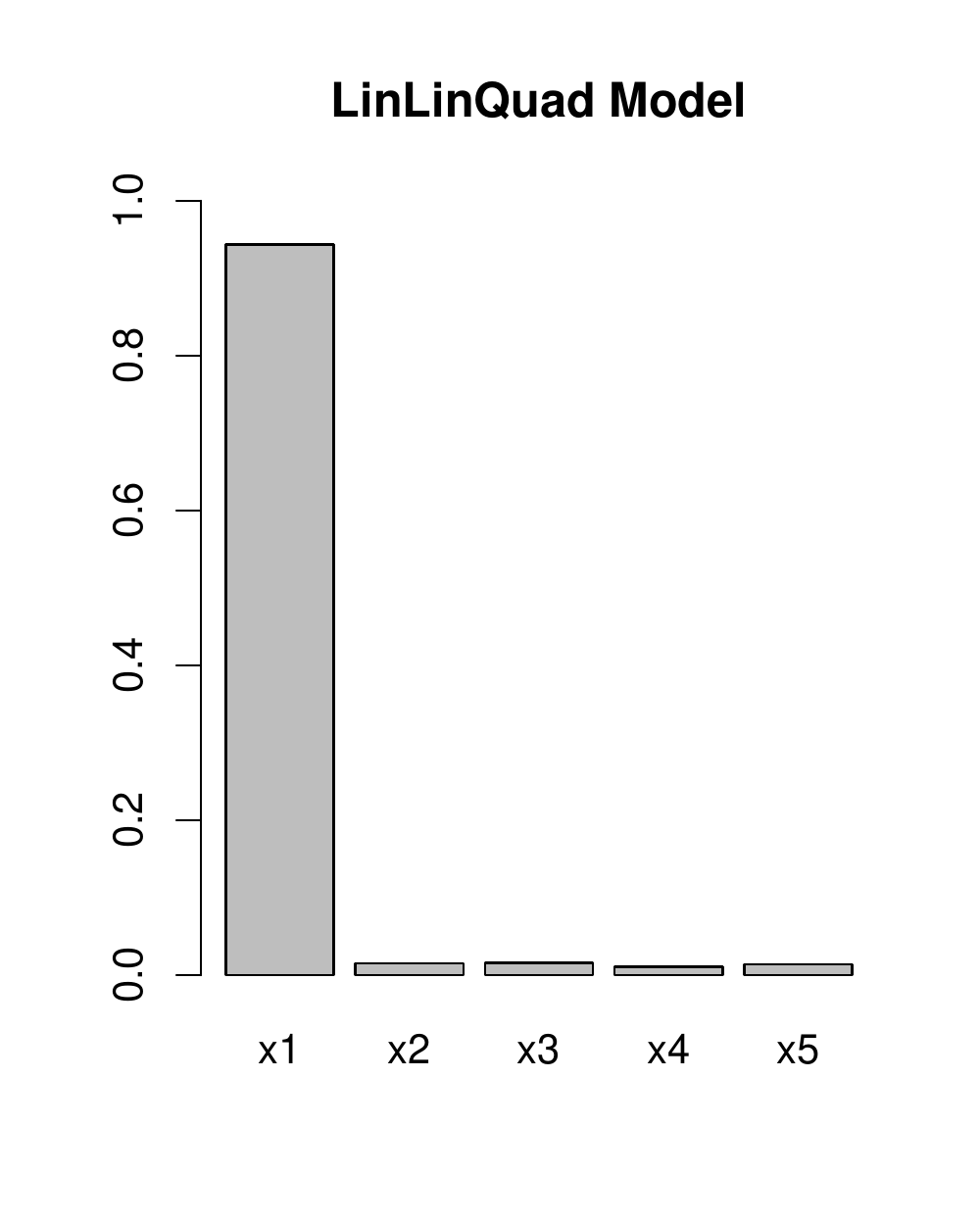}}
\caption[Estimated probabilities of split variable selection for multiple linear logistic regression option with bias correction.]{Estimated probabilities of split variable selection for multiple linear logistic regression option with bias correction (Algorithm \ref{A3:alg1}) under the models in Table \ref{T3:simu}.}\label{P3:SB4}
\end{figure}

\begin{figure}[!h]
%\begin{center}
\makebox[\textwidth][c]{\includegraphics[scale=0.5]{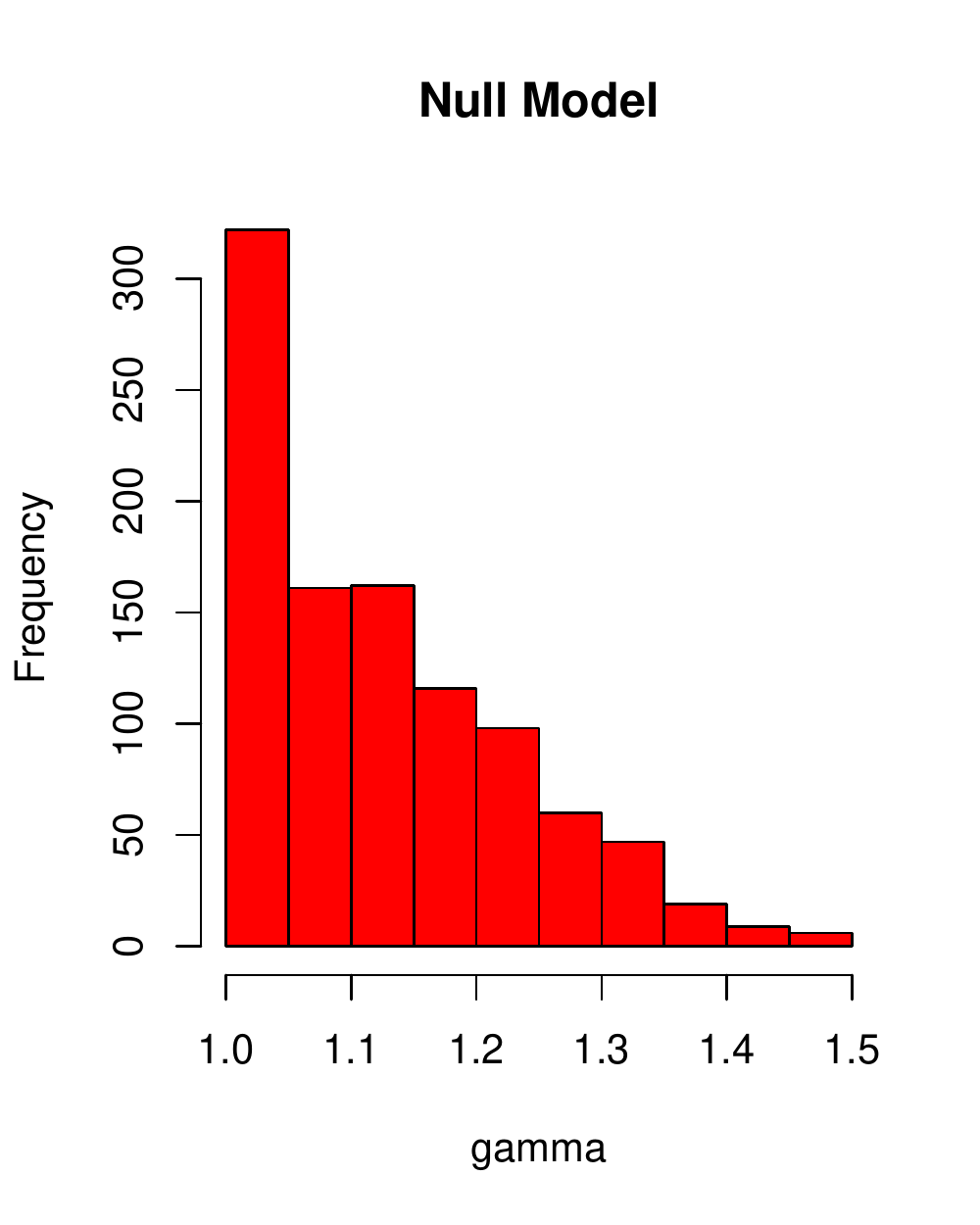}
\includegraphics[scale=0.5]{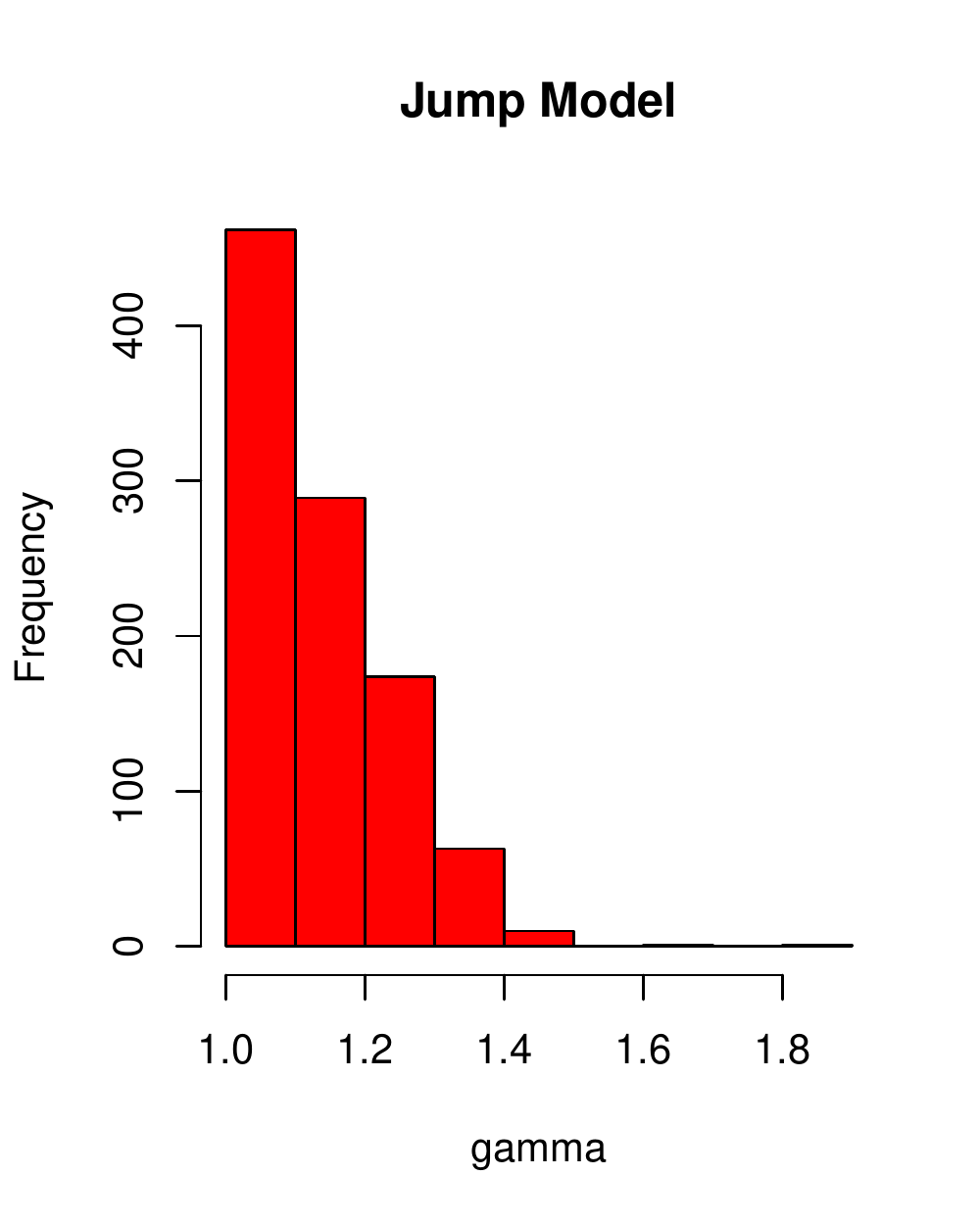}
\includegraphics[scale=0.5]{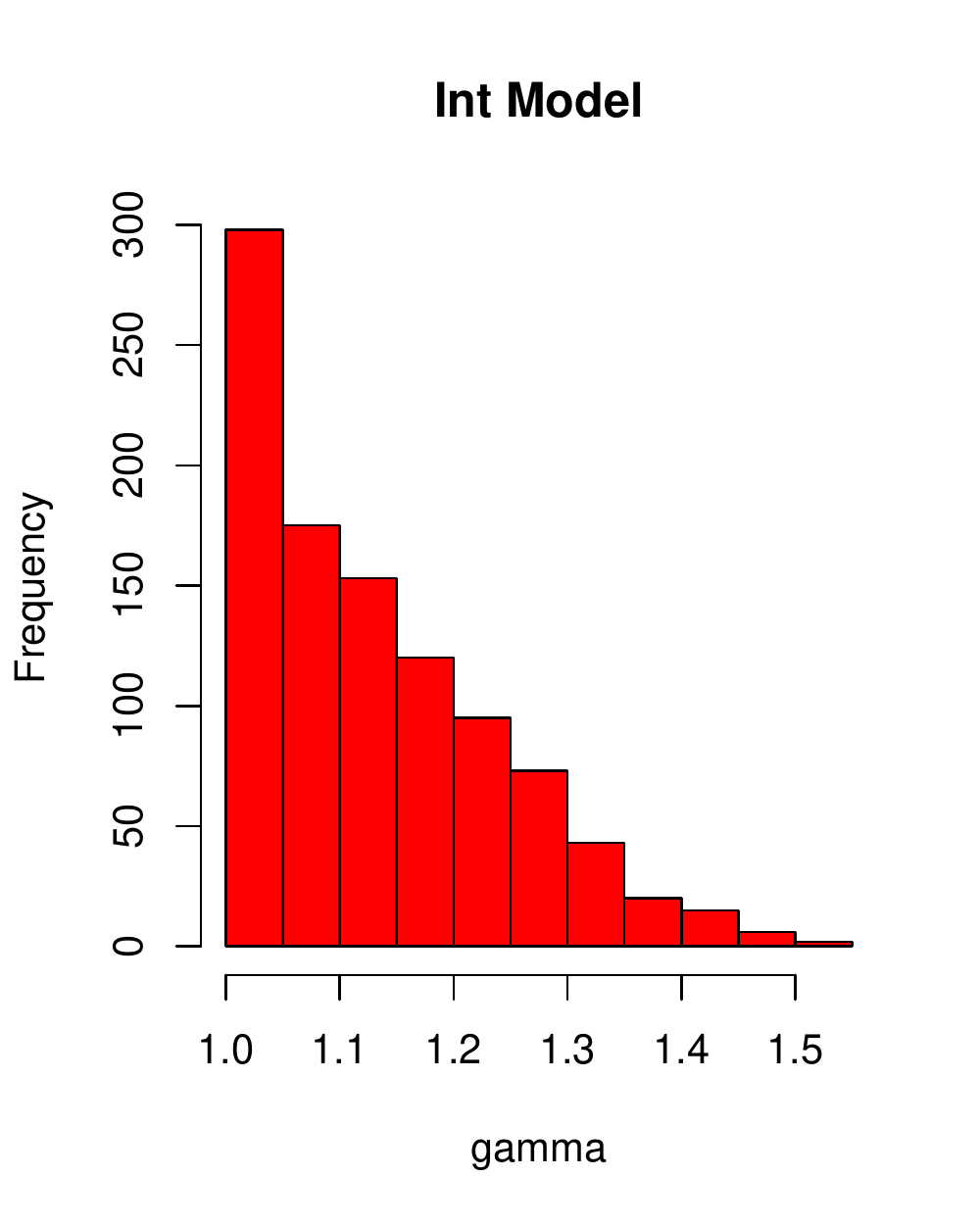}}
\makebox[\textwidth][c]{\includegraphics[scale=0.5]{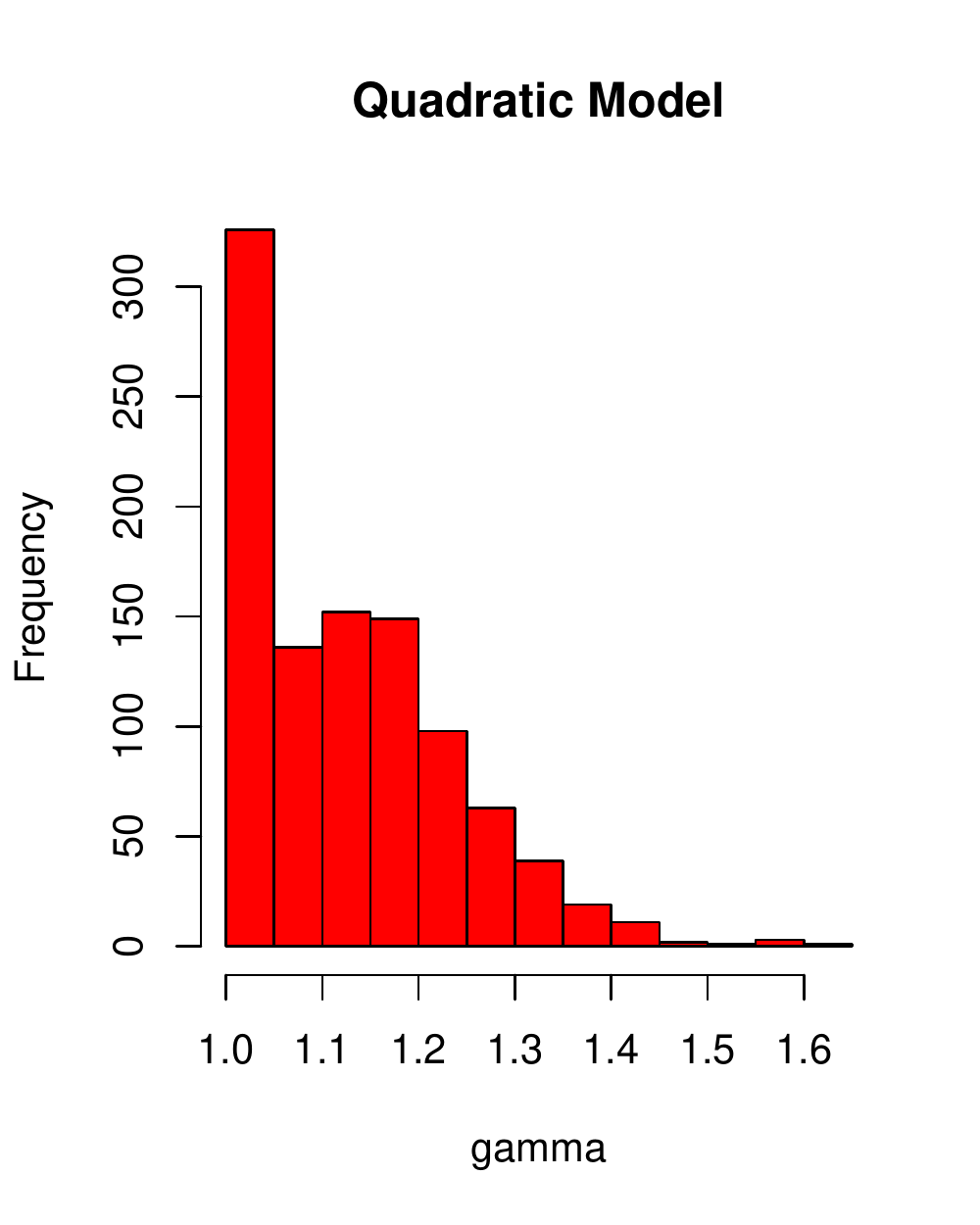}
\includegraphics[scale=0.5]{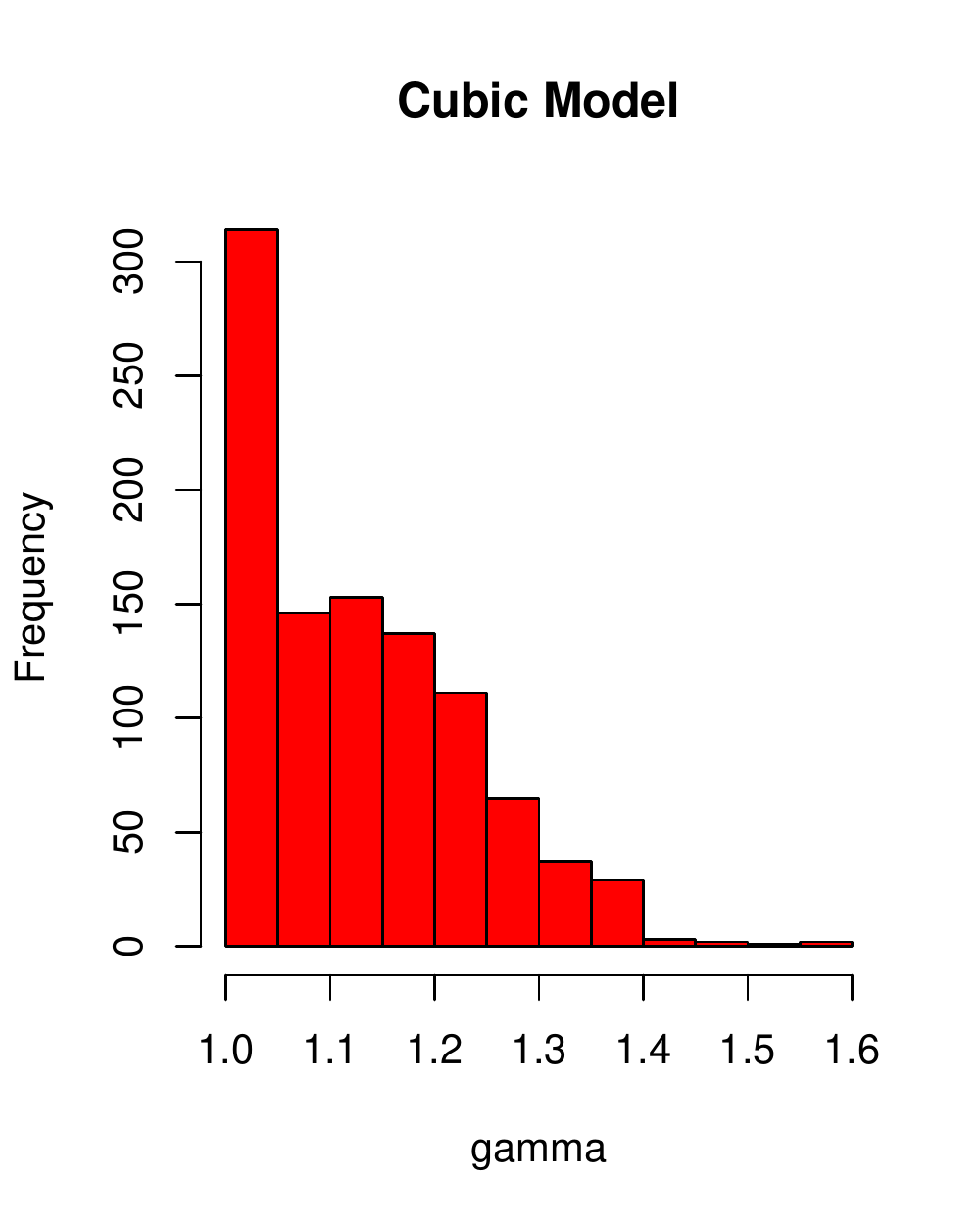}
\includegraphics[scale=0.5]{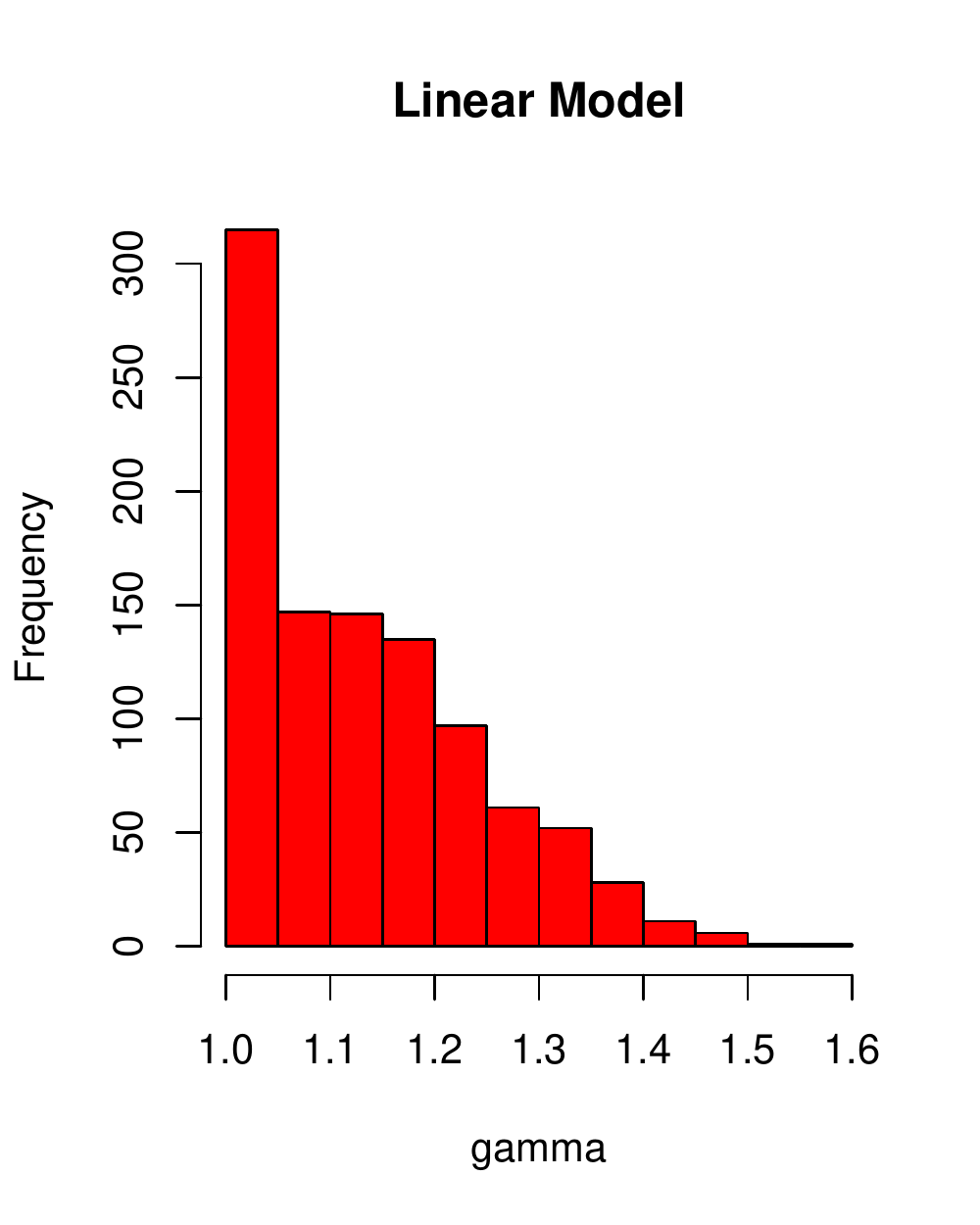}}
\makebox[\textwidth][c]{\includegraphics[scale=0.5]{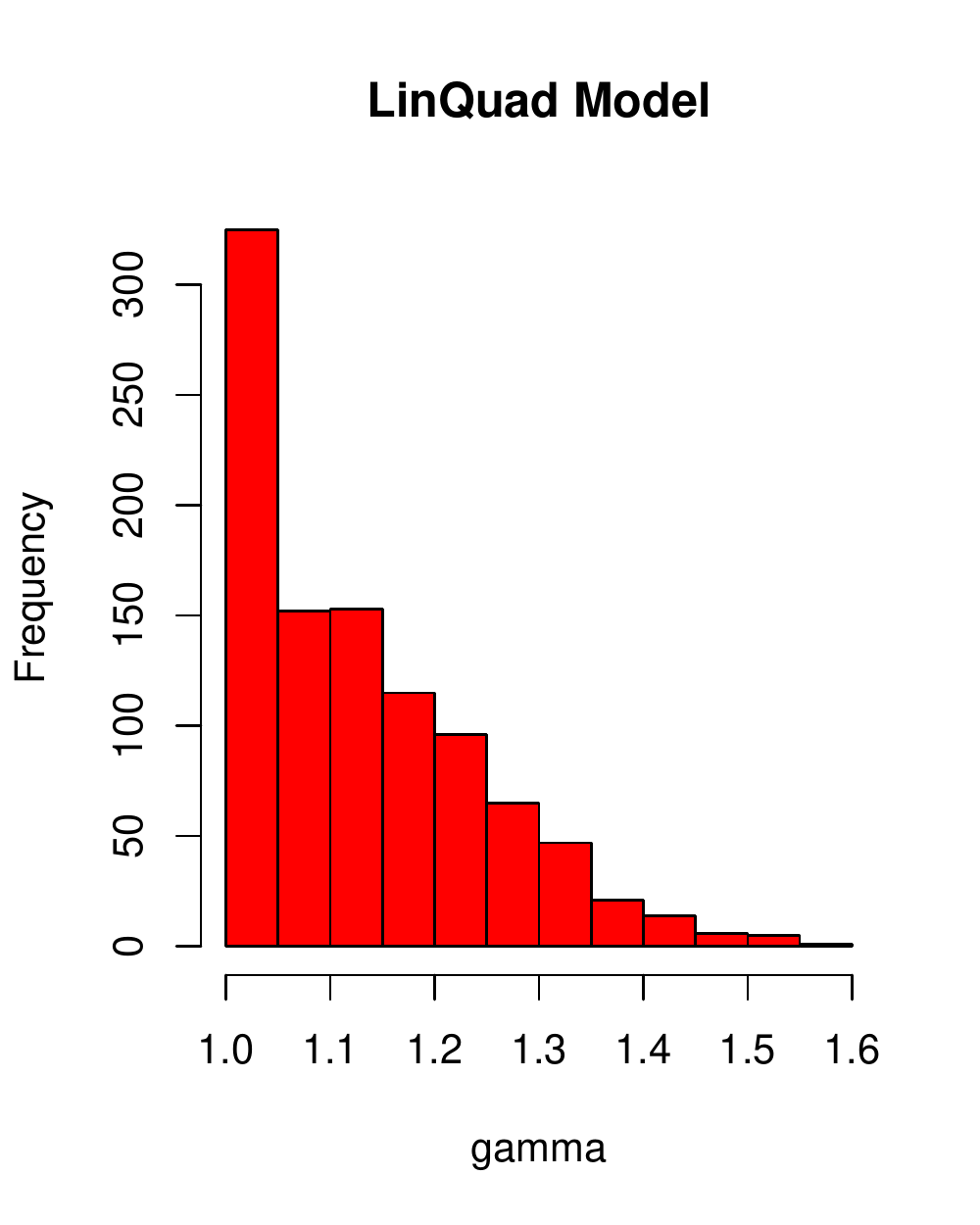}
\includegraphics[scale=0.5]{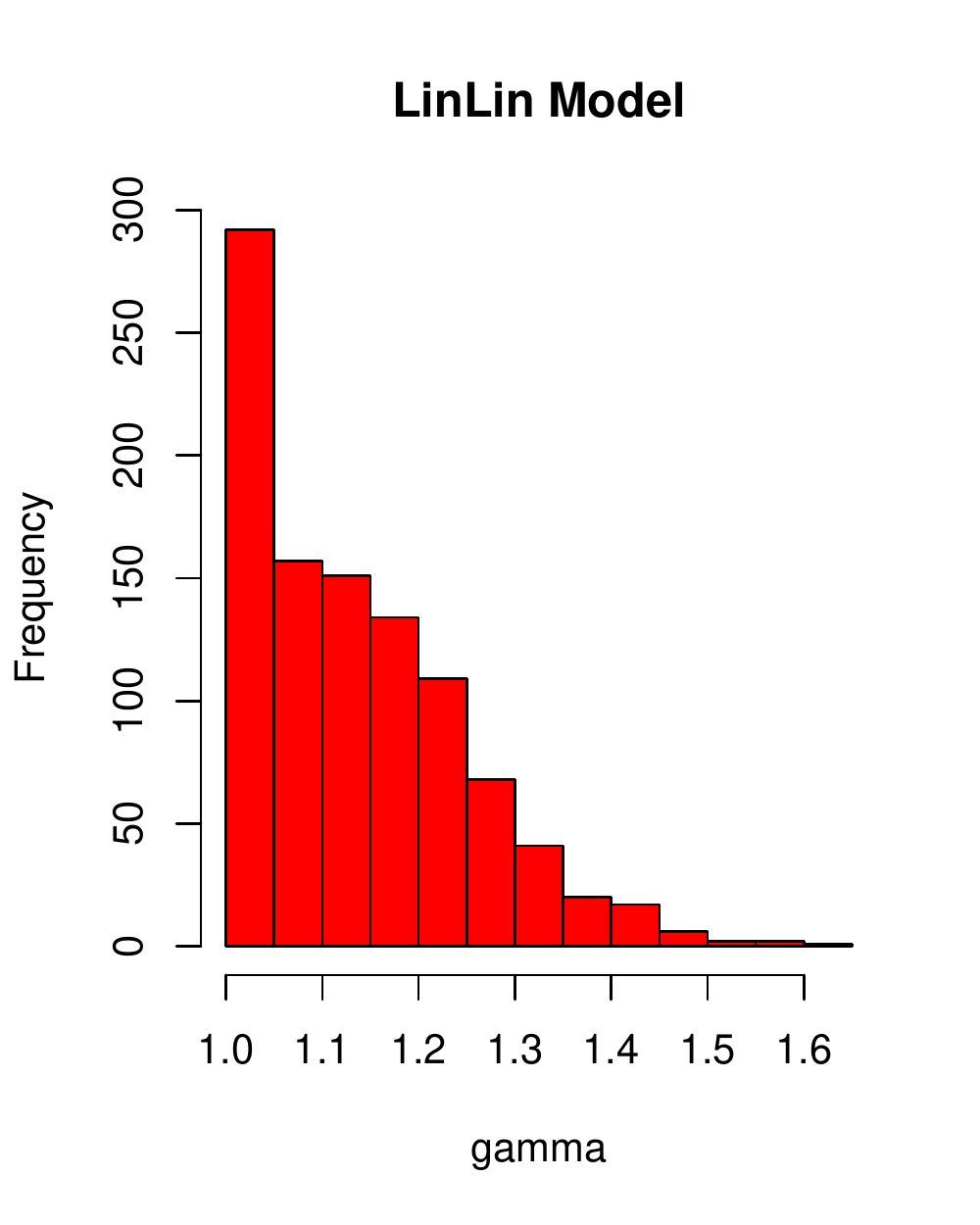}
\includegraphics[scale=0.5]{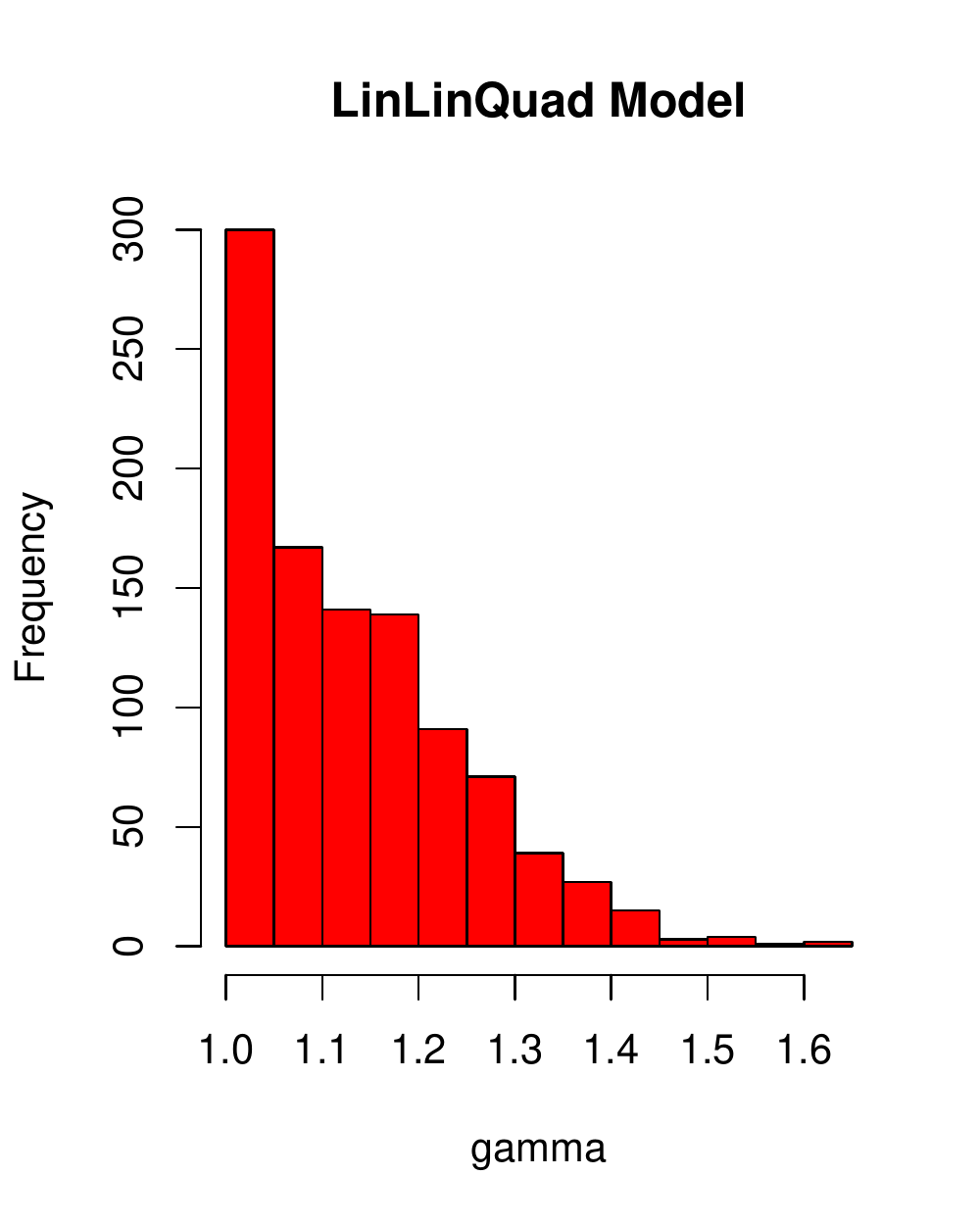}}
\caption{Simulated histograms of the multiplier $\gamma^*$ for bias correction.}\label{P3:SB5}
%\end{center}
\end{figure}

A third simulation experiment is carried out to evaluate the effect of our bootstrap bias correction algorithm for the multiple linear logistic regression option. The results are shown in Figure \ref{P3:SB4}. We use Algorithm \ref{A3:alg1} with 100 bootstrap replications and an equal-spaced grid of 1000 values of $\gamma$ in the interval $[1,\ 2]$. Compared with the selection probabilities in Figure \ref{P3:SB3}, we can see that in general the tendency of selecting categorical split variables is reduced. Figure \ref{P3:SB5} displays histograms of the multiplier $\gamma^*$ computed by Algorithm \ref{A3:alg1}. We observe that the selected value of $\gamma$ is almost always less than $1.5$.

\clearpage
\subsubsection{Summary}
The estimated selection probabilities of each predictor variable from the above three simulation experiments are summarized in Table \ref{T3:rslt}.

\begin{table}[!htb]
%\begin{center}
%\captionsetup{width=\textwidth}
\caption{Estimated probabilities of split variable selection of all three simulation experiments.}\label{T3:rslt}
\begin{adjustwidth}{-0.5cm}{}
%\begin{footnotesize}
\begin{small}
\begin{tabular}{ccp{0.07\linewidth}p{0.07\linewidth}p{0.07\linewidth}p{0.07\linewidth}p{0.07\linewidth}p{0.07\linewidth}p{0.07\linewidth}p{0.07\linewidth}p{0.07\linewidth}}
\hline
\hline \noalign{\smallskip}
Split &  & \multicolumn{9}{c}{Models} \\
\cline{3-11} \noalign{\smallskip}
Methods & $X_i$ & Null & Jump & Int & Quad &  Cubic  & Linear & LQ\textsuperscript{\dag} &  LL\textsuperscript{\dag} & LLQ\textsuperscript{\dag} \\
\hline
\multirow{5}{*}{Simple}
& $X_1$ & 0.181 & 0.478 & 0.715 & 0.800 & 0.340 & 0.213 & 0.000 & 0.000 & 0.005 \\
& $X_2$ & 0.190 & 0.142 & 0.078 & 0.053 & 0.193 & 0.227 & 0.000 & 0.743 &  0.584 \\
& $X_3$ & 0.202 & 0.141 & 0.057 & 0.047 & 0.158 & 0.201 & 1.000 & 0.257 &  0.411 \\
& $X_4$ & 0.191 & 0.112 & 0.069 & 0.050 & 0.151 & 0.157 & 0.000 & 0.000 &  0.000 \\
& $X_5$ & 0.236 & 0.127 & 0.081 & 0.050 & 0.158 & 0.202 & 0.000 & 0.000 &  0.000 \\
&&&&&&&&&\\
\multirow{5}{*}{Multiple}
& $X_1$ & 0.169 & 0.458 & 0.708 & 0.805 & 0.353 & 0.175 & 0.000 & 0.166 & 0.936 \\
& $X_2$ & 0.202 & 0.142 & 0.074 & 0.049 & 0.169 & 0.252 & 0.000 & 0.218 & 0.013 \\
& $X_3$ & 0.196 & 0.122 & 0.053 & 0.045 & 0.143 & 0.161 & 1.000 & 0.197 & 0.016 \\
& $X_4$ & 0.184 & 0.108 & 0.060 & 0.047 & 0.141 & 0.155 & 0.000 & 0.143 & 0.012 \\
& $X_5$ & 0.249 & 0.170 & 0.105 & 0.054 & 0.194 & 0.257 & 0.000 & 0.276 & 0.023 \\
&&&&&&&&&\\
\multirow{2}{*}{Multiple}
& $X_1$ & 0.184 & 0.477 & 0.729 & 0.815 & 0.357 & 0.185 & 0.000 & 0.169 & 0.944\\
& $X_2$ & 0.218 & 0.146 & 0.073 & 0.054 & 0.182 & 0.257 & 0.000 &  0.230 & 0.015\\
With
& $X_3$ & 0.206 & 0.130 & 0.057 & 0.047 & 0.147 & 0.170 &  1.000 &  0.205 & 0.016\\
\multirow{2}{*}{BBC}
& $X_4$ & 0.194 & 0.113 & 0.060 & 0.044 & 0.153 & 0.164 & 0.000 & 0.155 & 0.011\\
& $X_5$ & 0.198 & 0.134 & 0.081 & 0.040 & 0.161 & 0.224 & 0.000 & 0.241 & 0.014\\
\hline
& & \multicolumn{9}{l}{\textsuperscript{\dag}\footnotesize{LQ: LinQuad; LL: LinLin; LLQ: LinLinQuad.}}\\
&&&&&&&&&\\
\cline{2-11} \noalign{\smallskip}
 & \multicolumn{10}{c}{$\hat{p}$} \\
\cline{2-11} \noalign{\smallskip}
&  0.05 & \ 0.10 & \ 0.15 & \ 0.20 & \ 0.25 & \ 0.30 & \ 0.35 & \ 0.40 & \ 0.45 & \ 0.50 \\
& (0.95) & (0.90) & (0.85) & (0.80) & (0.75) & (0.70) & (0.65) & (0.60) & (0.55) & \\
\hline \noalign{\smallskip}
SE & \ 0.007 & \ 0.009 & \ 0.011 & \ 0.013 & \ 0.014 & \ 0.014 & \ 0.015 & \ 0.015 & \ 0.016 & \ 0.016 \\
\hline
\end{tabular}
%\end{center}
\end{small}
\end{adjustwidth}
\end{table}

%%%%%%%%%%%%%%%%%%%%%%%%%%%%%%%%%%%%%%%%%%%%%%%%%%%%%%%%%%%%%%%%%%%%%%%%%%%%%%%%
\section{Prediction Accuracy}
A question that frequently arises is how accurately does an algorithm predict in practice. To evaluate the prediction ability of PLUTO, we apply it and several competing algorithms on various real datasets using 10-fold cross-validation. This section presents the settings and results of the comparison.

\subsection{Model description} \label{sec:ch4:model}
Both the simple linear logistic regression option (Algorithm \ref{A2:alg2}) and multiple linear logistic regression option (Algorithm \ref{A2:alg3}) of PLUTO are included in the comparison. We refer to them as ``PLUTO\_S" and ``PLUTO\_M" respectively. The bias correction algorithm introduced in Section \ref{sec3} can be implemented in ``PLUTO\_S" and ``PLUTO\_M". This yields two new models: ``PLUTO\_S\_BC" and ``PLUTO\_M\_BC", where ``BC" stands for bias correction. If the $\theta\mbox{-SE}$ rule is applied while pruning the tree, a tail ``\_$\theta\mbox{SE}$" will be appended to the model name. For example, ``PLUTO\_S\_1SE" stands for the PLUTO simple linear tree without bias correction pruned with 1-SE rule. Theoretically, $\theta$ can take any non-negative value, but we only compare the common 0-SE, 0.5-SE and 1-SE rules in this study. Altogether, we include 12 different settings of PLUTO in the comparison.

Other algorithms include:
\begin{description}
  \item[MOB:] The Model-Based Recursive Partitioning based on multiple linear logistic regression models. It is implemented in R package ``partykit" \citep{partykit}. Default values for the arguments are used.
  \item[LOTUS:] LOTUS with its default settings. It fits a multiple linear logistic regression model with stepwise selection at each node. 10-fold cross-validation and 0-SE rule are applied for pruning. LOTUS crashes when there is complete separation in the data.
  \item[GLMNET:] Regularized multiple linear logistic regression with lasso penalty via coordinate descent. It is implemented in R package ``glmnet" \citep{GLMNET}. 10-fold cross-validation is applied for tuning the regularization parameter. Each categorical predictor is converted to a set of dummy variables by hand-coding.
  \item[GLM:] Multiple linear logistic regression without variable selection. Each categorical predictor is converted to a set of dummy variables automatically by the program. GLM can not predict a testing sample if any categorical variable takes a value that does not exist in the training sample.
\end{description}

For the tree algorithms, all predictors, categorical or numerical, serve as split variable candidates. On the other hand, only numerical predictors are used for node model fitting.

When an algorithm crashes due to its own internal flaw, we take the trivial estimate, the proportion of training samples that have outcomes equal to 1, to predict the testing sample.
%In this way, algorithms get penalized for not being compatible.

\subsection{Comparison measures}
We estimate the probability of success for each observation via 10-fold cross-validation and denote the fitted value by $\hat{p}_i$, where $i$ is an index of observations. Then the following measures are computed:
\begin{description}
  \item[Ratio of Deviance:] Although deviance is commonly used for comparing generalized linear models, it is not scale free. Instead, we take the ratio of deviance to the maximum deviance among all algorithms as a measure, which falls in the range $[0,1]$. Same as deviance, an algorithm with a smaller value of the Ratio of Deviance is preferred. Deviance can be computed by Equation \ref{E2:dev}. However, when $\hat{p}_i = 0,\ y_i = 1$, deviance will be ``infinity", no matter how accurate the rest of the observations are predicted. To remedy this problem, we adopt a trimmed version of deviance and denote it by $\mbox{DEV}'$.
\begin{equation}
\label{E4:dev}
\mbox{DEV}' = -2\sum_{i\in T} [y_i\log{(\hat{p}_i)}+(1-y_i)\log{(1-\hat{p}_i)}]
\end{equation}
where $T$ is a subset of the data with log-likelihood value $y_i\log{(\hat{p}_i)}+(1-y_i)\log{(1-\hat{p}_i)}$ below the 99th percentile.
  \item[Misclassification Error Rate:] The proportion of the data that is misclassified. It is a straightforward way to assess the prediction accuracy of classification algorithms.
  \item[AUROC:] Area Under the ROC curve. ROC stands for ``receiver operating characteristic". The ROC curve plots the true positive rate against the false positive rate of a binary classifier. AUROC measures the probability that the classifier will assign a higher probability of ``success" to a randomly selected positive sample than to a randomly selected negative sample.
\end{description}

\subsection{Data summary}
\begin{figure}[!ht]
                \makebox[\textwidth][c]{\includegraphics[scale=0.7]{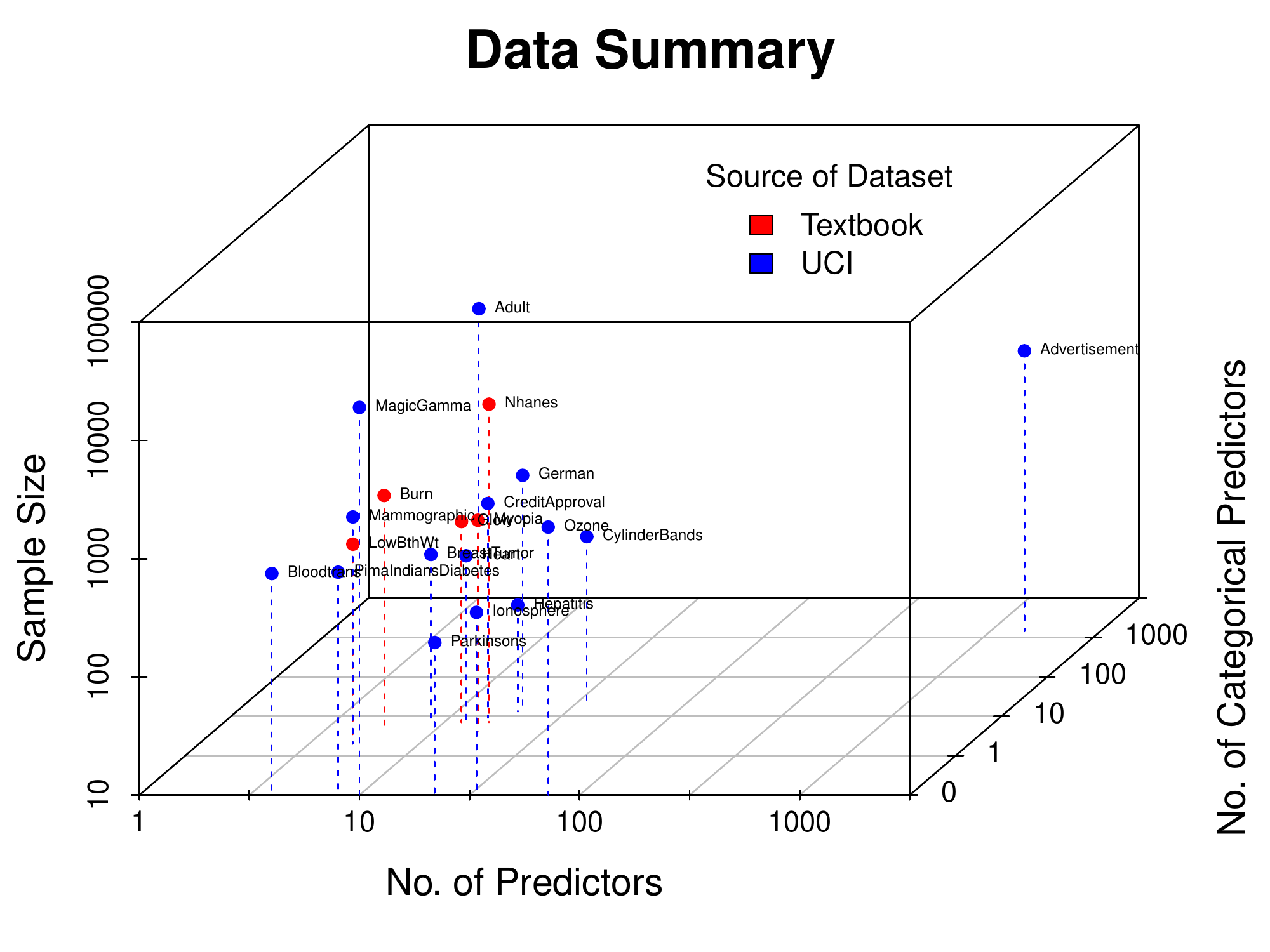}}
\caption{Summary of real datasets used for model comparison.}
\label{P4:datsum}
\end{figure}

We collected 20 datasets for this study from the UCI Machine Learning Repository \citep{uci} and textbooks \citep{hosmer2000,hosmer2013}. Table \ref{T4:datasum} shows the following features of each dataset in order: data name, sample size (\#Obs), total number of predictors (\#Prd), number of categorical predictors (\#Cat), number of numerical predictors (\#Num), whether the original data contains missing values, and the source of the data. Figure \ref{P4:datsum} shows that these datasets cover a wide range.

Some of the datasets do not contain categorical predictors, and the PLUTO bias correction algorithm is not applicable to them. For datasets that originally have missing values, we only use the set of complete observations for model fitting. The sample sizes given in Table \ref{T4:datasum} do not include observations with missing values.

\begin{table}[!htb]
%\begin{center}
\caption{Descriptions of datasets applied for model comparisons.}\label{T4:datasum}
\begin{adjustwidth}{-0.5cm}{}
\begin{small}
\begin{tabular}{lrrrrcc}
\hline\hline
\multicolumn{1}{l}{Dataset}&\multicolumn{1}{c}{\#Obs}&\multicolumn{1}{c}{\#Prd}&\multicolumn{1}{c}{\#Cat}&\multicolumn{1}{c}{\#Num}&\multicolumn{1}{c}{Missing}&\multicolumn{1}{c}{Source}\tabularnewline
\hline
Hepatitis&$   80$&$  19$&$  13$&$ 6$&Yes&UCI\tabularnewline
Parkinsons&$  195$&$  22$&$   0$&$22$&No&UCI\tabularnewline
Heart&$  270$&$  13$&$   6$&$ 7$&No&UCI\tabularnewline
BreastTumor&$  277$&$   9$&$   6$&$ 3$&Yes&UCI\tabularnewline
CylinderBands&$  277$&$  37$&$  17$&$20$&Yes&UCI\tabularnewline
Ionosphere&$  351$&$  34$&$   0$&$34$&No&UCI\tabularnewline
LowBthWt&$  488$&$   5$&$   2$&$ 3$&No& \citet{hosmer2000} \tabularnewline
Glow&$  500$&$  12$&$   7$&$ 5$&No& \citet{hosmer2013} \tabularnewline
Myopia&$  618$&$  16$&$   4$&$12$&No& \citet{hosmer2013} \tabularnewline
CreditApproval&$  653$&$  15$&$   9$&$ 6$&Yes&UCI\tabularnewline
Bloodtrans&$  748$&$   4$&$   0$&$ 4$&No&UCI\tabularnewline
PimaIndiansDiabetes&$  768$&$   8$&$   0$&$ 8$&No&UCI\tabularnewline
Mammographic&$  830$&$   5$&$   2$&$ 3$&Yes&UCI\tabularnewline
German&$ 1000$&$  20$&$  13$&$ 7$&No&UCI\tabularnewline
Burn&$ 1000$&$   6$&$   4$&$ 2$&No&\citet{hosmer2013}\tabularnewline
Ozone&$ 1848$&$  72$&$   0$&$72$&Yes&UCI\tabularnewline
Advertisement&$ 2359$&$1430$&$1427$&$ 3$&Yes&UCI\tabularnewline
Nhanes&$ 4915$&$  16$&$   7$&$ 9$&No&\citet{hosmer2013}\tabularnewline
MagicGamma&$19020$&$  10$&$   0$&$10$&No&UCI\tabularnewline
Adult&$30162$&$  14$&$   8$&$ 6$&Yes&UCI\tabularnewline
\hline
\end{tabular}
%\end{center}
\end{small}
\end{adjustwidth}
\end{table}

\subsection{Results}
\subsubsection{Model comparison on individual datasets}
\begin{figure}[p]
%\begin{adjustwidth}{-0.5cm}{}
\centering
                \includegraphics[width=\textwidth]{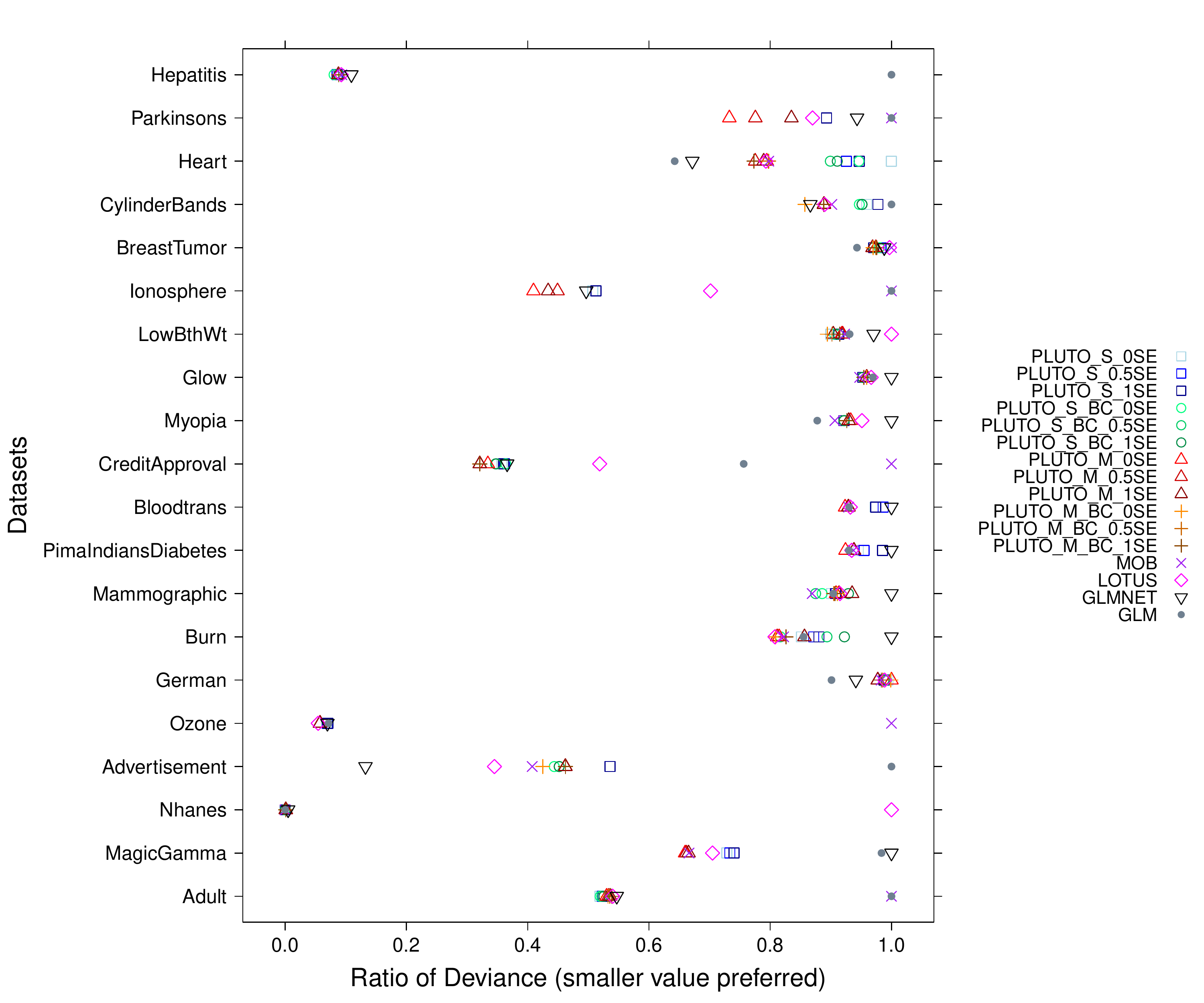}
%\end{adjustwidth}
\caption[Ratio of deviance of each algorithm for all datasets.]{Ratio of deviance of each algorithm for all datasets.}
\label{P4:dev}
\end{figure}

\begin{figure}[p]
%\begin{adjustwidth}{-0.5cm}{}
\centering
                \includegraphics[width=\textwidth]{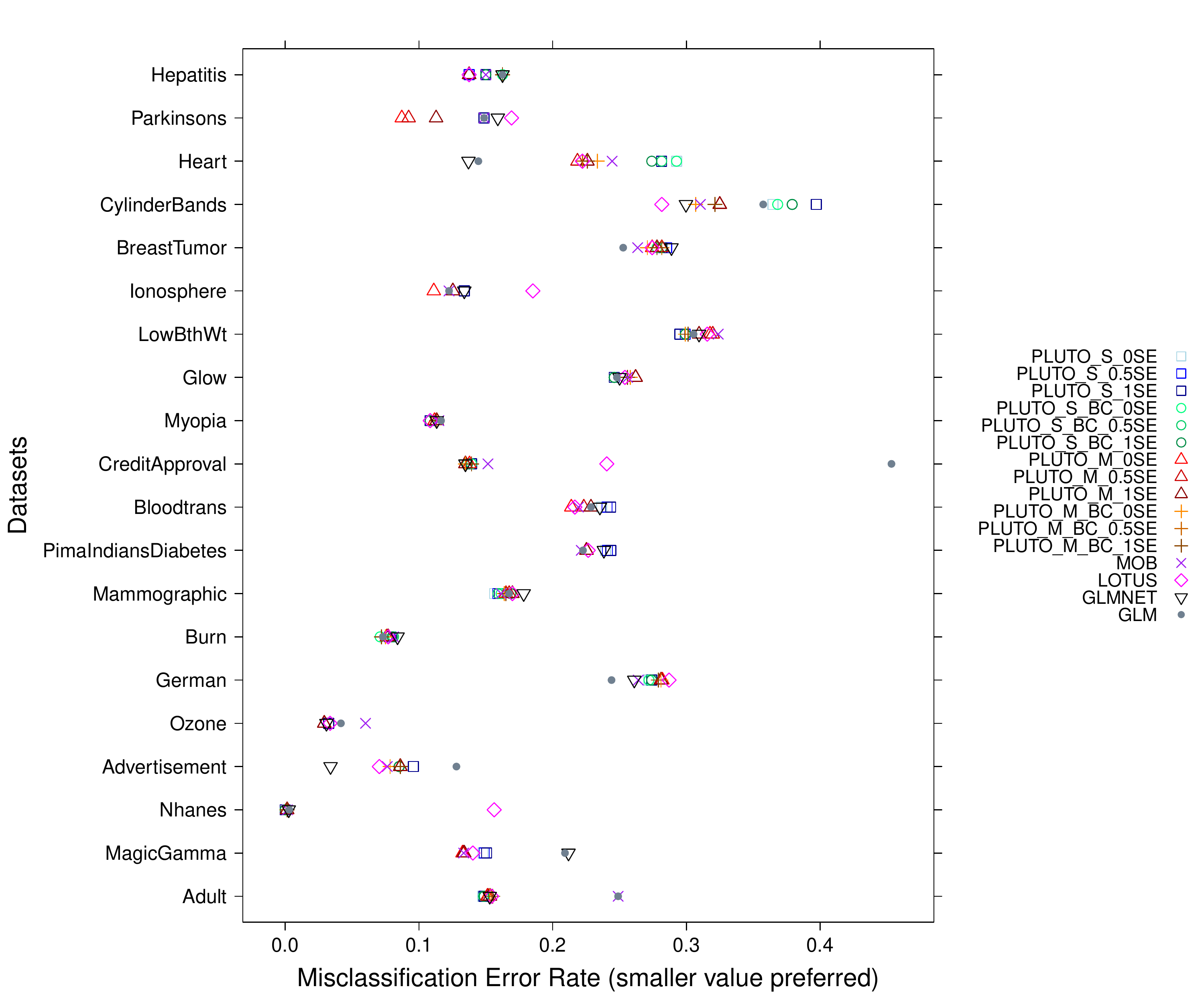}
%\end{adjustwidth}
\caption[Misclassification error rate of each algorithm for all datasets.]{Misclassification error rate of each algorithm for all datasets.}
\label{P4:misc}
\end{figure}

\begin{figure}[p]
%\begin{adjustwidth}{-0.5cm}{}
\centering
                \includegraphics[width=\textwidth]{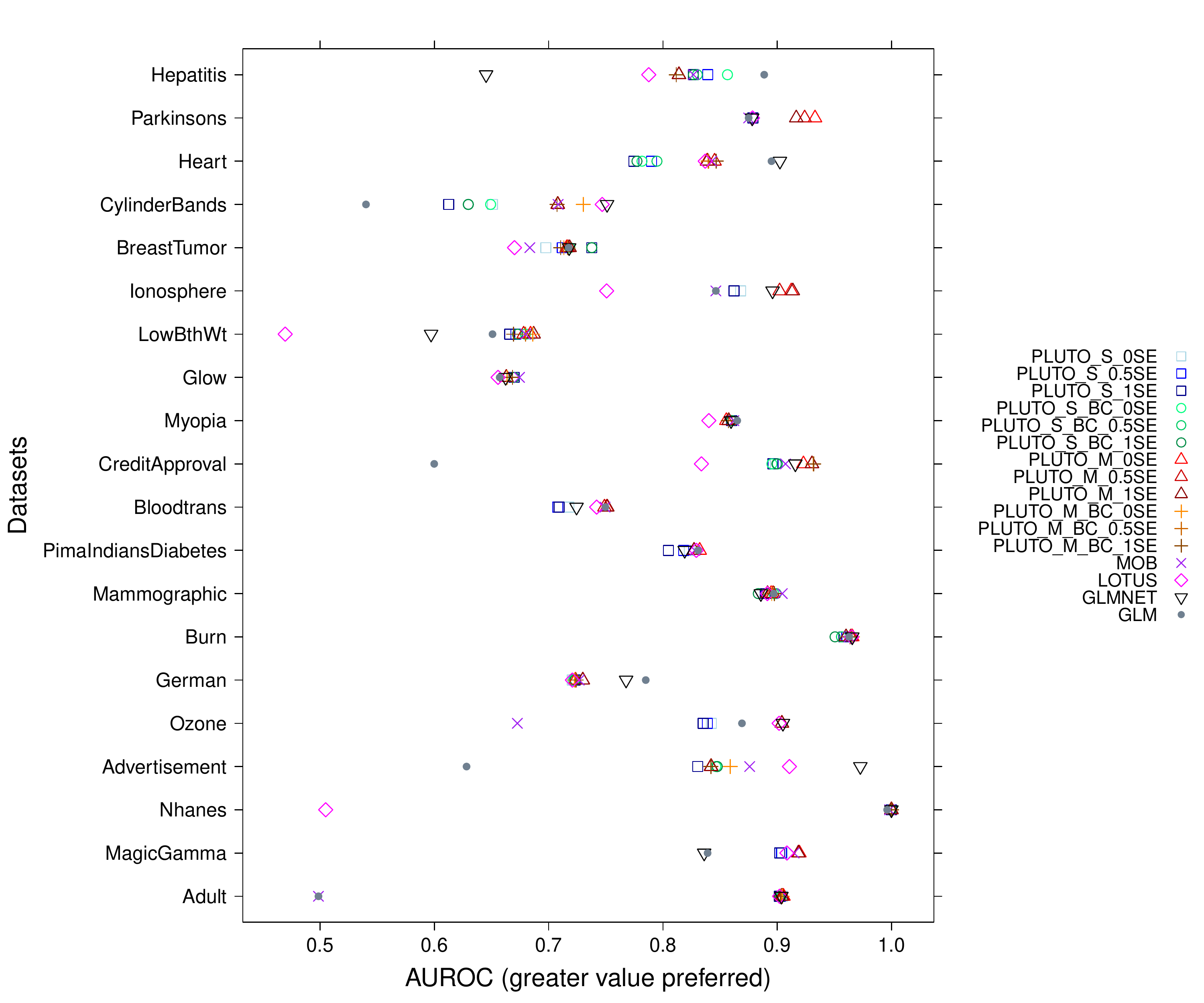}
%\end{adjustwidth}
\caption[AUROC of each algorithm for all datasets.]{AUROC of each algorithm for all datasets.}
\label{P4:auc}
\end{figure}

The estimated ratio of deviance, misclassification error rate and AUROC of each algorithm are displayed in Figures \ref{P4:dev}, \ref{P4:misc} and \ref{P4:auc} respectively.

The scatter plots show that none of the algorithms always performs better than the others. Besides, we can see that:
\begin{enumerate}
\item The multiple linear PLUTO models perform well against other algorithms. For a majority of the datasets, their prediction accuracies are above average. They also dominate other algorithms on $30\%$ to $40\%$ of the datasets, depending on the measures.
\item The simple linear PLUTO models do not perform as well as the multiple linear PLUTO models in general, and can fall in the bottom half at times. However, they provide acceptable prediction accuracies on most datasets, and sometimes even the best.
\item The accuracies of MOB, LOTUS, GLMNET and GLM are hard to determine from the scatter plots since each of them has highs and lows.
\item MOB crashes on the dataset ``Adult". LOTUS crashes on ``LowBthWt" and ``Nhanes". GLM crashes on ``CylinderBands", ``CreditApproval", ``Advertisement" and ``Adult". As explained in section \ref{sec:ch4:model}, these algorithms get penalized, and the penalty is reflected in their scores.
\end{enumerate}

\subsubsection{Model comparison on average}
Since the six PLUTO models with bias correction are not applicable to datasets without categorical variables, we first omit them from the comparison. Next, we take the arithmetic mean of each measure for the remaining twelve algorithms over all datasets. The results are given in Table \ref{T4:rslts.allmtd1}. One may question the penalty given to the algorithms that crash on certain datasets. To remedy this problem, we also provide means over the fourteen datasets where no crash occurs, which are presented in Table \ref{T4:rslts.allmtd2}.

\begin{table}[!htb]
\begin{center}
\caption{Arithmetic means of 10-fold cross-validation estimates of the ratio of deviance, misclassification error rate, and AUROC for all methods except PLUTO algorithms with bias correction, over all 20 datasets.}\label{T4:rslts.allmtd1}
\begin{tabular}{lp{0.15\linewidth}p{0.15\linewidth}l}
  \hline\hline
Model & RoD\textsuperscript{\dag} & MER\textsuperscript{\dag} & AUROC \\
  \hline
PLUTO\_S\_0SE & 0.704 & 0.178 & 0.821 \\
  PLUTO\_S\_0.5SE & 0.705 & 0.180 & 0.818 \\
  PLUTO\_S\_1SE & 0.707 & 0.180 & 0.817 \\
  PLUTO\_M\_0SE & 0.665 & 0.168 & 0.839 \\
  PLUTO\_M\_0.5SE & 0.668 & 0.169 & 0.839 \\
  PLUTO\_M\_1SE & 0.674 & 0.170 & 0.839 \\
  MOB & 0.809 & 0.178 & 0.802 \\
  LOTUS & 0.750 & 0.186 & 0.787 \\
  GLMNET & 0.705 & 0.171 & 0.830 \\
  GLM & 0.835 & 0.194 & 0.780 \\
   \hline
   \multicolumn{4}{l}{\textsuperscript{\dag}\footnotesize{RoD: Ratio of Deviance. MER: Misclassification Error Rate.}}
\end{tabular}
\end{center}
\end{table}

\begin{table}[!hb]
\begin{center}
\caption{Arithmetic means of 10-fold cross-validation estimates of the ratio of deviance, misclassification error rate, and AUROC for all methods except PLUTO algorithms with bias correction, over the 14 datasets that no method crashes on.}\label{T4:rslts.allmtd2}
\begin{tabular}{lp{0.15\linewidth}p{0.15\linewidth}l}
  \hline\hline
  Model & RoD\textsuperscript{\dag} & MER\textsuperscript{\dag} & AUROC \\
  \hline
  PLUTO\_S\_0SE & 0.772 & 0.179 & 0.818 \\
  PLUTO\_S\_0.5SE & 0.770 & 0.180 & 0.818 \\
  PLUTO\_S\_1SE & 0.774 & 0.181 & 0.817 \\
  PLUTO\_M\_0SE & 0.727 & 0.167 & 0.837 \\
  PLUTO\_M\_0.5SE & 0.731 & 0.169 & 0.837 \\
  PLUTO\_M\_1SE & 0.740 & 0.171 & 0.836 \\
  MOB & 0.854 & 0.174 & 0.812 \\
  LOTUS & 0.765 & 0.179 & 0.813 \\
  GLMNET & 0.801 & 0.177 & 0.819 \\
  GLM & 0.858 & 0.170 & 0.834 \\
   \hline
   \multicolumn{4}{l}{\textsuperscript{\dag}\footnotesize{RoD: Ratio of Deviance. MER: Misclassification Error Rate.}}
\end{tabular}
\end{center}
\end{table}

\begin{figure}[!ht]
                \makebox[\textwidth][c]{\includegraphics[scale=0.45]{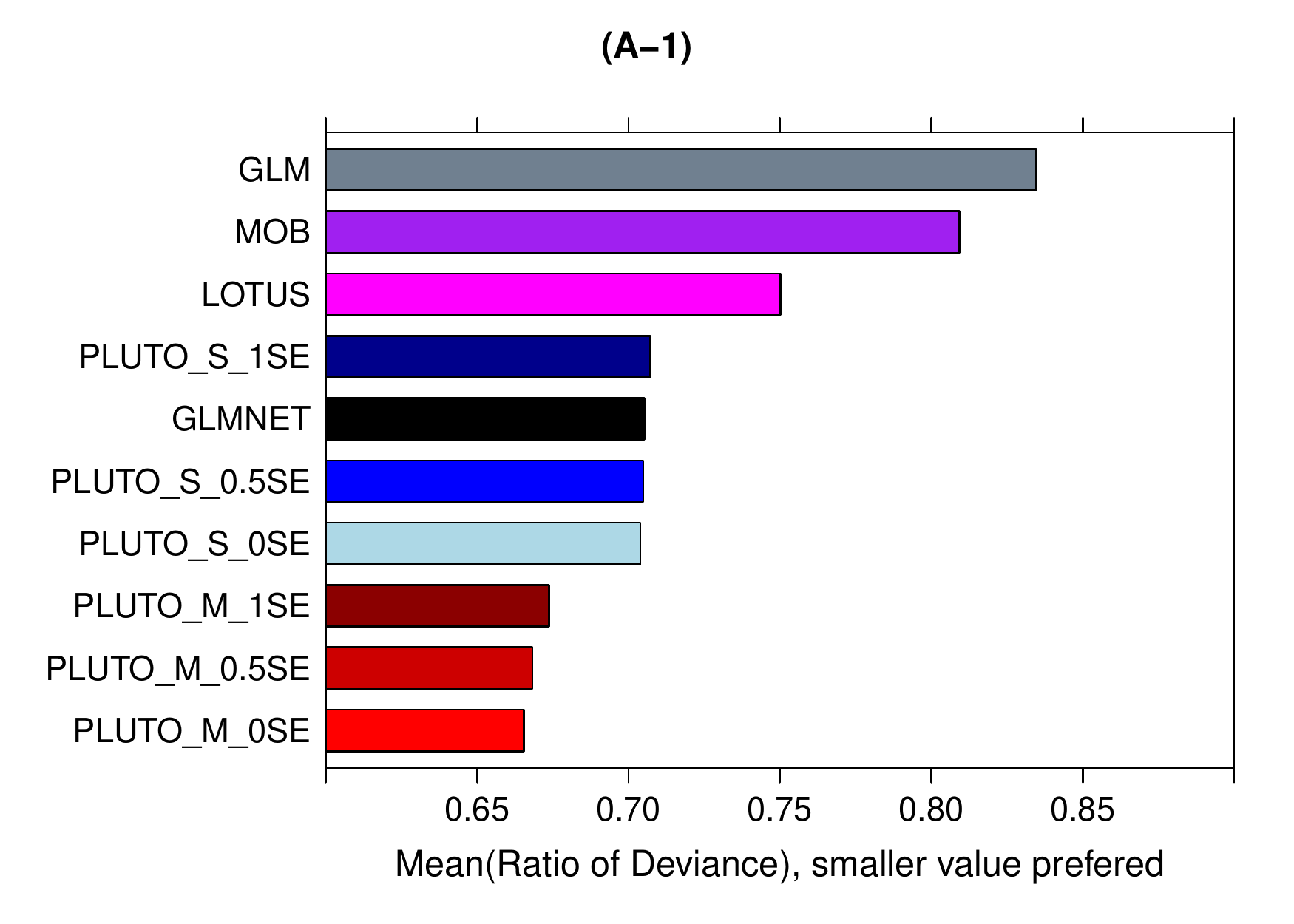}
                \includegraphics[scale=0.45]{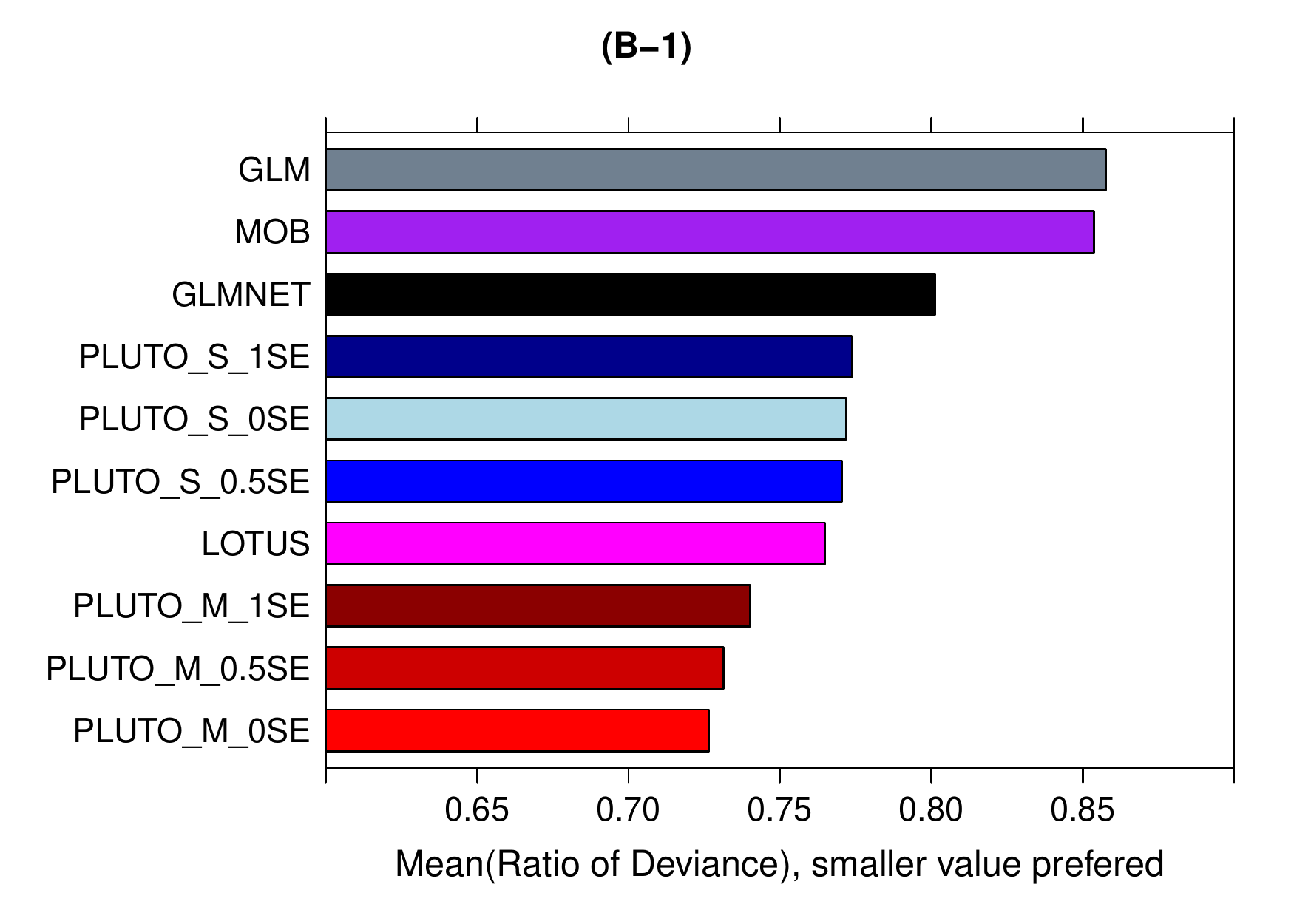}}
                \makebox[\textwidth][c]{\includegraphics[scale=0.45]{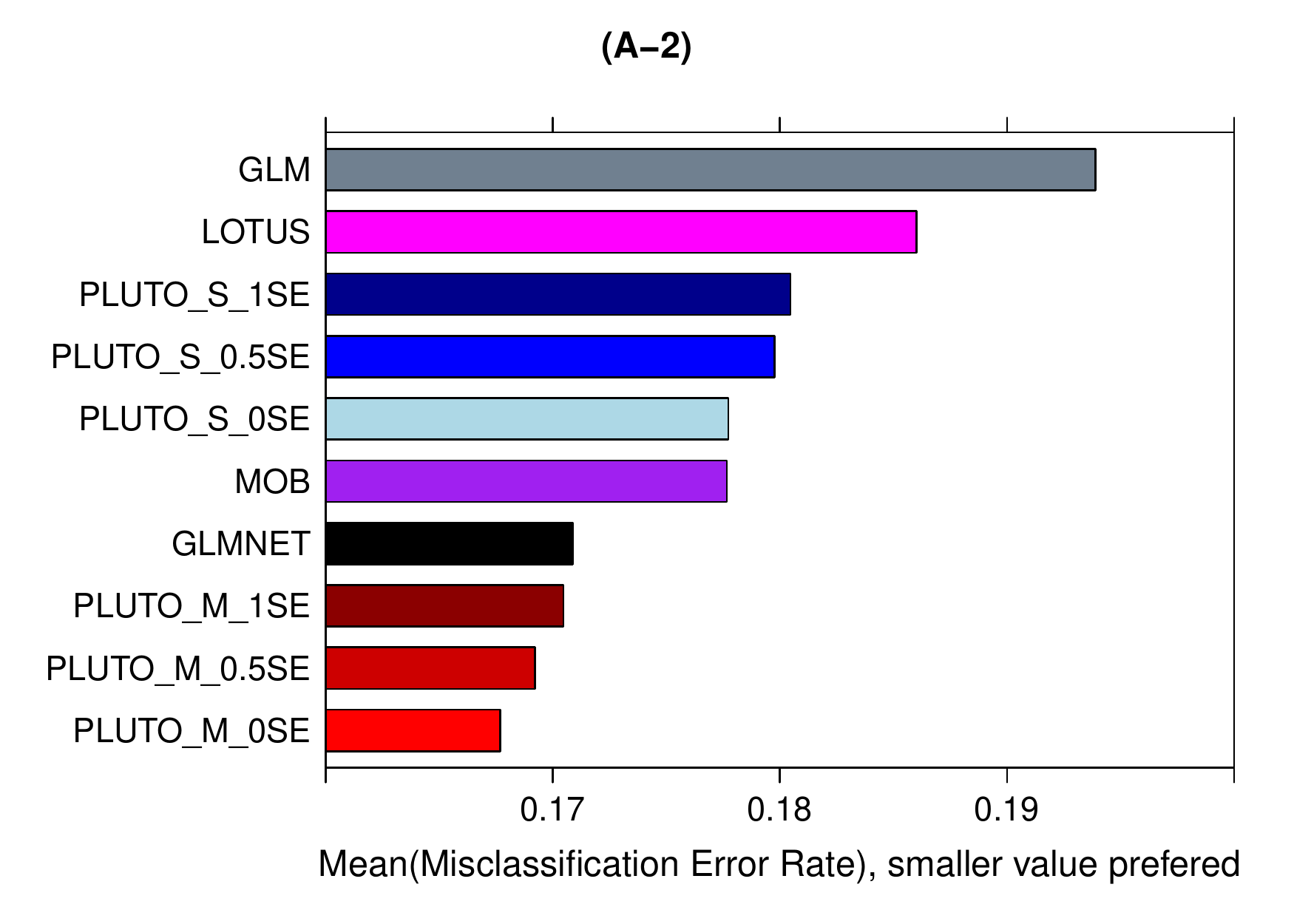}
                \includegraphics[scale=0.45]{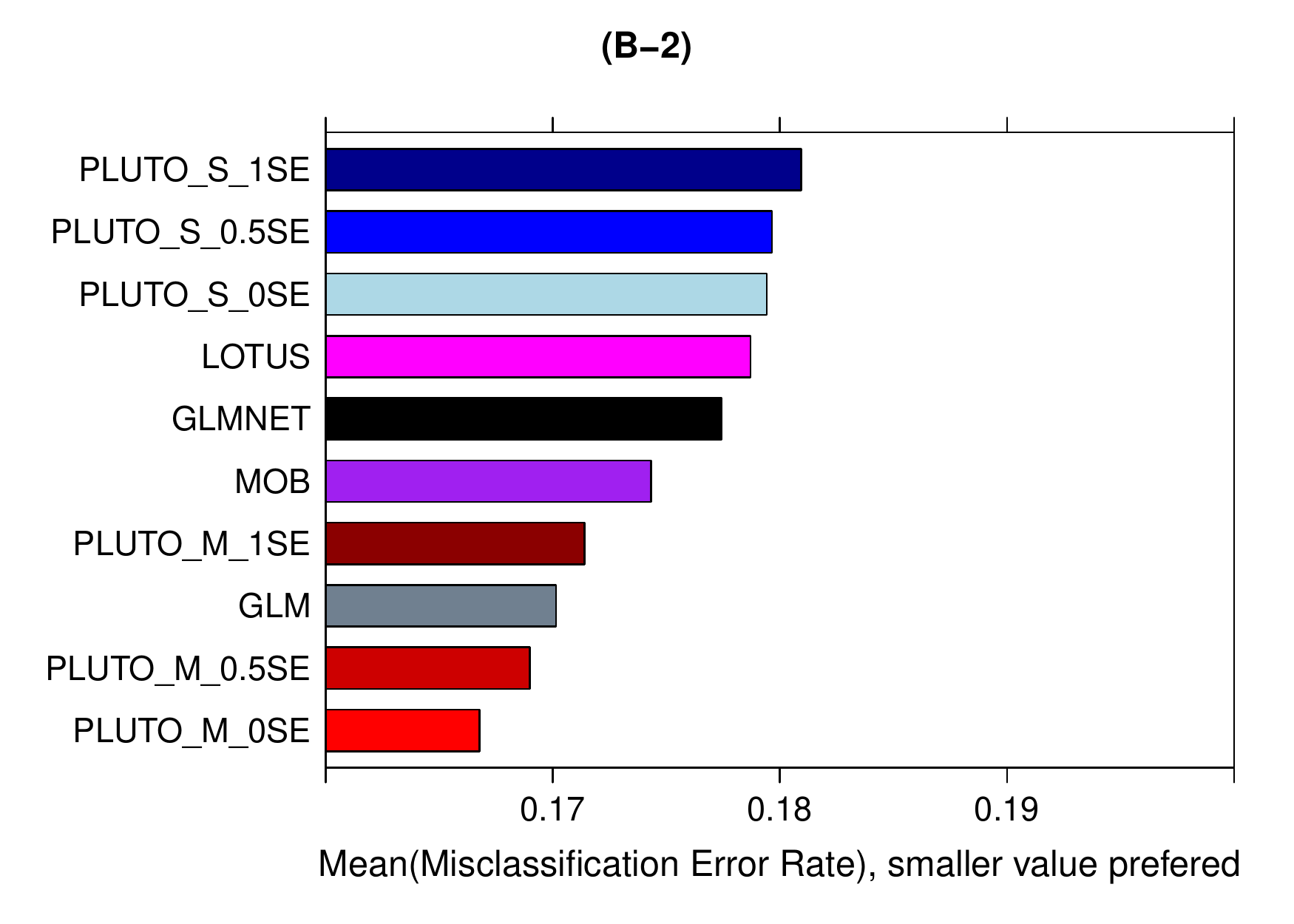}}
                \makebox[\textwidth][c]{\includegraphics[scale=0.45]{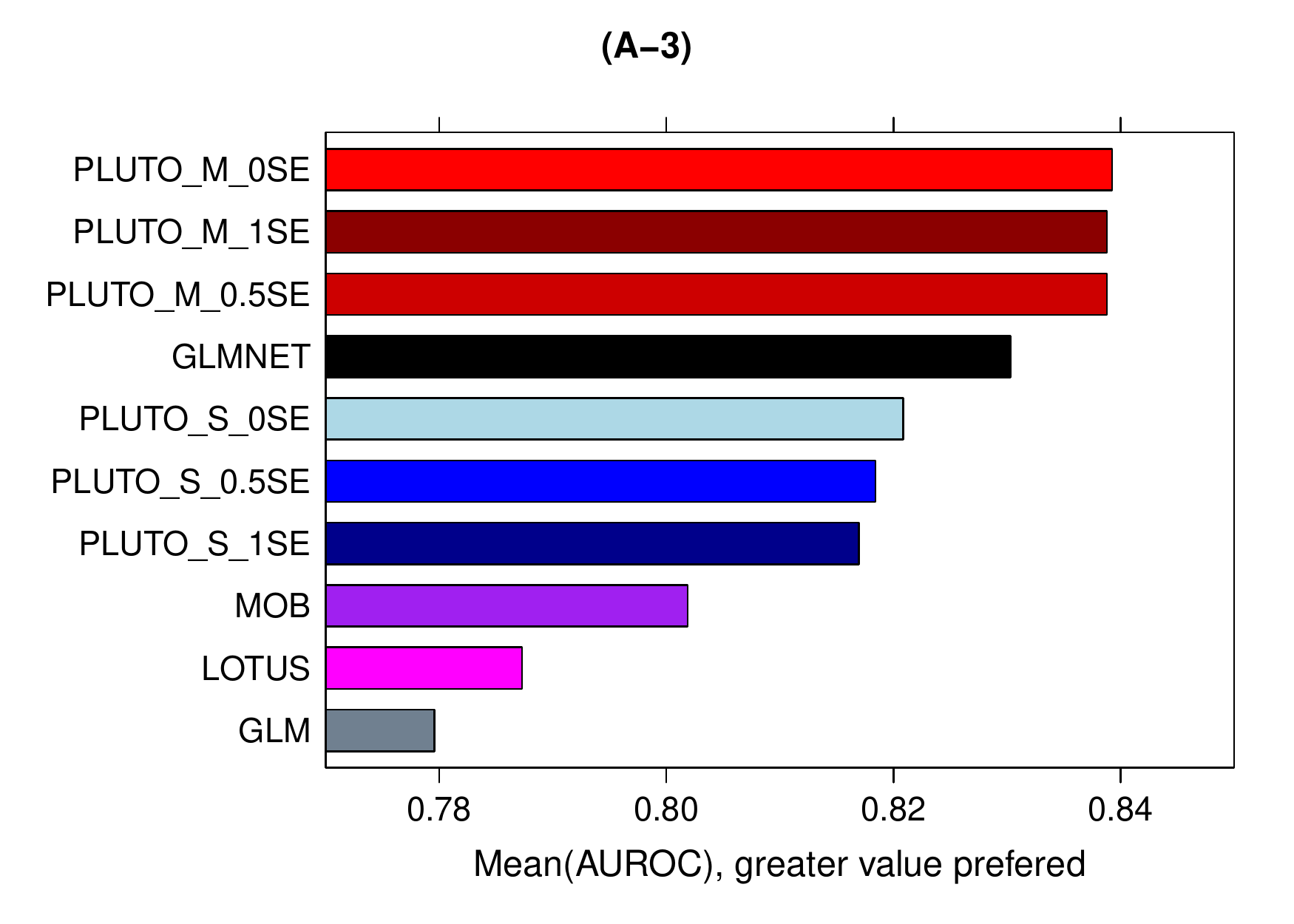}
                \includegraphics[scale=0.45]{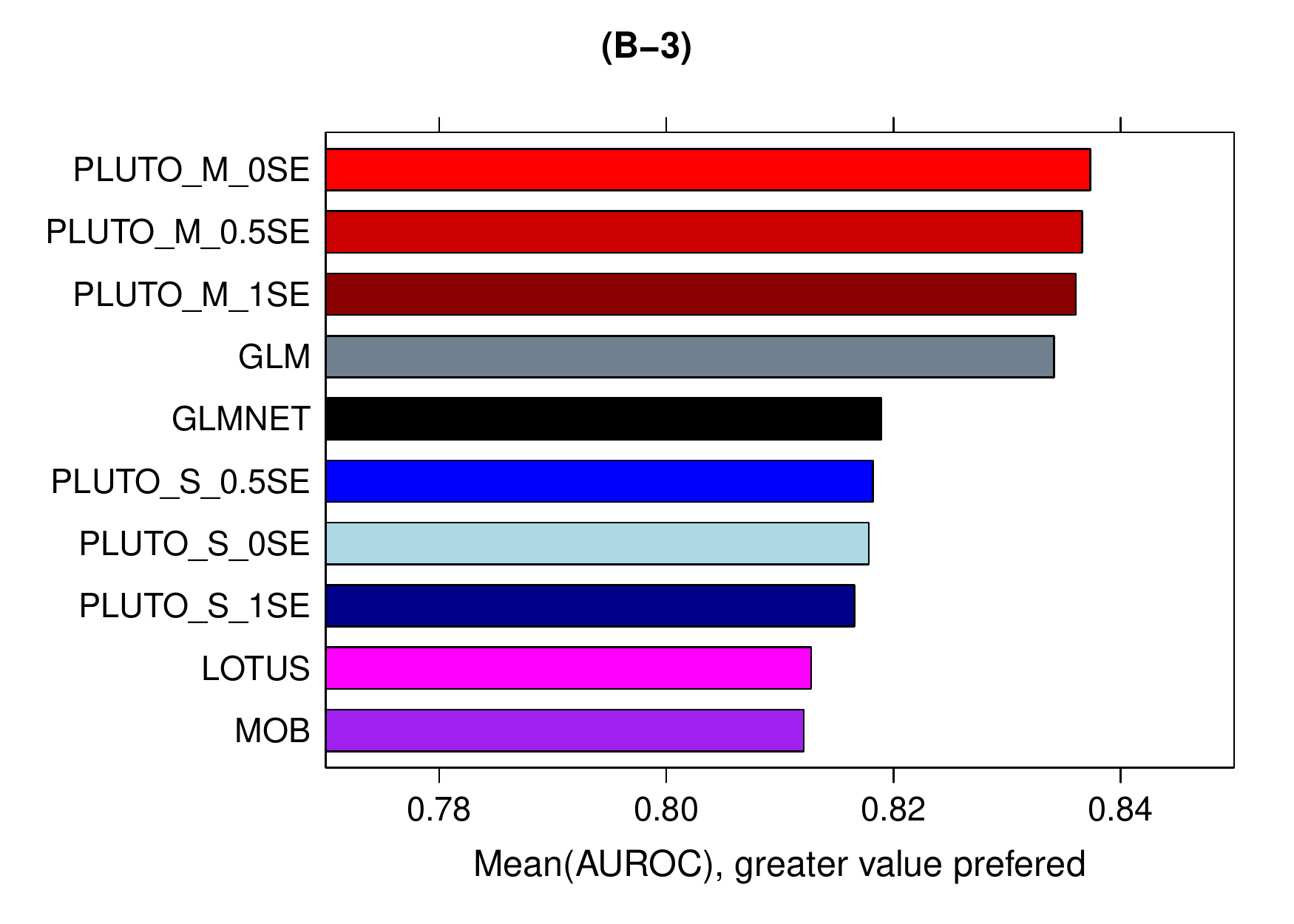}}
\caption[Bar charts of the mean measures for all algorithms except the PLUTO algorithms with bias correction.]{Bar charts of the mean measures for all algorithms except the PLUTO algorithms with bias correction. The measures in the plots are: (Top) ratio of deviance, (Middle) misclassification error rate, (Bottom) AUROC. In the left three plots, means are taken over all 20 datasets. While in the right three plots, means are taken over 14 datasets where no crash occurs.}
\label{P4:bar1}
\end{figure}

Tables \ref{T4:rslts.allmtd1} and \ref{T4:rslts.allmtd2} are visualized in Figure \ref{P4:bar1}, from which we find that:
\begin{enumerate}
\item Overall, the multiple linear PLUTO models (shown as red bars) dominate other algorithms in every plot. We also find that applying a higher value of SE for pruning reduces prediction accuracy in all plots but plot (A-3). In other words, the 0-SE rule is preferred for the multiple linear PLUTO models.
\item The simple linear PLUTO models (shown as blue bars) fall behind the multiple linear PLUTO models as expected. However, they still perform better in general than MOB and LOTUS except in plot (B-2), although MOB and LOTUS fit more complicated models in the nodes. Similar to the multiple linear PLUTO models, the majority of the plots suggest that the 0-SE rule results in better simple linear PLUTO trees.
\end{enumerate}

\clearpage
\subsection{Comparison of PLUTO models}
\begin{figure}[!ht]
                \centering
                \includegraphics[scale=0.5]{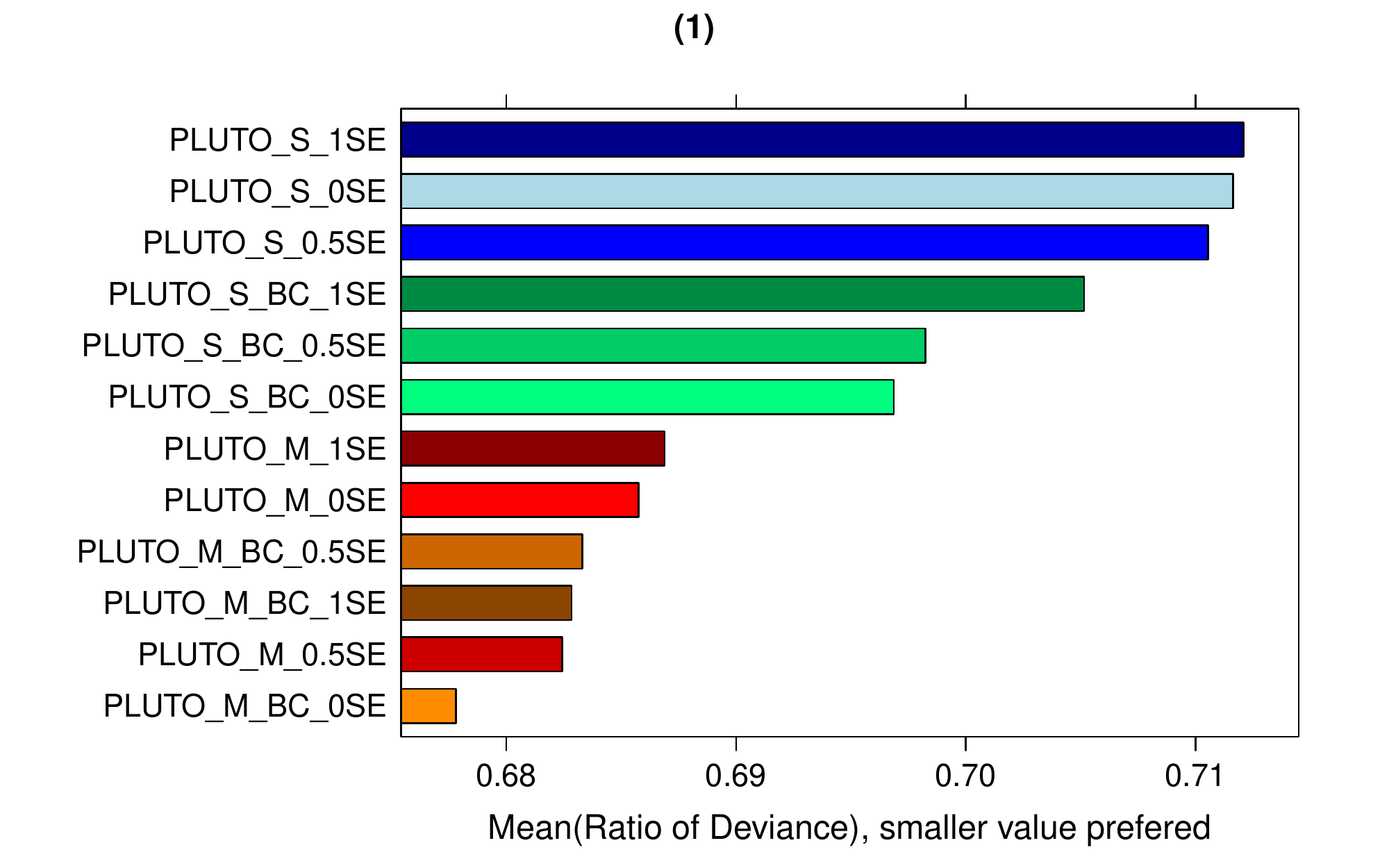}
                \includegraphics[scale=0.5]{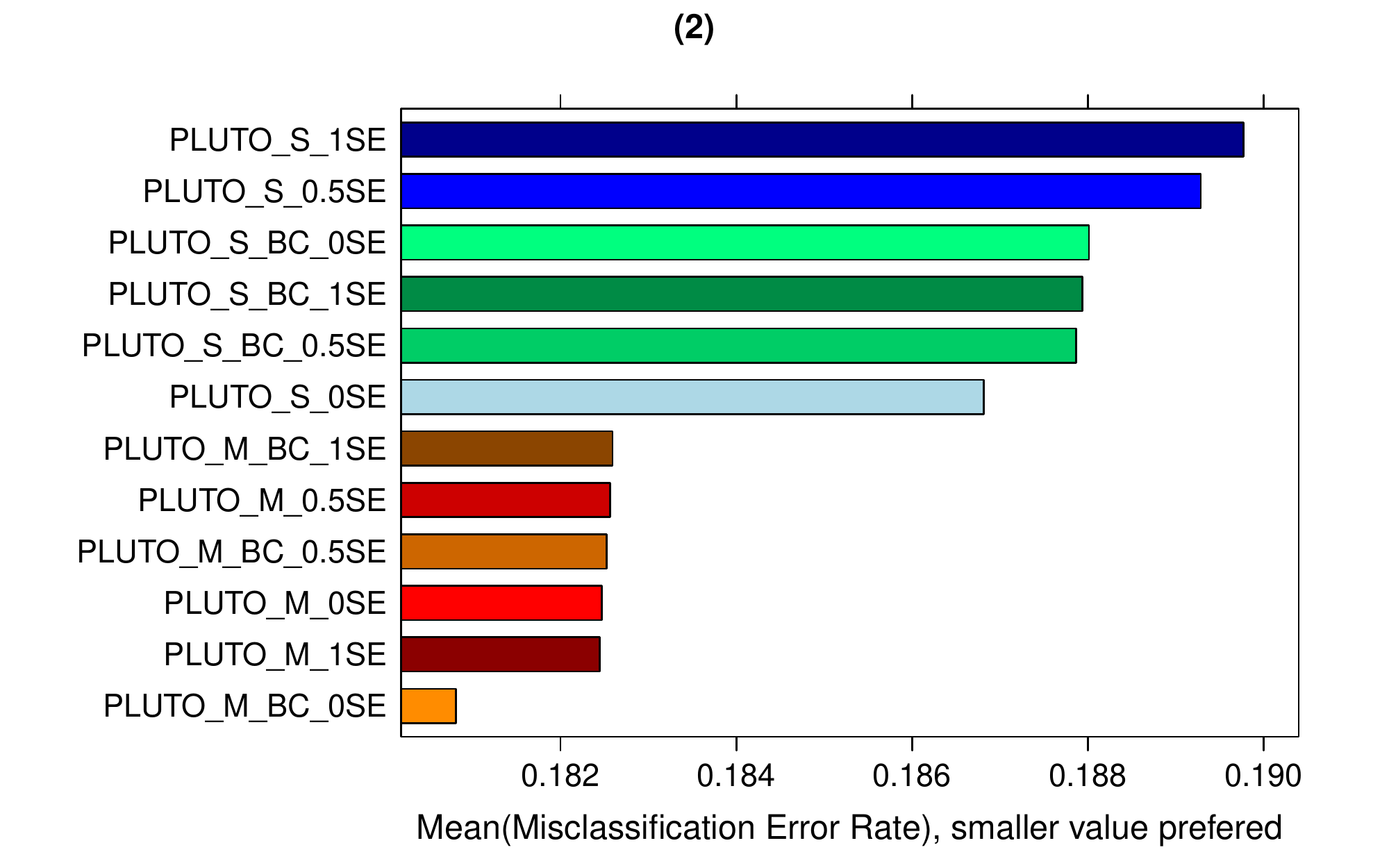}
                \includegraphics[scale=0.5]{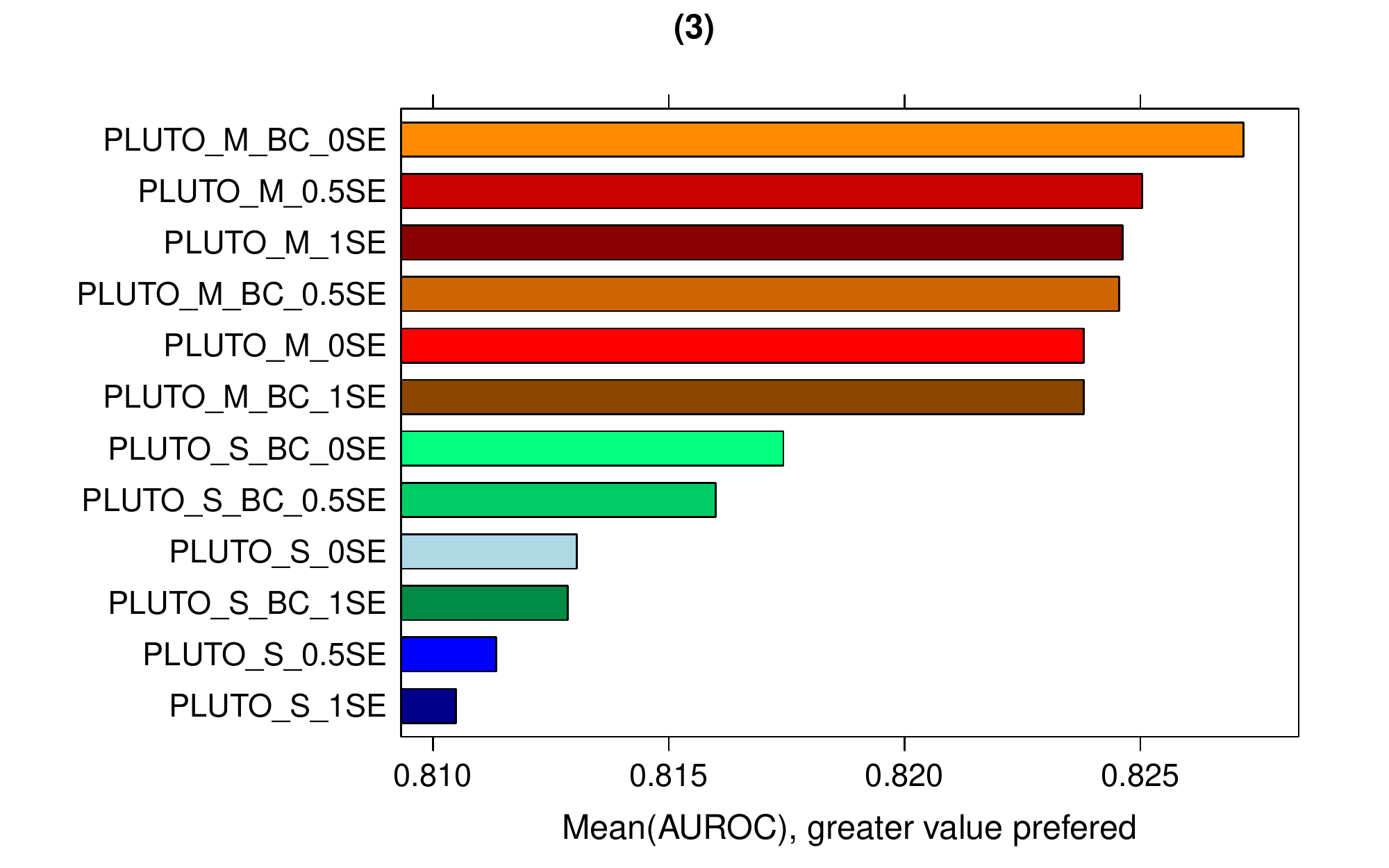}
\caption[Bar charts of the mean measures for 12 settings of PLUTO.]{Bar charts of the mean measures for 12 settings of PLUTO, over the 14 datasets that contain categorical predictors. The measures in the plots are: (Top) ratio of deviance, (Middle) misclassification error rate, (Bottom) AUROC.}
\label{P4:bar2}
\end{figure}

Lastly, we omit the six datasets on which the PLUTO models with bias correction are not applicable, and compare the models without bias correction against those with bias correction. Table \ref{T4:rslts.pluto} presents the results and Figure \ref{P4:bar2} gives a visual representation. The ``PLUTO\_M\_BC\_0SE" model provides the best prediction accuracy on average over the fourteen datasets with categorical predictors. The bias correction algorithm improves the performance of simple linear PLUTO models most of the time. However, models ``PLUTO\_M\_0.5SE" and ``PLUTO\_M\_1SE" work better in general than  ``PLUTO\_M\_BC\_0.5SE" and ``PLUTO\_M\_BC\_1SE", respectively.

Overall, the PLUTO bias correction algorithm does not necessarily improve the prediction accuracy. However, as discussed in Section \ref{sec3}, it has negligible selection bias.

\begin{table}[!htb]
\begin{center}
\caption{Arithmetic means of 10-fold cross-validation estimates of the ratio of deviance, misclassification error rate, and AUROC for different settings of PLUTO, over the 14 datasets that contain categorical predictors.}\label{T4:rslts.pluto}
\begin{tabular}{lp{0.15\linewidth}p{0.15\linewidth}l}
  \hline\hline
Model & RoD\textsuperscript{\dag} & MER\textsuperscript{\dag} & AUROC \\
  \hline
PLUTO\_S\_0SE & 0.712 & 0.187 & 0.813 \\
  PLUTO\_S\_0.5SE & 0.711 & 0.189 & 0.811 \\
  PLUTO\_S\_1SE & 0.712 & 0.190 & 0.810 \\
  PLUTO\_S\_BC\_0SE & 0.697 & 0.188 & 0.817 \\
  PLUTO\_S\_BC\_0.5SE & 0.698 & 0.188 & 0.816 \\
  PLUTO\_S\_BC\_1SE & 0.705 & 0.188 & 0.813 \\
  PLUTO\_M\_0SE & 0.686 & 0.182 & 0.824 \\
  PLUTO\_M\_0.5SE & 0.682 & 0.183 & 0.825 \\
  PLUTO\_M\_1SE & 0.687 & 0.182 & 0.825 \\
  PLUTO\_M\_BC\_0SE & 0.678 & 0.181 & 0.827 \\
  PLUTO\_M\_BC\_0.5SE & 0.683 & 0.183 & 0.825 \\
  PLUTO\_M\_BC\_1SE & 0.683 & 0.183 & 0.824 \\
   \hline
   \multicolumn{4}{l}{\textsuperscript{\dag}\footnotesize{RoD: Ratio of Deviance. MER: Misclassification Error Rate.}}
\end{tabular}
\end{center}
\end{table}

\clearpage
%%%%%%%%%%%%%%%%%%%%%%%%%%%%%%%%%%%%%%%%%%%%%%%%%%%%%%%%%%%%%%%%%%%%%%%%%%%%
\section{Example: Analysis of Census Income Data} \label{sec5}
From the previous section, we verified that PLUTO provides excellent prediction accuracy. Our tree-based model shows great data interpretation ability as well. For illustration, we employ PLUTO to analyze a real dataset in this section.

\subsection{Data description and visualization}
The census income dataset, also known as the ``adult" dataset, is available from the UCI Machine Learning Repository \citep{uci}. It is extracted by Barry Becker from the 1994 US Census Bureau database \citep{Kohavi96}. The original data were obtained from the Current Population Survey (CPS), a nationwide survey that collects numerous demographic, social, and economic characteristics of the population.

The census income dataset was prepared to predict whether or not an individual has a yearly income over \$50,000. This dataset contains 48,842 observations, which were randomly split into a training set and a testing set using $\mathcal{MLC++}$.\footnote{$\mathcal{MLC++}$ is a machine learning library in C++ \citep{Kohavi97}.} The training set contains 32,561 observations (2/3 of the total), leaving the testing set 16,281 observations (1/3 of the total). About $7\%$ of the observations have missing values. In this study, we remove all the instances with missing values. After removal, there are 30,162 observations in the complete training set, and 15,060 in the complete testing set. The dataset is made up of 15 variables, including the dependent variable. Table \ref{T5:variable} describes each variable. Among the 14 independent variables, 8 are categorical.

%table of variable
%\begin{table}[!htb]
\begin{center}
\begin{small}
%\begin{adjustwidth}{-1cm}{}
%\caption{Variables for census income data}\label{T5:variable}
%\begin{tabular}{l p{5cm} c p{7cm}}
\begin{longtable}{l p{5cm} c p{5cm}}
\caption{Variable descriptions of census income data}\label{T5:variable}\\

\hline\hline Name & Values & Type\textsuperscript{\dag} & Levels/Ranges\\ \hline
\endfirsthead

\multicolumn{4}{c}%
{{\bfseries \tablename\ \thetable{} -- continued from previous page}} \\
\hline\hline Name & Values & Type\textsuperscript{\dag} & Levels/Ranges\\ \hline
\endhead

\hline \multicolumn{4}{l}{\textsuperscript{\dag}\footnotesize{Variable type:``N"--numerical predictor, ``C"--categorical predictor, ``R"--response.}}\\
\endfoot

Age & & N & Min=17, Max=90 \\
Workclass & Type of employer & C &
		1. Private\newline
		2. Federal-gov\newline
		3. State-gov\newline
		4. Local-gov\newline
		5. Self-emp-not-inc\newline
		6. Self-emp-inc\newline
		7. Without-pay \\
Fnlwgt & Number of people the census takers believe that observation represents & N & Min=13,492 , Max=1,490,400 \\
Education & Highest level of education achieved & C &
        1. Preschool\newline
        2. 1st-4th\newline
        3. 5th-6th\newline
        4. 7th-8th\newline
        5. 9th\newline
        6. 10th\newline
        7. 11th\newline
        8. 12th\newline
        9. HS-grad\newline
        10. Some-college\newline
        11. Assoc-voc\newline
        12. Assoc-acdm\newline
        13. Bachelors\newline
        14. Masters\newline
        15. Prof-school\newline
        16. Doctorate \\
Education-num & Highest level of education in numerical form & N & Min=1 , Max=16 \\
Marital & Marital status & C &
		1. Married-civ-spouse\newline
		2. Married-AF-spouse\newline
		3. Married-spouse-absent\newline
		4. Separated\newline
		5. Divorced\newline
		6. Widowed\newline
		7. Never-married \\
Occupation & Descriptions of the individual's occupation & C &
		1. Adm-clerical\newline
		2. Armed-Forces\newline
		3. Craft-repair\newline
		4. Exec-managerial\newline
		5. Farming-fishing\newline
		6. Handlers-cleaners\newline
		7. Machine-op-inspect\newline
		8. Other-service\newline
		9. Priv-house-serv\newline
		10. Prof-specialty\newline
		11. Protective-serv\newline
		12. Sales\newline
		13. Tech-support\newline
		14. Transport-moving \\
Relationship & Descriptions of the individual's role in the household & C &
		1. Husband\newline
 		2. Wife\newline
		3. Own-child\newline
		4. Other-relative\newline
		5. Unmarried\newline
		6. Not-in-family \\
Race & Descriptions of the individual's race & C &
		1. White\newline
		2. Black\newline
		3. Asian-Pac-Islander\newline
		4. Amer-Indian-Eskimo\newline
		5. Other \\
Sex & Gender & C &
		1. Female\newline
        2. Male \\
Capital-gain & Capital gains recorded & N & Min=0 , Max=99,999 \\
Capital-loss & Capital Losses recorded & N & Min=0 , Max=4,356 \\
Hours & Hours worked per week & N & Min=1 , Max=99 \\
Country & Country of origin for the individual &  C & 41 levels/countries\\
Income & Whether or not the person has annual income over \$50,000 & R &
        1. $\leq50K$\newline
        2. $>50K$\\
%\end{tabular}
\end{longtable}
%\end{adjustwidth}
\end{small}
\end{center}
%\end{table}

The census income data is unbalanced. In the training set, 22,654 observations have annual income $\leq50K$, which consist of $75\%$ of the total training samples. Only $25\%$ of the observations have annual income $>50K$.

\begin{figure}[!h]
%\begin{center}
\makebox[\textwidth][c]{\includegraphics[width=0.45\textwidth]{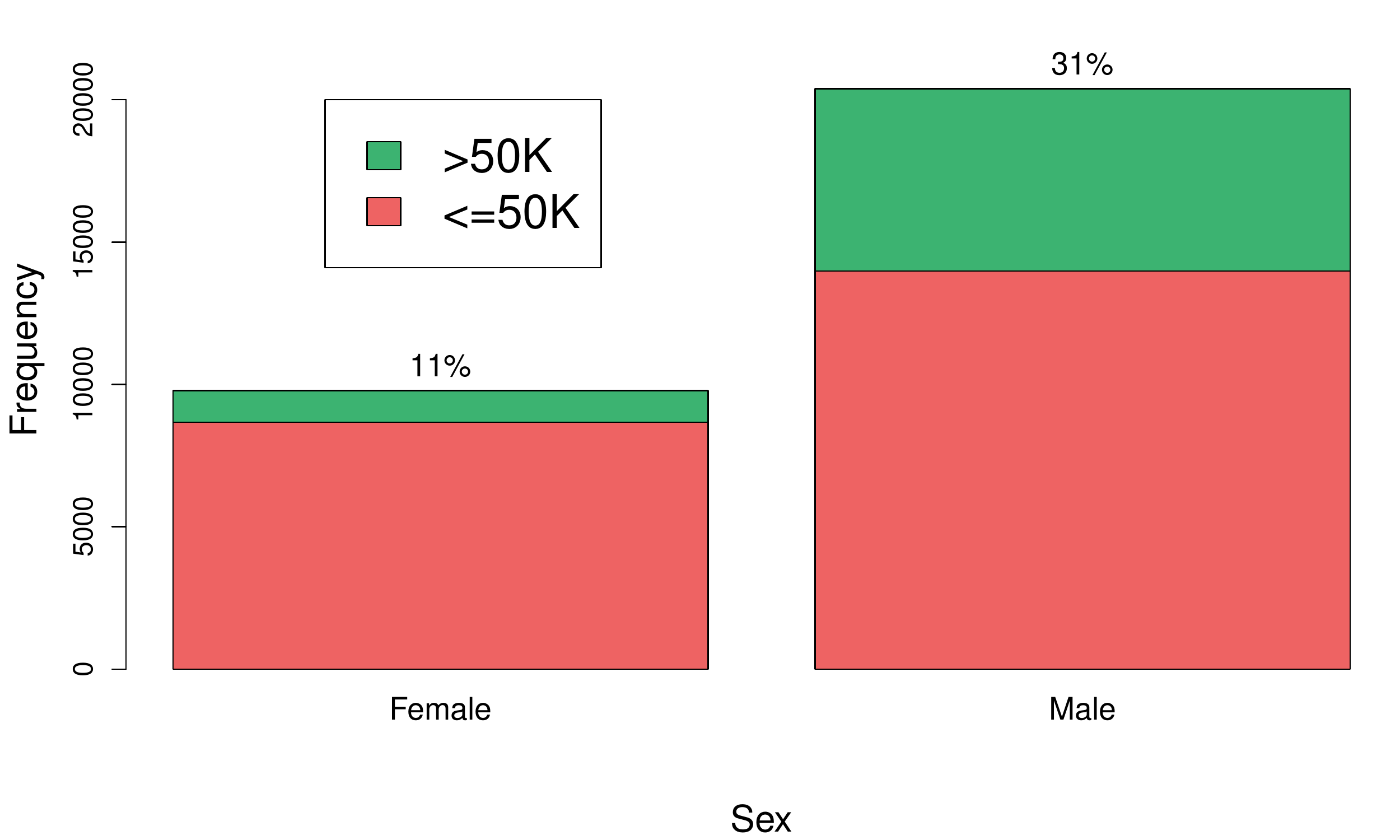}
\includegraphics[width=0.45\textwidth]{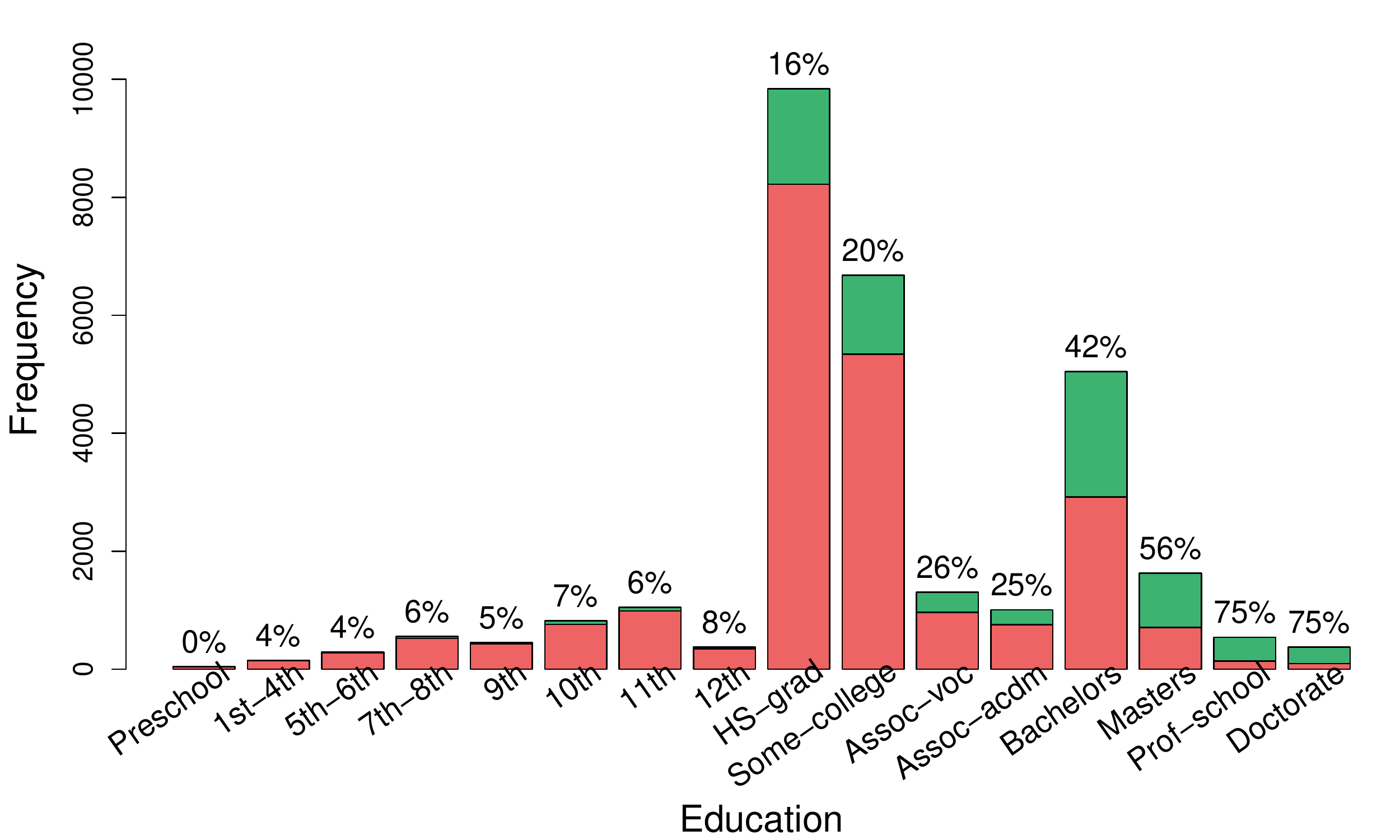}}
\makebox[\textwidth][c]{\includegraphics[width=0.45\textwidth]{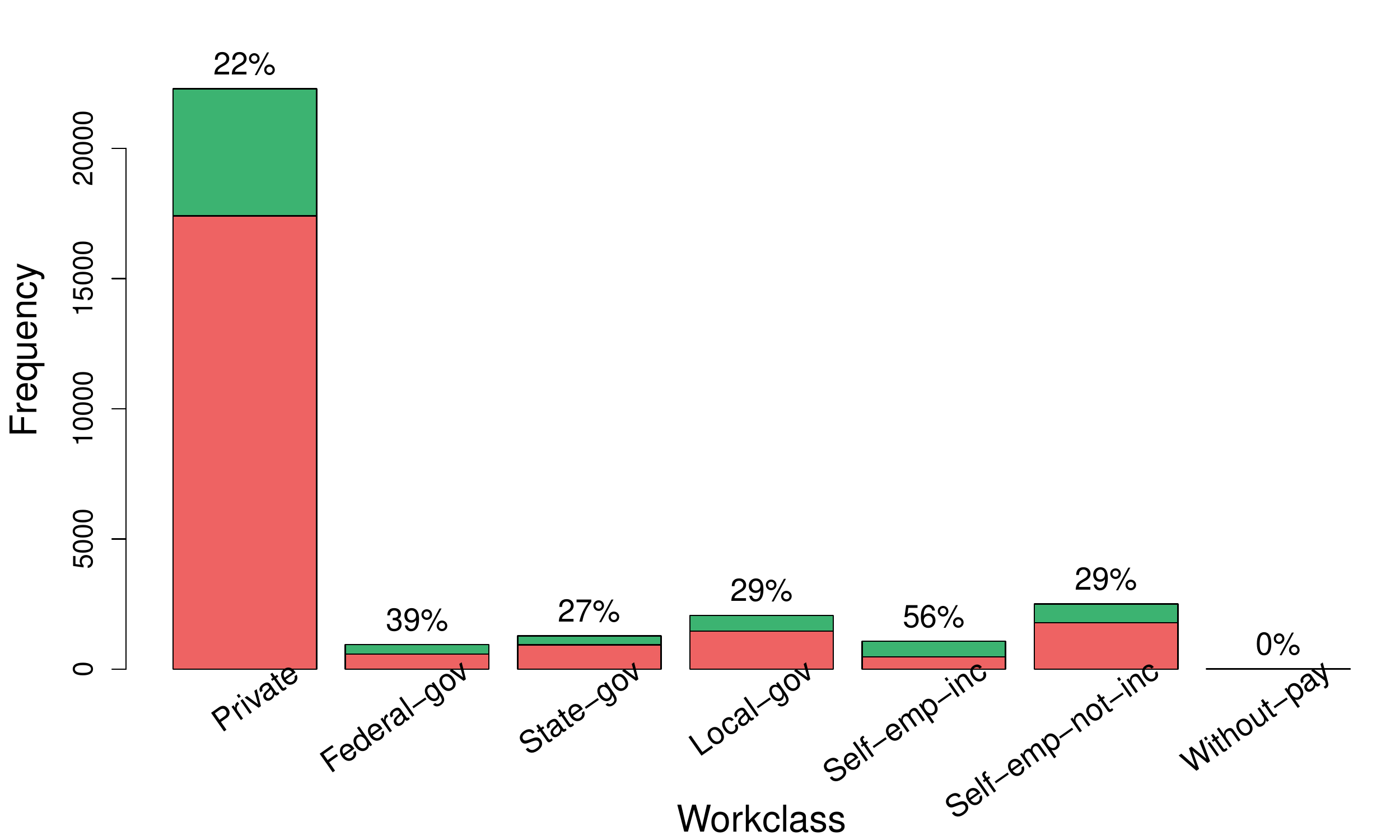}
\includegraphics[width=0.45\textwidth]{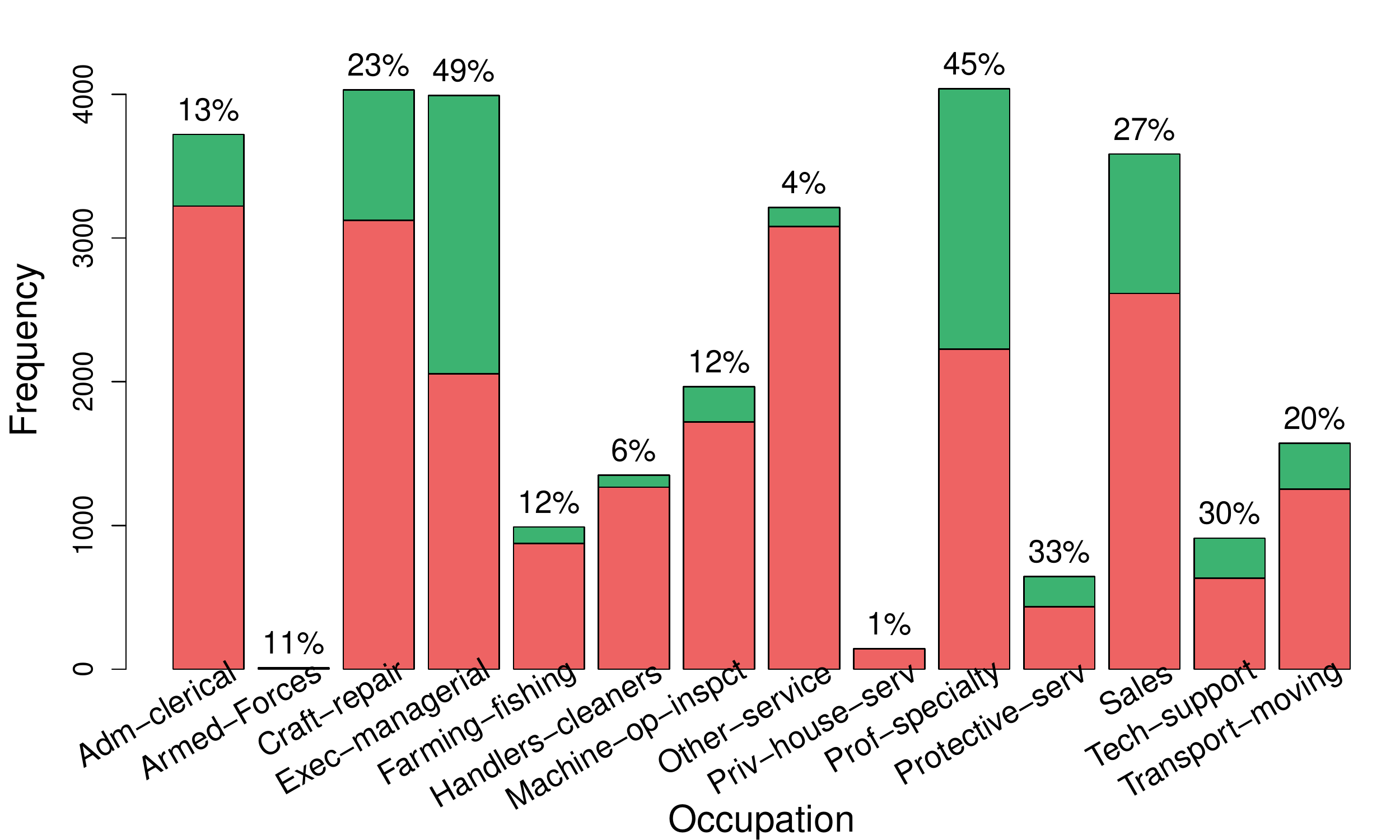}}
\makebox[\textwidth][c]{\includegraphics[width=0.45\textwidth]{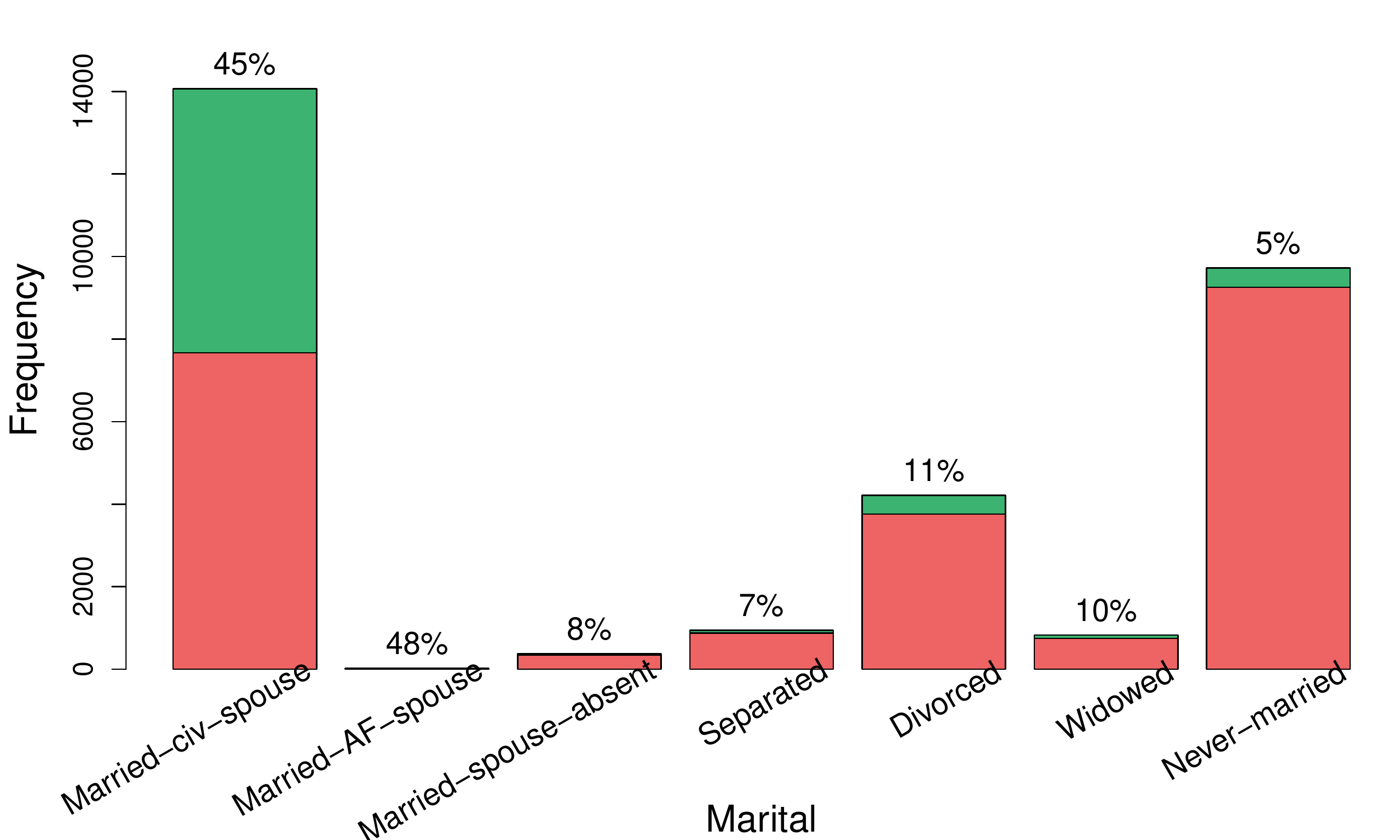}
\includegraphics[width=0.45\textwidth]{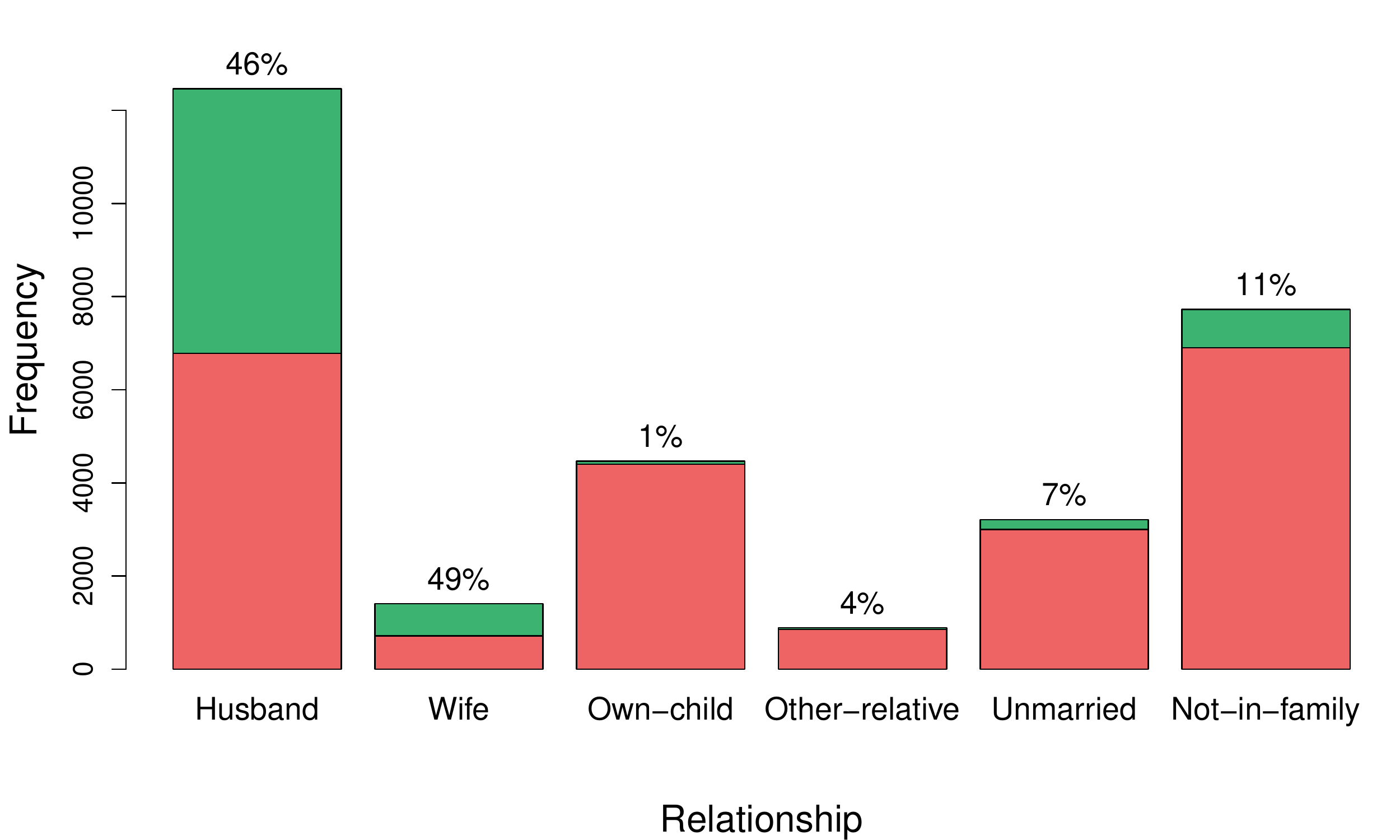}}
\makebox[\textwidth][c]{\includegraphics[width=0.45\textwidth]{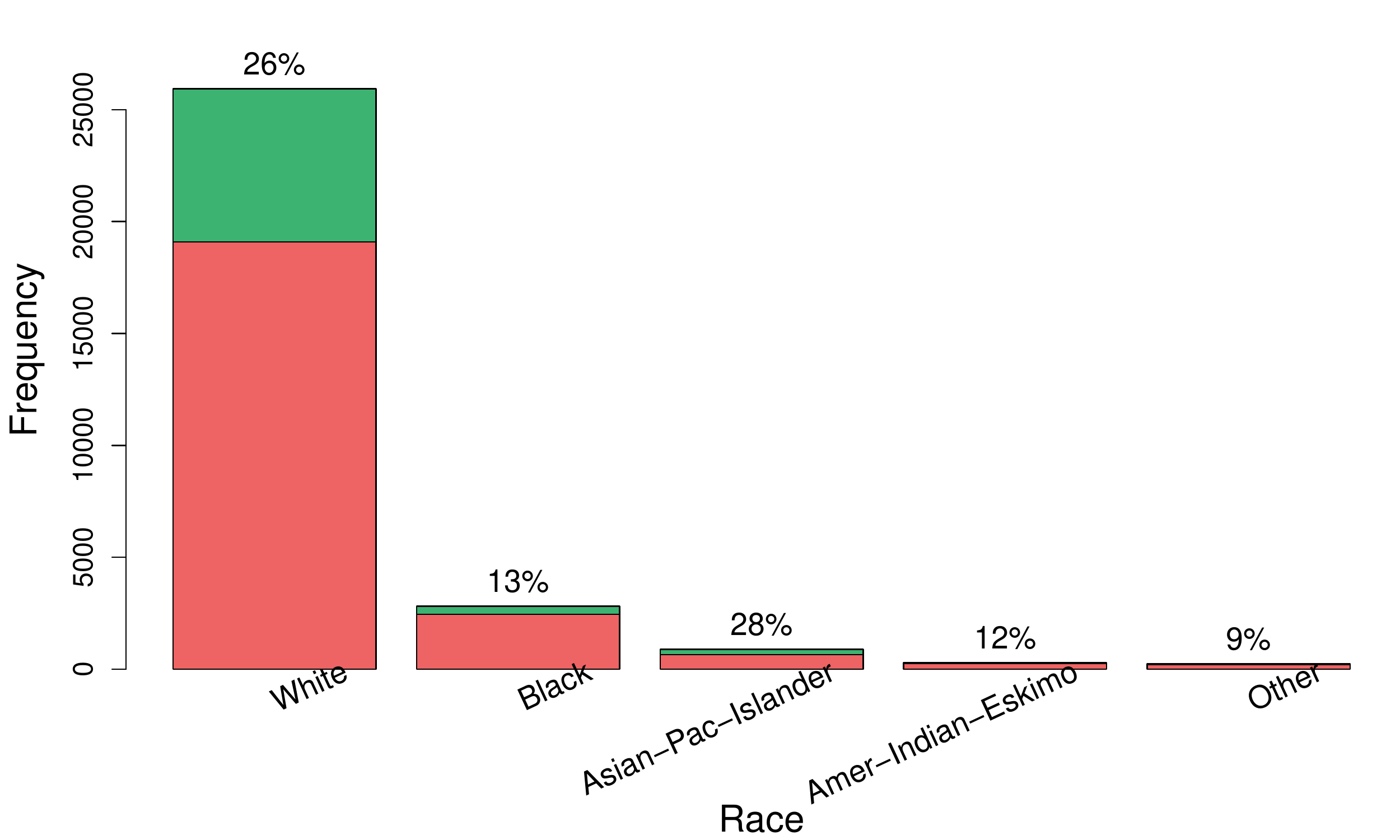}}
\caption[Distributions of income levels among groups of categorical variables.]{Distributions of income levels among groups of categorical variables. The percentage on each bar represents
the rate of people with annual income over 50K in that group.}\label{P5:barplots}
%\end{center}
\end{figure}

The stacked bar plots in Figure \ref{P5:barplots} show the population distribution and the distribution of income levels among groups of categorical variables. We found that the variable ``Country" is rather difficult to visualize and is omitted because it contains an excessive number of values. Among these values, the United States represents the majority of the observations and a number of countries have very small frequencies. Of the 30,162 training samples, 20,380 are male ($68\%$) while only 9,782 are female ($32\%$). About $31\%$ of the men make over 50K in a year, which is higher than the rate of women ($11\%$). Among the education groups, high school graduates, college dropouts, and bachelors have high frequencies. Persons with higher education levels tend to earn more. The majority of the samples are employed by private companies ($74\%$), among which about $22\%$ have annual income above 50K. The high income rates for self-employed and government workers are greater. Persons in managerial and professional specialty occupations have higher income. On the other hand, people in service occupations have the lowest chance to make over 50K per year. The bar plots regarding marital status and family relationships suggest that spouses in healthy and complete families earn more. In the last plot, we find that the majority of the survey participants are white ($86\%$). White and Asian-Pacific-Islander are the two races that have the top high income rates.

\begin{figure}[!h]
                %\centering
                \makebox[\textwidth][c]{\includegraphics[scale=0.5]{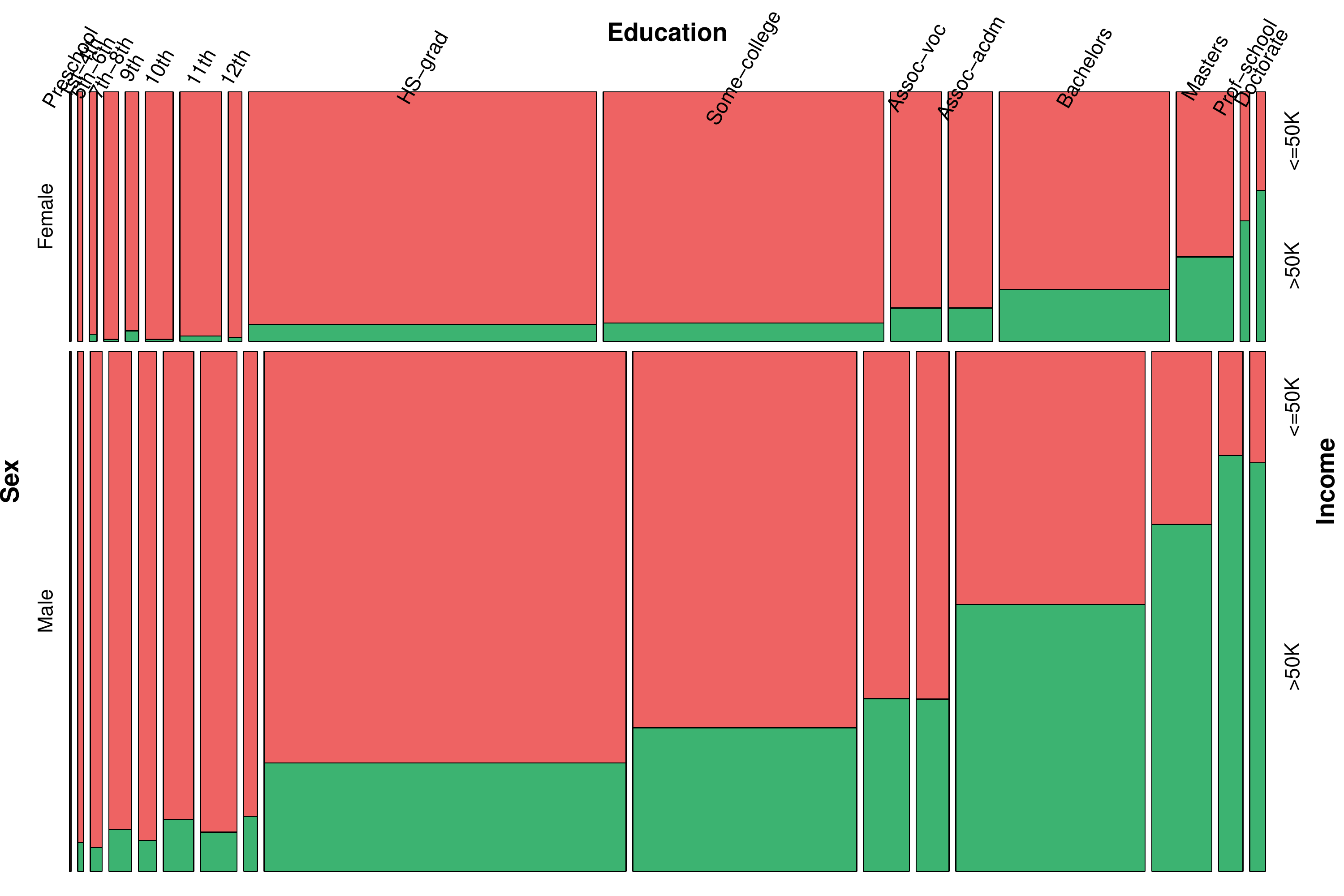}}
\caption{Distributions of income levels among different combinations of sex groups and education groups.}
\label{P5:mosaic1}
\end{figure}

To explore any possible interactions between categorical variables, we use mosaic plots to visualize the population distribution and the distribution of income levels among combinations of categorical groups. Figure \ref{P5:mosaic1} shows the distribution of population among a combination of ``Sex", ``Education", and ``Income" variables. More men have a degree equal to or higher than Bachelors compared with women. The high income rate of women is consistently lower than that of men across all education types, suggesting no interactions between ``Sex" and ``Education".

\begin{figure}[!ht]
                %\centering
                \makebox[\textwidth][c]{\includegraphics[scale=0.5]{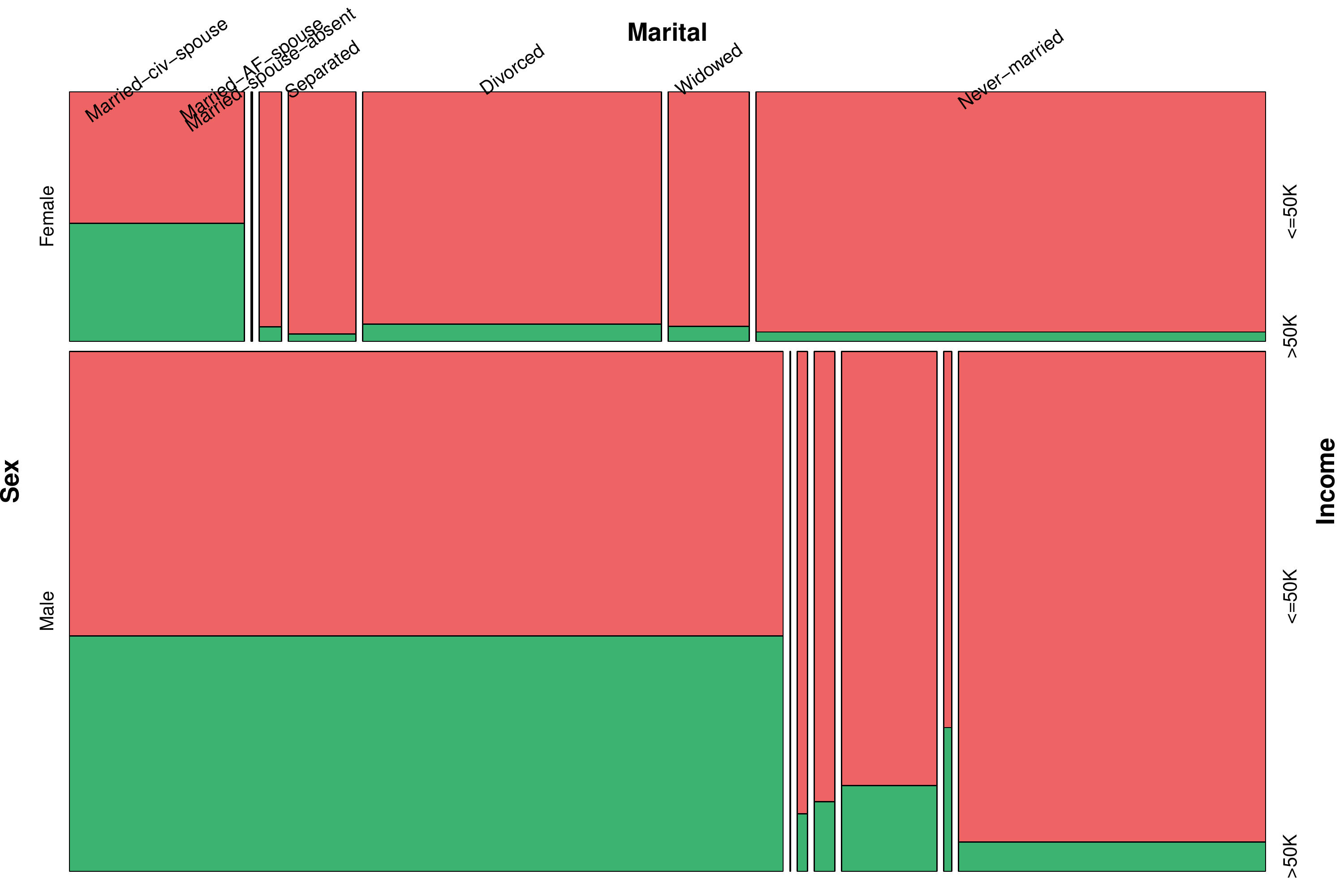}}
\caption{Distributions of income levels among different combinations of sex groups and marital status.}
\label{P5:mosaic2}
\end{figure}

Figure \ref{P5:mosaic2} shows the distribution of population among a combination of ``Sex", ``Marital", and ``Income" variables. About $62\%$ of male workers are married. On the other hand, among female workers, only $15\%$ are married. Although the overall high income rate for women is much lower than men as shown in Figure \ref{P5:barplots}, from Figure \ref{P5:mosaic2}, we actually find that the high income rates among married female workers ($47\%$ and $50\%$ respectively for civilian spouse and military spouse) are greater than their male counterparts ($45\%$ and $44\%$). The difference implies that there may be interactions between ``Sex" and ``Marital".

\begin{figure}[!ht]
                \makebox[\textwidth][c]{\includegraphics[scale=0.55]{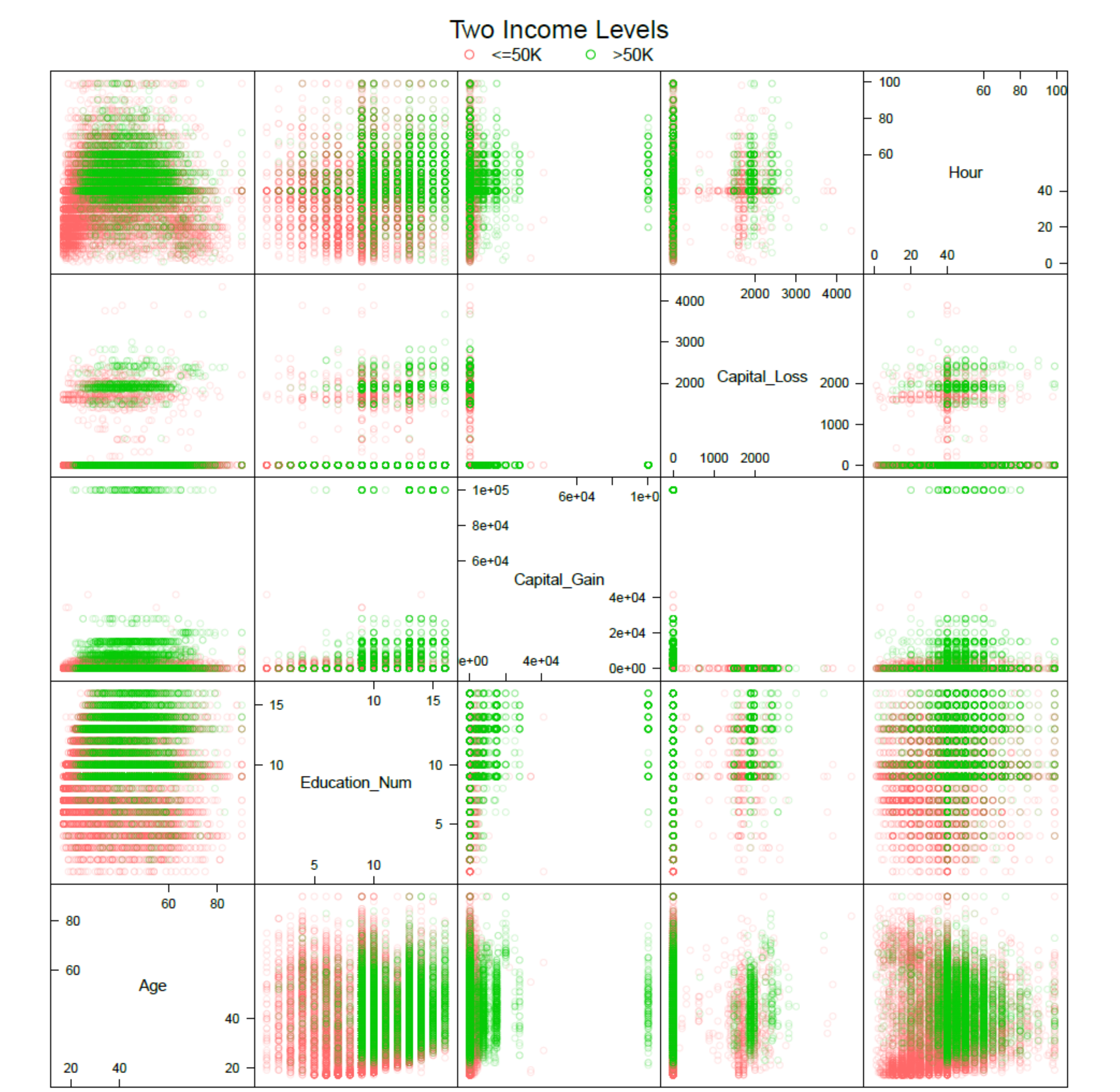}}
\caption{Scatter plot matrix of census income data.}
\label{P5:SPM}
\end{figure}

Figure \ref{P5:SPM} is a scatter plot matrix for all numerical variables of the data. We observe that both ``Capital-gain" and ``Capital-loss" are heavily skewed. The high income rate increases as education level increases, while the rest of the picture seems to be quite ``foggy". More advanced statistical models would be beneficial to explore the patterns in this dataset.

\subsection{Previous study}
The census income dataset was analyzed in \citet{Kohavi97}. They built classifiers by applying various machine learning algorithms to the complete training set. Then they used the classifier to predict the complete testing sample and obtained the misclassification error rate results as shown in Table \ref{T5:pre.rslts}. Among all the algorithms that they compared, FSS Naive Bayes \citep{Kohavi95featuresubset}, which is feature subset selection on top of Naive-Bayes, has the lowest error rate. The runner-up is NBTree \citep{Kohavi96}, a tree hybrid with Naive-Bayes at the leaves.
\begin{table}[!htb]
\begin{center}
\caption[Misclassification error rate results on the census income dataset of algorithms available in $\mathcal{MLC++}$.]{Misclassification error rate results on the census income dataset of algorithms available in $\mathcal{MLC++}$ \citep{Kohavi97}.}\label{T5:pre.rslts}
\begin{tabular}{l p{2cm} l}
  \hline\hline
  Algorithm     &     &     Error($\%$)\\
  \hline
 C4.5          &   &       15.54\\
 C4.5-auto    &    &       14.46\\
 C4.5 rules    &   &       14.94\\
 Voted ID3 (0.6) & &       15.64\\
Voted ID3 (0.8)   &&      16.47\\
T2                 &&     16.84\\
1R                 & &    19.54\\
 NBTree             & &    14.10\\
CN2                  &&   16.00\\
HOODG                &&   14.82\\
FSS Naive Bayes      &&   14.05\\
IDTM (Decision table) &&  14.46\\
Naive-Bayes           &&  16.12\\
Nearest-neighbor (1)  &&  21.42\\
Nearest-neighbor (3)  &&  20.35\\
OC1                   &&  15.04\\
   \hline
\end{tabular}
\end{center}
\end{table}

The objective of their study was merely the comparison of the prediction performance for the machine learning algorithms. Hence, it lacks data analysis and interpretation.

\citet{Antonov14} analyzed the census income data. They built decision tree classifier and naive Bayes classifier to the training dataset. Based on these classifiers, they employed the permutation method to measure the variable importance. However, they neglected the tree structure itself, which may provide insight into the data. Second, the tree method they applied is based on CART \citep{cart}. As we discussed earlier, CART has the selection bias problem, which may jeopardize interpretation. Lastly, the classifiers that they trained offer dichotomous prediction only, of which performance measures such as deviance and AUROC are not used. In contrast, logistic regression trees can employ these measures to improve variable ranking.

\subsection{Analysis using PLUTO}
In this subsection, we use PLUTO to analyze the census income data. The goal of our analysis is to develop PLUTO models that can identify the high income worker and to explore the role of each predictor.

First, we modify the data by excluding the variable ``Fnlwgt" from our analysis because weights of observations are not employed in the current version of PLUTO.

\subsubsection{PLUTO with simple linear logistic regression leaves}
We start with the best simple linear logistic regression option of PLUTO. After growing a large tree, we prune it back using the minimal cost-complexity method of CART. In Table \ref{T5:prune.rslts}, partial results of the pruning process are listed, including the tree sizes, the cross-validated estimates of deviance, and the corresponding standard error estimates. Tree size represents the number of terminal(leaf) nodes in the tree, denoted by $|\tilde{T}|$.

\begin{table}[!htb]
\begin{center}
\caption{Partial results of minimal cost-complexity pruning on Model 1 for the census income data.}
\label{T5:prune.rslts}
\begin{tabular}{p{.15\linewidth}p{.15\linewidth}p{.2\linewidth}}
  \hline\hline \noalign{\smallskip}
Tree No. & $|\tilde{T}|$ & $\hat{\mathcal{D}} \pm \mbox{SE}$ \\
  \hline
1 & 227 &     $ 2766 \pm 64 $ \\
  24 & 200 &  $ 2645 \pm 65 $ \\
  60 & 150 &  $ 2464 \pm 51 $ \\
  93 & 101 &  $ 2209 \pm 46 $ \\
  124 & 50 &  $ 2017 \pm 33 $ \\
  133 & 40 & $ 1955 \pm 23 $ \\
  134 & 39 & $ 1953 \pm 24 $ \\
  135\textsuperscript{\dag} & 38\textsuperscript{\dag} & $ 1945 \pm 25 $ \\
  136\textsuperscript{*} & 36\textsuperscript{*} & $ 1948 \pm 23 $ \\
  137 & 35 & $ 1958 \pm 20 $ \\
  138 & 33 & $ 1962 \pm 18 $ \\
  139 & 32 & $ 1959 \pm 21 $ \\
  140 & 31 & $ 1961 \pm 21 $ \\
  141 & 30 & $ 1964 \pm 20 $ \\
  142 & 29 & $ 1959 \pm 18 $ \\
  143\textsuperscript{**} & 27\textsuperscript{**} & $ 1962 \pm 18 $ \\
  144 & 26 & $ 1972 \pm 17 $ \\
  152 & 16 & $ 2027 \pm 18 $ \\
  163 & 1 & $ 3011 \pm 8 $ \\
   \hline \noalign{\smallskip}
   \multicolumn{3}{l}{\textsuperscript{\dag}\footnotesize{Minimal $\hat{\mathcal{D}}$ (0-SE) tree (Model 1).}}\\
   \multicolumn{3}{l}{\textsuperscript{*}\footnotesize{0.5-SE tree.}}\\
   \multicolumn{3}{l}{\textsuperscript{**}\footnotesize{1-SE tree.}}\\
\end{tabular}
\end{center}
\end{table}

\begin{landscape}
\begin{figure}[!ht]
\begin{adjustwidth}{-2cm}{}
                \includegraphics[scale=0.7]{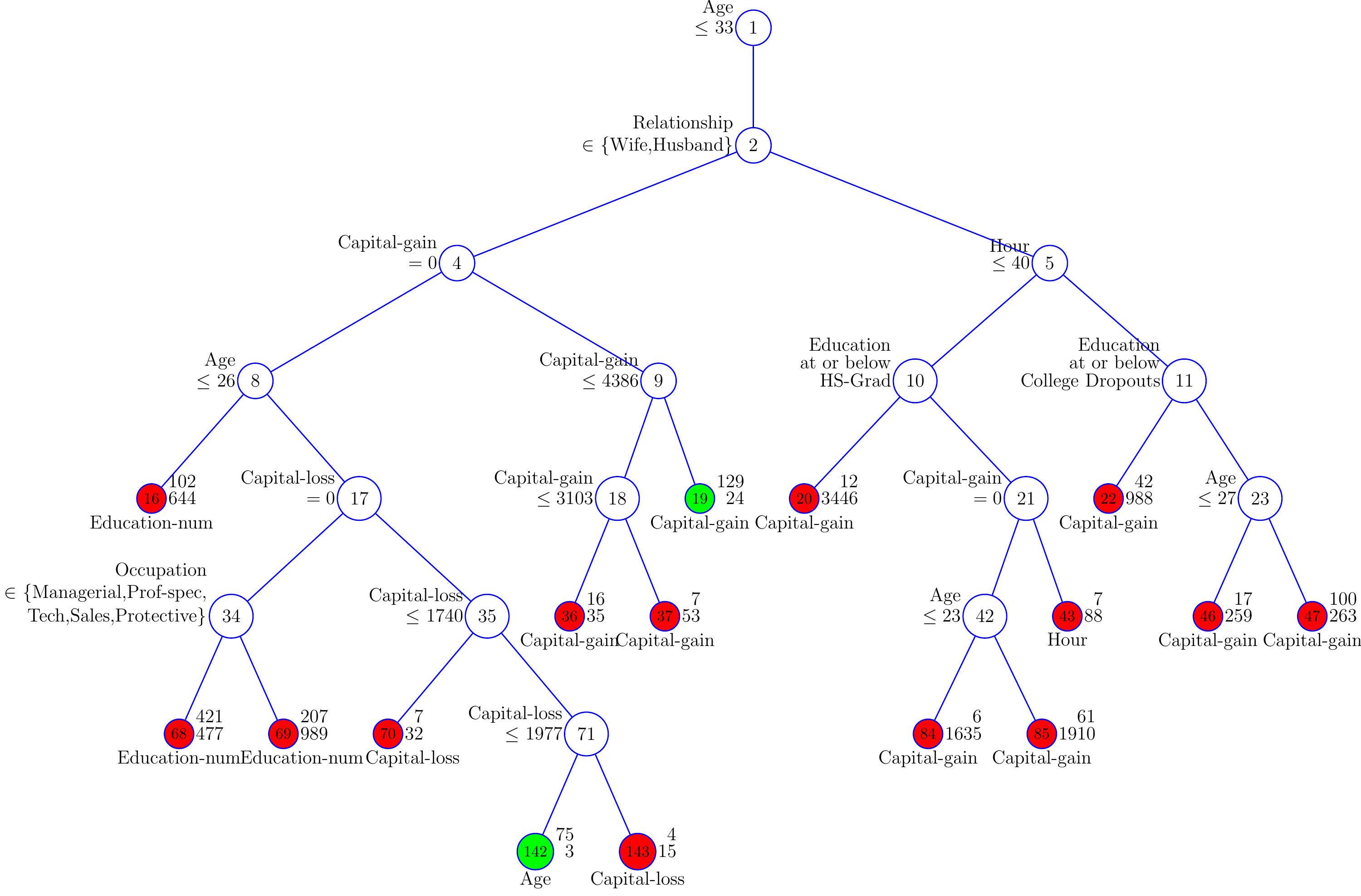}
\end{adjustwidth}
\caption[Model 1. PLUTO 0-SE tree with simple linear logistic regression leaves for the census income data, part (i)]{\small Model 1. PLUTO 0-SE tree with simple linear logistic regression leaves for the census income data, part (i). The branch beyond Node 2. At each splitting node, an observation goes to the left if and only if the criterion to the left of the node is satisfied. At each terminal node, the variable beneath represents the best regressor. The numbers to the right of the terminal nodes are numbers of observations that have annual income $> 50K$ (top) and $\leq 50K$ (bottom). \textcolor[rgb]{0.00,0.59,0.00}{Green} circle indicates more observations with annual income above 50K. \textcolor[rgb]{1.00,0.00,0.00}{Red} otherwise.}
\label{P5:SLnode2}
\end{figure}
\end{landscape}

\begin{landscape}
\begin{figure}[!ht]
\begin{adjustwidth}{-3cm}{}
                \includegraphics[scale=0.62]{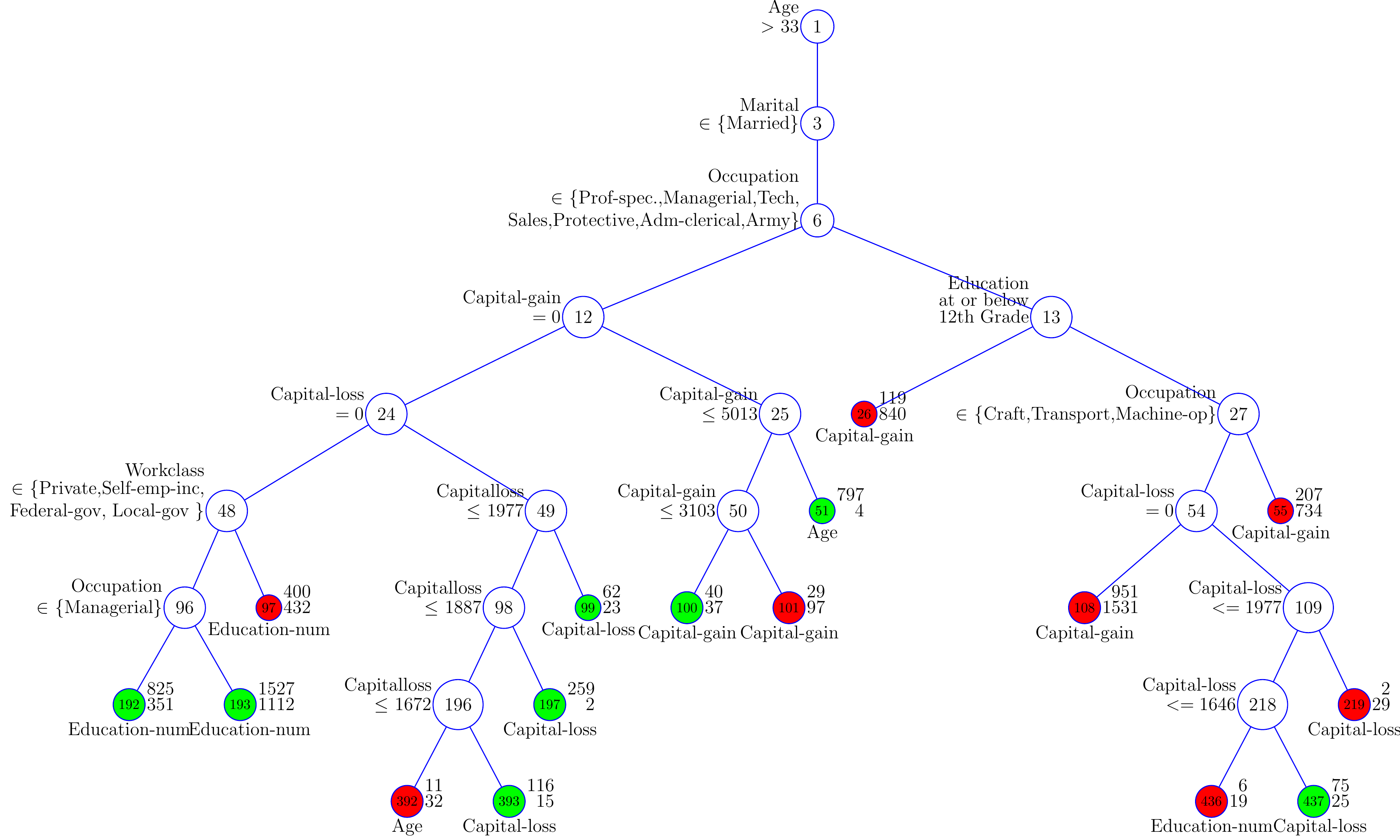}
\end{adjustwidth}
\caption[Model 1. PLUTO 0-SE tree with simple linear logistic regression leaves for the census income data, part (ii).]{\small Model 1. PLUTO 0-SE tree with simple linear logistic regression leaves for the census income data, part (ii). The branch beyond Node 6. At each splitting node, an observation goes to the left if and only if the criterion to the left of the node is satisfied. At each terminal node, the variable beneath represents the best regressor. The numbers to the right of the terminal nodes are numbers of observations that have annual income $> 50K$ (top) and $\leq 50K$ (bottom). \textcolor[rgb]{0.00,0.59,0.00}{Green} circle indicates more observations with annual income above 50K. \textcolor[rgb]{1.00,0.00,0.00}{Red} otherwise.}
\label{P5:SLnode6}
\end{figure}
\end{landscape}

\begin{figure}[!ht]
                \makebox[\textwidth][c]{\includegraphics[scale=0.95]{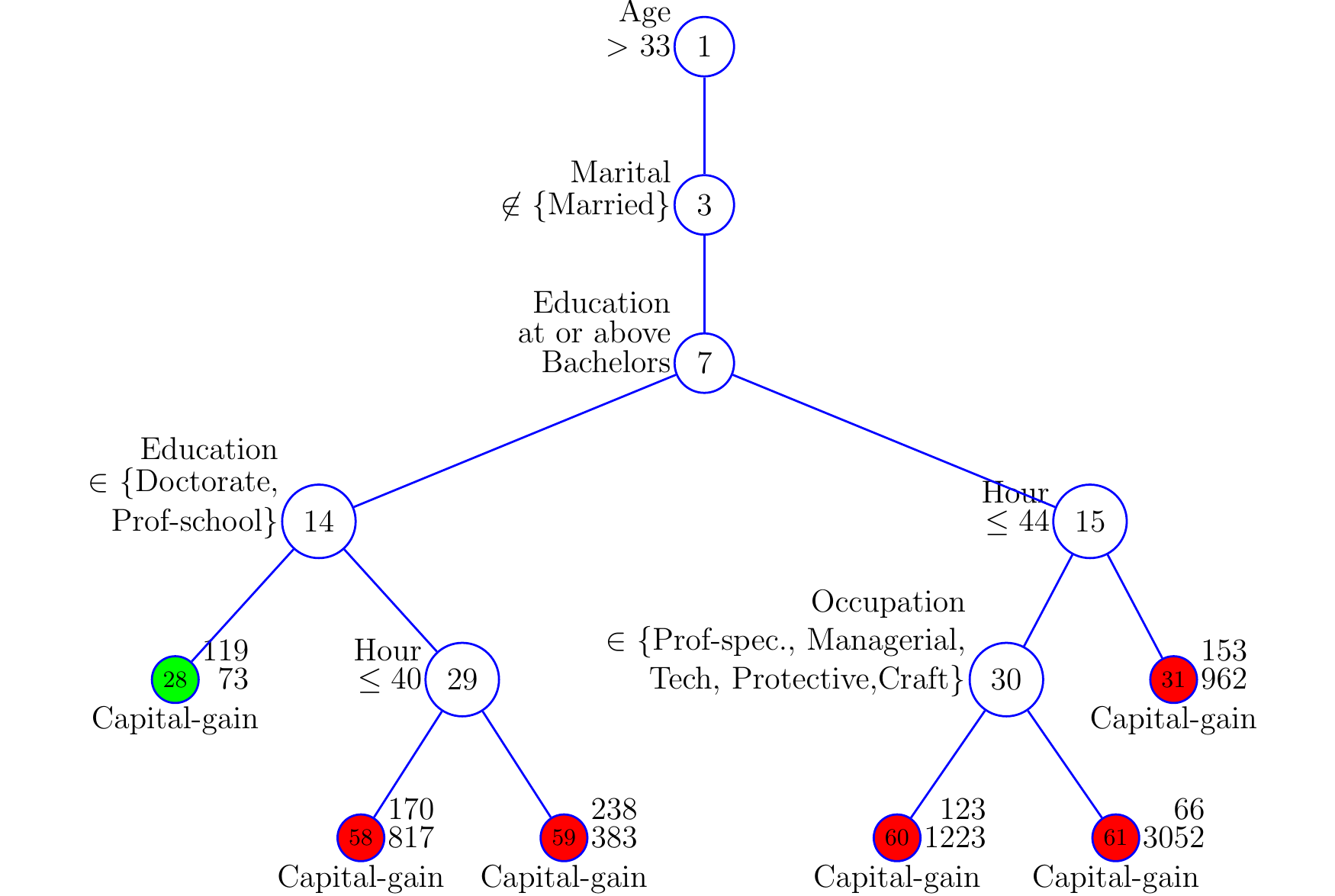}}
\caption[Model 1. PLUTO 0-SE tree with simple linear logistic regression leaves for the census income data, part (iii).]{Model 1. PLUTO 0-SE tree with simple linear logistic regression leaves for the census income data, part (iii). The branch beyond Node 7. At each splitting node, an observation goes to the left if and only if the criterion to the left of the node is satisfied. At each terminal node, the variable beneath represents the best regressor. The numbers to the right of the terminal nodes are numbers of observations that have annual income $> 50K$ (top) and $\leq 50K$ (bottom). \textcolor[rgb]{0.00,0.59,0.00}{Green} circle indicates more observations with annual income above 50K. \textcolor[rgb]{1.00,0.00,0.00}{Red} otherwise.}
\label{P5:SLnode7}
\end{figure}

We call the PLUTO 0-SE tree with simple linear option ``Model 1". It is a large tree with 38 terminal nodes, and it takes Figures \ref{P5:SLnode2}, \ref{P5:SLnode6}, and \ref{P5:SLnode7} to fully display the tree structure. At the root node of Model 1, the data first splits on Age, where an observation goes to the left branch (Node 2) if and only if he/she is at or below 33 years old. Figure \ref{P5:SLnode2} shows that for the younger population, the majority of the observations have annual income less than or equal to 50K. Participants above 33 are sent to the right branch (Node 3), where the data are further partitioned on Marital Status. In the $\{\mbox{Age}>33\}\cap\{\mbox{Married}\}$ group (Figure \ref{P5:SLnode6}), occupations including Professional specialty, Executive Managerial, Tech-support, Sales, Protective Service, Admin clerical and Army provide higher income. In the $\{\mbox{Age}>33\}\cap\{\mbox{Spouse-absent, Separated, Divorced, Widowed, Never-married}\}$ group, only those with Doctorate or Professional school education have a good chance of making over 50K a year. Capital-gain is selected most frequently as the best regressor in the terminal nodes, and the tree structure indicates that there are interactions between the regressor Capital-gain and the split variables.

\begin{figure}[!ht]
%\begin{adjustwidth}{-0.5cm}{}
\centering
                \includegraphics[width=\textwidth]{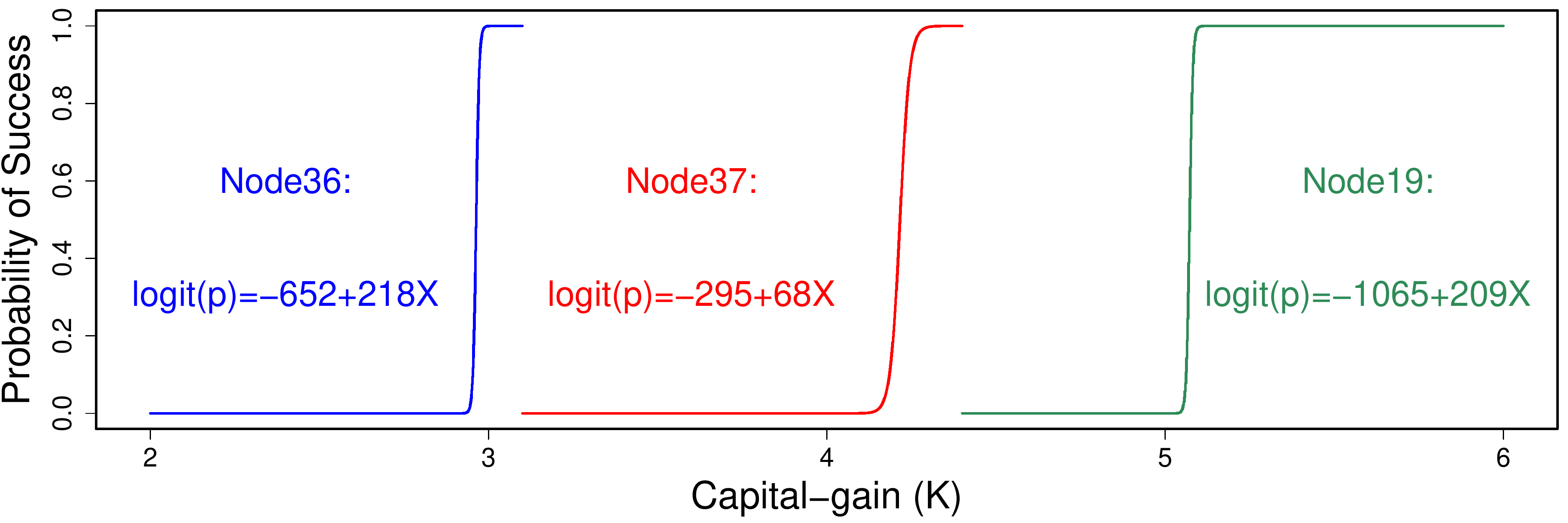}
%\end{adjustwidth}
\caption{S-curves of Nodes 36, 37, and 19 of Model 1 for the census income data.}
\label{P5:node9}
\end{figure}

Furthermore, in the branches starting at Nodes 4, 12, and 54, Capital-gain/Capital-loss are included as both the split variables and the fitting variables. We take a closer look at the branch starting at Node 9, and draw the s-curves of its terminal nodes in Figure \ref{P5:node9}. The difference in the slopes suggest that Capital-gain has nonlinear effect on the income.

The tree diagram makes it easier to identify the roles of each independent variable and to detect the interactions among them. However, we can see that as the depth of the tree grows, the interpretability of the tree model drops. This is a trade-off between the interpretability and the prediction accuracy. When prediction is not of primary interest, we can set the program to select a smaller tree, e.g., the PLUTO 2-SE tree, which will be shorter.

\subsubsection{PLUTO with regularized multiple linear logistic regression leaves}
\begin{landscape}
\begin{figure}[!ht]
\begin{adjustwidth}{-2.5cm}{}
                \includegraphics[scale=0.62]{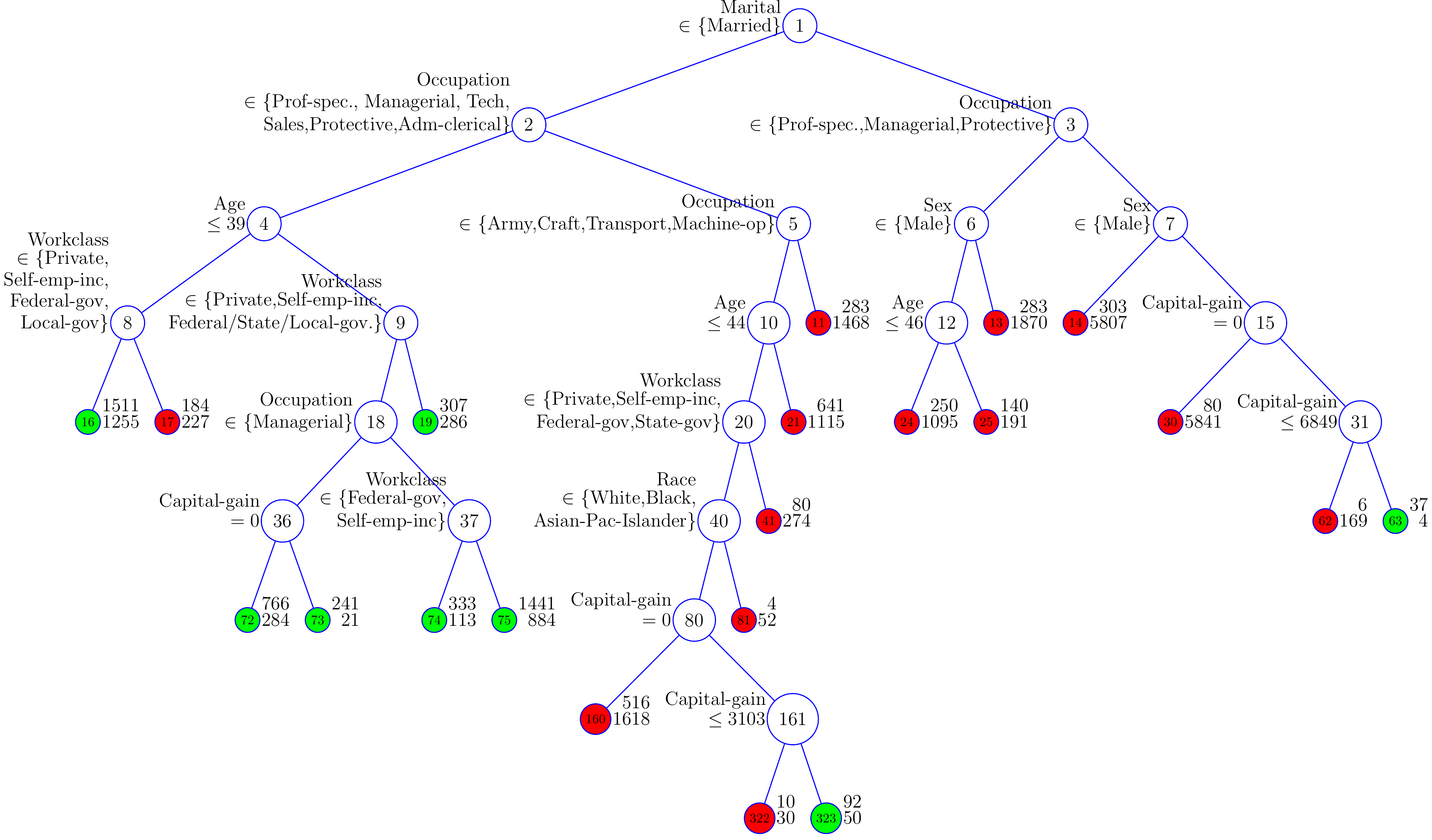}
\end{adjustwidth}
\caption[Model 2. PLUTO 0-SE tree with regularized multiple linear logistic regression leaves for the census income data.]{\small Model 2. PLUTO 0-SE tree with regularized multiple linear logistic regression leaves for the census income data. At each splitting node, an observation goes to the left if and only if the criterion to the left of the node is satisfied. At each terminal node, the numbers to the right are the numbers of observations that have annual income $> 50K$ (top) and $\leq 50K$ (bottom). \textcolor[rgb]{0.00,0.59,0.00}{Green} circle indicates more observations with annual income above 50K. \textcolor[rgb]{1.00,0.00,0.00}{Red} otherwise.}
\label{P5:ML0SE}
\end{figure}
\end{landscape}

The PLUTO regularized multiple linear option provides another solution for us to obtain a shorter tree without compromising prediction accuracy. The pruned PLUTO 0-SE tree with regularized multiple linear logistic regression leaves is displayed in Figure \ref{P5:ML0SE}, which contains 21 terminal nodes. We denote this tree as ``Model 2".

\begin{figure}[!ht]
\begin{adjustwidth}{-0.5cm}{}
                \includegraphics[scale=0.7]{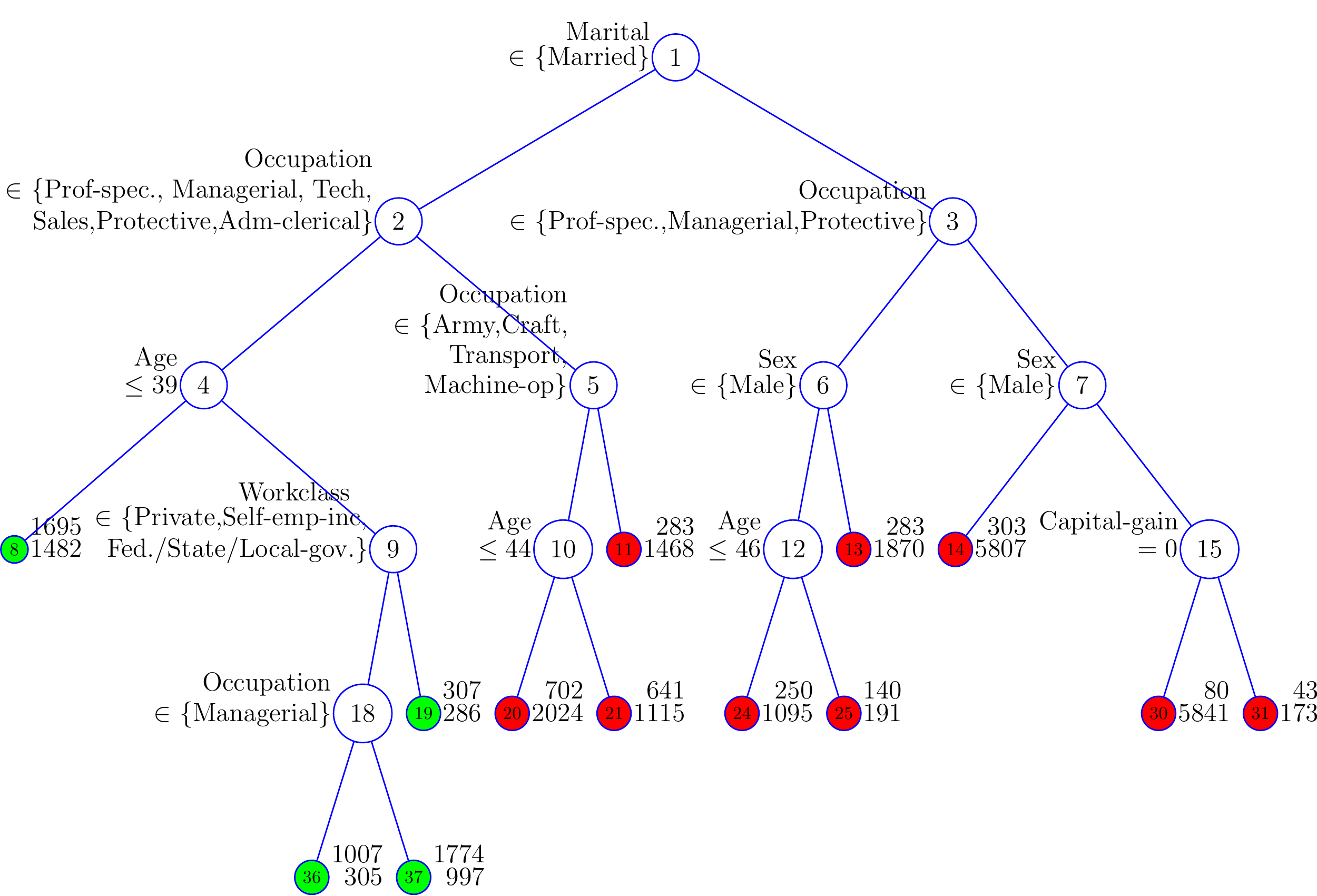}
\end{adjustwidth}
\caption[Model 3. PLUTO 1-SE tree with regularized multiple linear logistic regression leaves for the census income data.]{Model 3. PLUTO 1-SE tree with regularized multiple linear logistic regression leaves for the census income data. At each splitting node, an observation goes to the left if and only if the criterion to the left of the node is satisfied. At each terminal node, the numbers to the right are the numbers of observations that have annual income $> 50K$ (top) and $\leq 50K$ (bottom). \textcolor[rgb]{0.00,0.59,0.00}{Green} circle indicates more observations with annual income above 50K. \textcolor[rgb]{1.00,0.00,0.00}{Red} otherwise.}
\label{P5:ML1SE}
\end{figure}

Model 2 is smaller than Model 1. However, 21 multiple linear logistic regression node models may
still be excessive for interpretation. Therefore, we trim down the tree more using the 1-SE rule. The resulting PLUTO 1-SE tree, denoted as ``Model 3", is shown in Figure \ref{P5:ML1SE}. Model 3 has 13 terminal nodes. Table \ref{T5:MLnodes} lists the fitted multiple linear logistic regression model in these terminal nodes.

In Model 3, the data at first split on Marital Status, separating ``Married" from the rest. Both groups then split on Occupation. We observe from Figure \ref{P5:ML1SE} that the observations in the $\{\mbox{Married}\}\ \cap$ \{Professional specialty, Executive Managerial, Tech-support, Sales, Protective Service, Admin clerical\} group have greater high income rate compared with the other groups. The estimated coefficients for all the numerical variables, except Age, are in general consistent across all terminal nodes. The variable Age has negative estimated coefficients at Nodes 36, 37, 19, 21, and 25, indicating that the high income rate decreases as age increases. We find that all these nodes are groups where only mid-aged and senior persons are included. Therefore, Age has a nonlinear effect on income.

\begin{table}[!htb]
\begin{center}
\caption{Fitted multiple linear logistic regression models in the terminal nodes of PLUTO Model 3.}\label{T5:MLnodes}
\begin{tabular}{c|p{.13\linewidth}p{.13\linewidth}p{.13\linewidth}p{.13\linewidth}p{.13\linewidth}p{.13\linewidth}|}
\cline{2-7}
 & \multicolumn{6}{|c|}{Estimated Coefficients}\\
\hline
\multicolumn{1}{ |c|  }{Node ID} & (Intercept) & Age & Education-num & Capital-gain(K) & Capital-loss(K) & Hour \\
\hline
\multicolumn{1}{ |c|  }{8}  &\ -8.061 & \ 0.115 & \ 0.305 & \ 0.270 & \ 0.699 & \ 0.016 \\
\multicolumn{1}{ |c|  }{36} &\ -2.203 & -0.022 & \ 0.260 & \ 0.216 & \ 0.452 & \ 0.029 \\
\multicolumn{1}{ |c|  }{37} &\ -2.128 & -0.025 & \ 0.264 & \ 0.230 & \ 0.714 & \ 0.018 \\
\multicolumn{1}{ |c|  }{19} &\ -2.070 & -0.022 & \ 0.207 & \ 0.243 & \ 1.088 & \ 0.013 \\
\multicolumn{1}{ |c|  }{20} &\ -7.332 & \ 0.069 & \ 0.223 & \ 0.376 & \ 0.547 & \ 0.035 \\
\multicolumn{1}{ |c|  }{21} &\ -1.981 & -0.033 & \ 0.248 & \ 0.263 & \ 0.459 & \ 0.018 \\
\multicolumn{1}{ |c|  }{11} &\ -5.420 & \ 0.012 & \ 0.264 & \ 0.205 & \ 0.717 & \ 0.014 \\
\multicolumn{1}{ |c|  }{24} & -11.267 & \ 0.108 & \ 0.275 & \ 0.295 & \ 0.469 & \ 0.053 \\
\multicolumn{1}{ |c|  }{25} &\ -2.304 &  -0.024 & \ 0.212 & \ 0.161 & \ 0.518 & \ 0.010 \\
\multicolumn{1}{ |c|  }{13} &\ -8.836 & \ 0.032 & \ 0.310 & \ 0.369 & \ 0.637 & \ 0.033 \\
\multicolumn{1}{ |c|  }{14} & -10.611 & \ 0.059 & \ 0.309 & \ 0.445 & \ 0.759 & \ 0.046 \\
\multicolumn{1}{ |c|  }{30} & -13.169 & \ 0.055 & \ 0.399 &  & \ 0.608 & \ 0.064 \\
\multicolumn{1}{ |c|  }{31} & -14.243 & \ 0.042 & \ 0.463 & \ 0.514 &  & \ 0.065 \\
\hline
\end{tabular}
\end{center}
\end{table}

\subsection{Variable importance ranking}
\begin{figure}[H]
                \centering
                \includegraphics[width=0.8\textwidth]{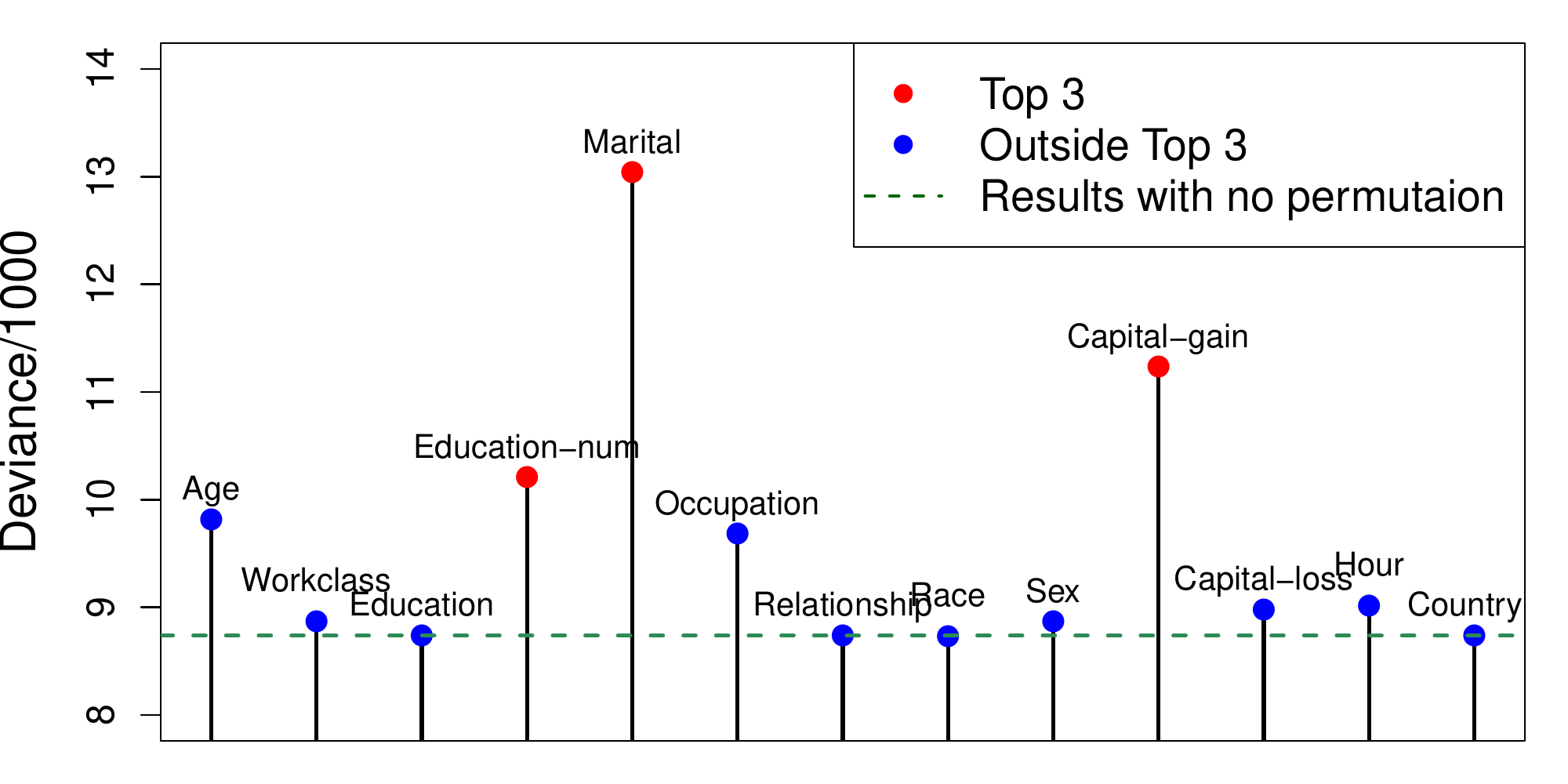}
                \includegraphics[width=0.8\textwidth]{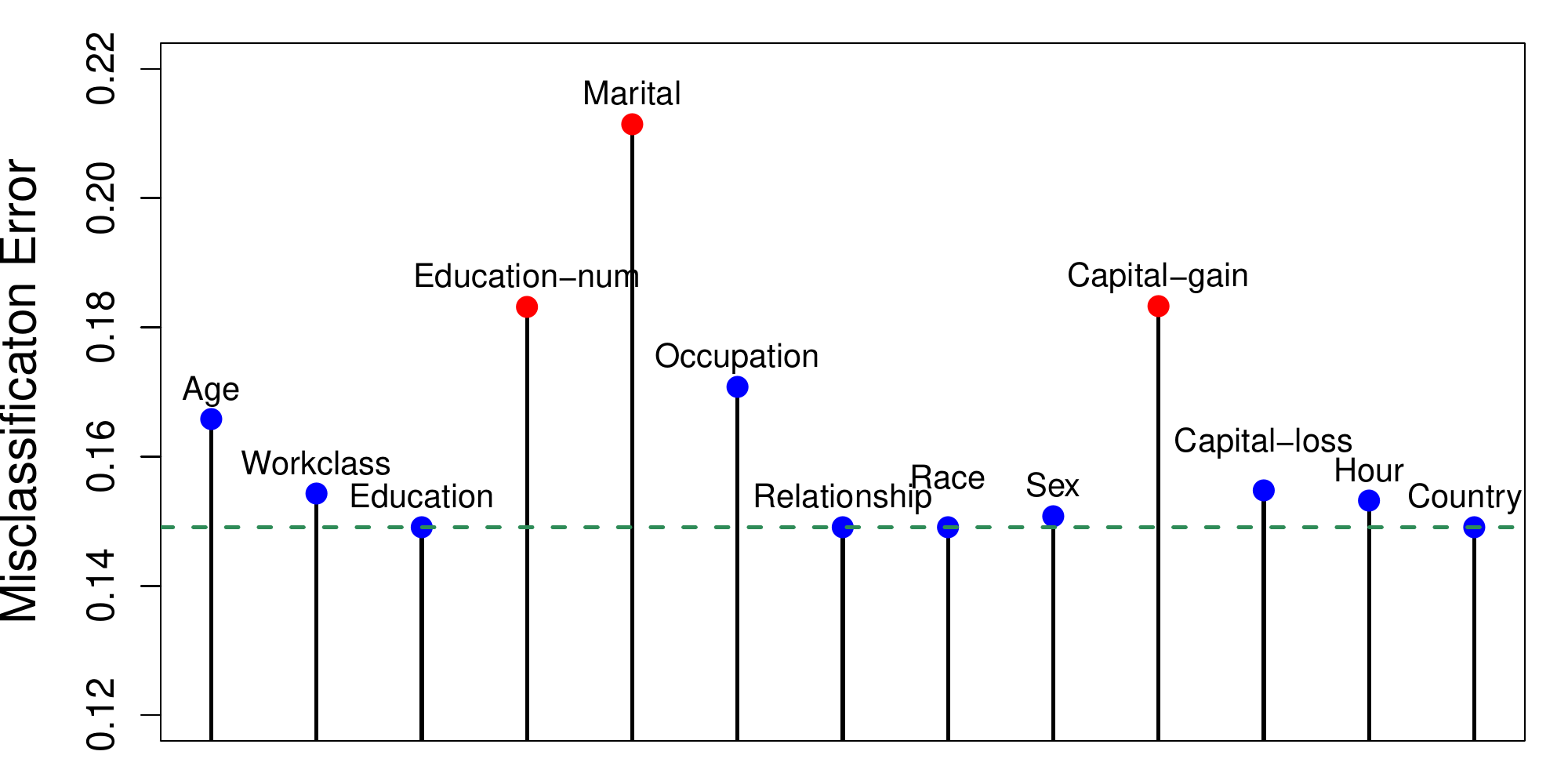}
                \includegraphics[width=0.8\textwidth]{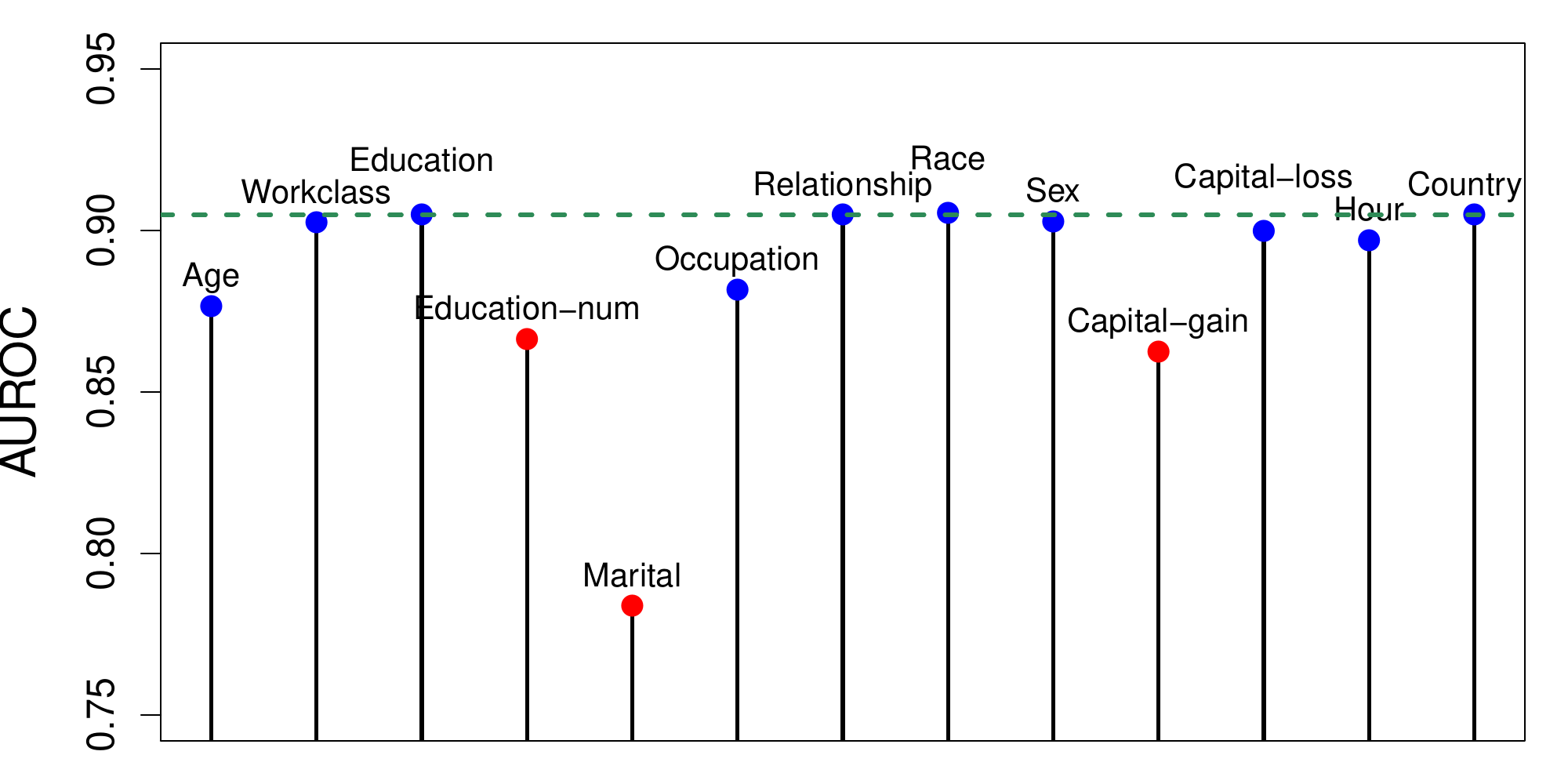}
\caption[Variable importance ranking for census income data]{Variable importance ranking for census income data based on trimmed deviance(top), misclassification error rate(middle), and AUROC(bottom).}
\label{P5:Imp}
\end{figure}

We employ the PLUTO importance ranking algorithm to explore variable importance in the census income data. Figure \ref{P5:Imp} displays the importance rankings based on trimmed deviance (top $1\%$ values discarded), misclassification error rate and AUROC, respectively. We used Model 2 on the independent testing dataset to compute these measures.

All three measures identify Marital, Capital-gain, and Education-num as the top three most important variables. PLUTO takes the average of the three sets of ranking scores, and obtains the final rank of importance, presented in Table \ref{T5:IMP}.

\begin{table}[!htb]
\begin{center}
\caption{PLUTO importance ranking for census income data}\label{T5:IMP}
\begin{tabular}{cccccc}
\hline
Rank & 1 & 2 & 3 & 4 & 5 \\
Variable & Marital & Capital-gain & Education-num & Age & Occupation \\
\hline
Rank &6 & 7 & 8 & 9 & 10 \\
Variable & Capital-loss & Hour & Workclass & Sex & Education\\
\hline
Rank & 11 & 12 &13 & &\\
Variable  & Relationship & Country & Race & &\\
\hline
\end{tabular}
\end{center}
\end{table}

\subsection{Summary}
The PLUTO tree models and PLUTO importance ranking show that Marital, Capital-gain, Education, Age, and Occupation are important in terms of determining whether a person has annual income above 50K. These results agree with common sense. Model 1 suggests that Capital-gain and Capital-loss have nonlinear effects on income. Model 3 further detected that Age may also have a nonlinear effect on income.

To compare the prediction accuracy between PLUTO and the algorithms listed in Table \ref{T5:pre.rslts}, we use Model 1, Model 2 and Model 3 to predict the same testing dataset. GLM, GLMNET/LASSO, LOTUS and MOB at their default settings, are also included in this comparison. However, MOB crashes on this dataset.
\begin{table}[!htb]
\begin{center}
\caption{Trimmed deviance, misclassification error rate and AUROC of the PLUTO, GLM, GLMNET and LOTUS algorithms on the census income testing dataset}\label{T5:comp}
\begin{tabular}{crrr}
  \hline\hline
  Model & $\mbox{DEV}'$ & MER\textsuperscript{\dag} & AUROC \\
  \hline
  PLUTO\_S\_0SE (Model 1) & 8477 & 0.145  &0.904 \\
  PLUTO\_M\_0SE (Model 2) & 8738 & 0.149  &0.905 \\
  PLUTO\_M\_1SE (Model 3) & 8805 & 0.151  &0.904 \\
  LOTUS                   & 8814 & 0.151  & 0.903 \\
  GLMNET                  & 8981 & 0.152  & 0.903\\
  GLM                     & 8870 & 0.153  & 0.902 \\
   \hline
   \multicolumn{4}{l}{\textsuperscript{\dag}\footnotesize{MER: Misclassification Error Rate.}}
\end{tabular}
\end{center}
\end{table}
%discussion

Table \ref{T5:comp} shows the prediction results on the census income data. The PLUTO algorithms more accurately predicted the testing datasets compared with GLM, GLMNET and LOTUS. PLUTO Model 1 out-performs the majority of the algorithms in Table \ref{T5:pre.rslts} based on the misclassification error rate, although FSS Naive Bayes and Naive Bayes Tree have lower values. However, PLUTO is not developed solely as a classification algorithm. Therefore, using misclassification error rate as the only measure of fit may give advantage to other classification algorithms.

%%%%%%%%%%%%%%%%%%%%%%%%%%%%%%%%%%%%%%%%%%%%%%%%%%%%%%%%%%%%%%%%%%%%%
\section{Conclusion and Future Work}
In this paper, we propose a new algorithm that combines recursive partitioning and logistic regression that we call PLUTO. Our algorithm is flexible and allows predictors to play different roles as needed while building the tree model. PLUTO also provides two node model options: best simple linear logistic regression model and regularized multiple linear logistic regression model. We employ the cyclical coordinate descent algorithm \citep{GLMNET} to fit multiple linear logistic regression with elastic net penalties, which enables PLUTO to handle high-dimensional data efficiently.

PLUTO controls selection bias by separating the split variable selection from split point/subset selection. It applies an adjusted chi-squared test to find the split variable instead of exhaustive search. To further control selection bias, we adopt the GUIDE \citep{guide} bootstrap bias correction method. PLUTO uses the minimal cost-complexity pruning algorithm of CART \citep{cart} to determine the final tree, which prevents PLUTO from underfitting or overfitting.

By comparing PLUTO with competing algorithms on twenty real datasets, we find that overall, the multiple linear PLUTO models predict more accurately than other algorithms. An application of PLUTO to the census income dataset further demonstrates the data interpretation ability of PLUTO.

In this paper we assume that the dataset is complete. When missing values occur in $X$, there are four ways to deal with them: (1) remove the observations containing missing values and use the complete cases only to grow the tree; (2) impute the missing values with sample means/modes; (3) impute the missing values by fitting regression or classification models with $X$ as the response; (4) consider the missing values as a separate class of the predictor and incorporate it in our split selection algorithms. The selection of missing value methods may be affected by various factors of the datasets such as the sample size, number of predictors, and type of predictors. Further study is needed to compare and evaluate these missing value methods.

We can extend the idea in this paper to multinomial and ordinal logistic regression. Multinomial logistic regression tree will allow response variables with more than two discrete outcomes, while ordinal logistic regression tree can be applied to ranking data. Both are useful tools for decision making.

\clearpage
\bibliographystyle{unsrtnat}
\bibliography{myref}

\end{document}